\documentclass{elsarticle}

\usepackage[utf8]{inputenc}
\usepackage{amsmath,amsfonts,amssymb,amsthm}
	\numberwithin{equation}{section} 
\usepackage{hyperref}       
    \hypersetup{
      pdftitle={Value functions for depth-limited solving in imperfect-information games},
      pdfauthor={Vojta Kovarik, Dominik Seitz, Viliam Lisy}
    }
\usepackage{url}            
\usepackage{booktabs}       
\usepackage{nicefrac}       
\usepackage{microtype}      

\usepackage[table]{xcolor}
\usepackage{wrapfig}
\usepackage{float}
\usepackage{rotating}

\usepackage{etoolbox}
\apptocmd{\thebibliography}{\raggedright}{}{}

\usepackage{todonotes}

\usepackage{graphicx}

\usepackage{xcolor}
\usepackage{forest} 
	\usetikzlibrary{shapes} 
	\usetikzlibrary{calc} 
\usepackage{xparse}	

\newlength{\nodesize}
\colorlet{chance_color}{black}
\colorlet{pl0_color}{chance_color}
\colorlet{chance_text}{white}
\colorlet{pl1_color}{magenta!50}
\colorlet{pl2_color}{cyan!50}
\colorlet{pl0_infoset_color}{pl0_color}
\colorlet{pl1_infoset_color}{magenta!75}
\colorlet{pl2_infoset_color}{cyan!75}
\forestset{
	basenode/.style = {draw,
		inner sep = 0,
		outer ysep = 0,
		minimum size = \nodesize,
		anchor = north
	},
	playernode/.style={basenode, 
		shape = regular polygon,
		regular polygon sides = 3,
	},
	pl1/.style={playernode, fill=pl1_color},
	pl2/.style={playernode, fill=pl2_color, shape border rotate=180},
	chance/.style = {basenode,
		fill=pl0_color, text=chance_text,
		circle,
		minimum size=0.75*\nodesize,
		l sep=0.4765\nodesize,
	},
	terminal/.style = {basenode,
		shape = regular polygon,
		regular polygon sides = 4,
		l sep=0.47\nodesize,
		minimum size = 1.07\nodesize
	}
}
\tikzset{
	partition/.style = {
		draw,
		rounded corners = 5,
		inner sep=0.1\nodesize,
	},
	infoset/.style = {
		partition,
		draw=pl1_infoset_color,
	},
	augmented/.style = {
		dashed
	},
	opponent/.style = {
		draw=pl2_infoset_color,
		inner sep=0.225\nodesize,
	},
	pl1_cl_infoset/.style = {infoset, yshift=-0.035\nodesize},
	pl2_cl_infoset/.style = {infoset,
		opponent,
		inner sep=0.1\nodesize,
		yshift=0.035\nodesize
	},
	public_state/.style = {
		partition,
		draw=pl0_infoset_color,
		inner sep=0.125\nodesize,
		rounded corners = 7,
	},
}
\tikzset{
	line_infoset/.style = {
		draw=pl1_infoset_color,
		dashed
	},
	opp_line_infoset/.style = {
		draw=pl2_infoset_color,
		dashed
	}
}
\newcommand{\corners}[1]{#1.corner 1)(#1.corner 2)(#1.corner 3}

\pgfdeclarelayer{bg}    
\pgfsetlayers{bg,main}  
    \usetikzlibrary{matrix}

\usepackage{thm-restate}

\theoremstyle{plain}
\newtheorem{theorem}{\protect\theoremname}
\newtheorem{lemma}{\protect\lemmaname}[section]
\newtheorem*{lemma*}{\protect\lemmaname}
\newtheorem{proposition}[lemma]{\protect\propositionname}

\newtheorem{corollary}[lemma]{\protect\corollaryname}

\theoremstyle{definition}
\newtheorem{definition}[lemma]{\protect\definitionname}
\newtheorem{example}[lemma]{\protect\examplename}
\newtheorem{remark}[lemma]{\protect\remarkname}
\newtheorem{problem}[lemma]{\protect\problemname}
\newtheorem{notation}[lemma]{\protect\notationname}
\newtheorem*{remark*}{\protect\remarkname}

\providecommand{\corollaryname}{Corollary}
\providecommand{\claimname}{Claim}
\providecommand{\definitionname}{Definition}
\providecommand{\lemmaname}{Lemma}
\providecommand{\notationname}{Notation}
\providecommand{\remarkname}{Remark}
\providecommand{\problemname}{Problem}
\providecommand{\propositionname}{Proposition}
\providecommand{\examplename}{Example}
\providecommand{\theoremname}{Theorem}
\providecommand{\conjecturename}{Conjecture}

\newcommand{\argmax}{\operatornamewithlimits{argmax}}

\newcommand{\R}{\mathbb{R}}
\newcommand{\defword}[1]{\textbf{\boldmath{#1}}}

\newcommand{\abs}[1]{\left|#1\right|}
\newcommand{\mc}{\mathcal}
\newcommand{\DLCFR}{$\textrm{DL-CFR}^+_{\textrm{NN}}\,$}
\newcommand{\gv}[2]{\textnormal{gv}_{#1}(#2)}
\newcommand{\vv}{\mathbf{v}}        
\newcommand{\VV}{\mathbf{V}}
\newcommand{\vf}[3]{\vv_{#1}^{#2}(#3)}
\newcommand{\Vf}[3]{\mathbf{V}_{#1}^{#2}(#3)}
\newcommand{\vfNP}[2]{\vv_{#1}^{#2}}

\newcommand{\rp}[3]{P_{#1}^{#2}(#3)}
\newcommand{\hv}[3]{v_{#1}^{#2}(#3)}
\newcommand{\hvNP}[2]{v_{#1}^{#2}}
\newcommand{\Iv}[3]{V_{#1}^{#2}(#3)}
\newcommand{\IvNP}[2]{V_{#1}^{#2}}
\newcommand{\hav}[4]{q_{#1}^{#2}(#3,#4)}

\newcommand{\Iav}[4]{Q_{#1}^{#2}(#3,#4)}

\newcommand{\expl}[3]{\textnormal{expl}_{#1}^{#2}(#3)}
\newcommand{\range}[2]{\textnormal{rng}^{#1}{(#2)}}
\newcommand{\ims}[1]{\textnormal{ims}(#1)}
\newcommand{\Succ}[1]{\textnormal{Desc}(#1)}
\newcommand{\hvCf}[3]{v_{#1,\textnormal{cf}}^{#2}(#3)}
\newcommand{\IvCf}[3]{V_{#1\textnormal{cf}}^{#2}(#3)}
\newcommand{\IvCfNP}[2]{V_{#1\textnormal{cf}}^{#2}}
\newcommand{\havCf}[4]{q_{#1,\textnormal{cf}}^{#2}(#3,#4)}
\newcommand{\IavCf}[4]{Q_{#1\textnormal{cf}}^{#2}(#3,#4)}

\newcommand{\trunk}{{\mathcal T}}
\newcommand{\trunkS}{\sigma^\trunk}
\newcommand{\DLu}[2]{u_{#1}^{ \vv}(#2)}
\newcommand{\DLhv}[3]{\hv{#1}{#2,\vv}{#3}}
\newcommand{\DLIv}[3]{\Iv{#1}{#2,\vv}{#3}}
\newcommand{\DLIvCf}[3]{\IvCf{#1}{#2,\vv}{#3}}
\newcommand{\DLIavCf}[4]{\IavCf{#1}{#2,\vv}{#3}{#4}}
\newcommand{\OptE}[2]{\textnormal{OE}{(#1,#2)}}
\newcommand{\portfolio}{\mathbb P}
\newcommand{\pub}{{\textnormal{pub}}}
\newcommand{\priv}{{\textnormal{priv}(p)}}

\newcounter{vkNoteCounter}

\newcounter{dsNoteCounter}

\newcommand{\edit}[1]{\textcolor{purple}{#1}}
 \renewcommand{\edit}[1]{#1}

\journal{Artificial Intelligence Journal}

\begin{document}

\begin{frontmatter}



\title{Value Functions for Depth-Limited Solving\texorpdfstring{\\}{ }in \edit{Zero-Sum} Imperfect-Information Games}


\author{Vojtěch Kovařík\fnref{label1}}
\author{Dominik Seitz\fnref{label1}}
\author{Viliam Lisý\corref{cor1}}
    \ead{viliam.lisy@agents.fel.cvut.cz}
\author{\\Jan Rudolf}
\author{Shuo Sun}
\author{Karel Ha}
\fntext[label1]{These two authors contributed equally.}
\cortext[cor1]{To whom correspondence should be addressed.}
\address{Artificial Intelligence Center, FEE,\\
    Czech Technical University in Prague,\\
    Prague, Czech Republic
}

\begin{abstract}
    We provide a formal definition of depth-limited games together with an accessible and rigorous explanation of the underlying concepts, both of which were previously missing in imperfect-information games.
The definition works for an arbitrary extensive-form game and is not tied to any specific game-solving algorithm.
Moreover, this framework unifies and significantly extends three approaches to depth-limited solving that previously existed in extensive-form games and multiagent reinforcement learning
but were not known to be compatible.
A key ingredient of these depth-limited games are value functions.
Focusing on two-player zero-sum imperfect-information games, we show how to obtain optimal value functions and prove that public information provides both necessary and sufficient context for computing them.
We provide a domain-independent encoding of the domains that allows for approximating value functions even by simple feed-forward neural networks, which are then able to generalize to unseen parts of the game.
We use the resulting value network to implement a depth-limited version of counterfactual regret minimization.
In three distinct domains, we show that
    the algorithm's exploitability is roughly linearly dependent on the value network's quality
    and that it is not difficult to train a value network with which depth-limited CFR's performance is as good as that of CFR with access to the full game.

\end{abstract}


\end{frontmatter}


\section{Introduction}\label{sec:intro}

Sequential decision making is a key challenge in AI research. As the number of consequent decisions increases, the size of the state space blows up exponentially, to the point where even modern computer clusters soon become unable to even enumerate all the states. In perfect information problems, this issue is typically overcome by replacing the states below a certain depth by a value function.
This technique vastly reduces the effective size of the game, which is essential for both minimax-like search and reinforcement learning. In imperfect information problems, value functions are much more complex since they depend on the agent’s belief about the current state. The situation gets even more complicated in multiagent imperfect information problems, where values additionally depend on each agent’s belief about other agents’ beliefs (etc.). Despite these challenges, recent results in poker \cite{DeepStack,Libratus,DeepRole} illustrate that depth-limited approaches can be successful even in this setting.

Unfortunately, many of the key concepts that enabled the recent results were introduced informally or tied to specific domains.
As a result, it is unclear how to adapt these results to new domains, the process is time-consuming, and the theoretical properties of the resulting algorithms are unknown.
For example, the value function used in \cite{DeepStack} takes as input the probabilities of different poker hands that a player could be holding.
This makes sense in poker, but what would be the input if we tried a similar approach in blind chess, scrabble, or phantom tic-tac-toe?
How would we train a value function in these domains and how would we know whether it is ``good''?
And what if we wanted to use the value function in combination with a different algorithm? How would we do it, and would it ``work''?
What does ``good'' and ``work'' mean in this context?
In this work, we aim to answer these questions by providing a solid theory of depth limited games and value functions and illustrating it by experiments on several domains.

\subsection{Outline and Contributions}\label{sec:outline-and-contributions}

In summary, this paper
    (1) provides a theory of depth-limited methods and value functions that unifies three recent approaches \cite{DeepStack,brown2018depth,wiggers2016structure},
    (2) formulates all required concepts in an accessible and domain-independent way, and
    (3) experimentally demonstrates that depth-limited solving is a viable and robust option for a range of imperfect information games beyond poker,
        and can be done without the need for hand-crafting domain-specific features.
        
\medskip

In more detail, we start with a brief background on EFGs (\textbf{Section~\ref{sec:EFGs}}).
In \edit{\textbf{Section~\ref{sec:background:values}}}, we define the key concepts -- expected utilities of histories and information sets, reach probabilities, and beliefs  -- in a unified way that is consistent with previous literature but much more intuitive and easy to use.
We then provide several technical propositions which \edit{capture the intuitive properties of values in EFGs and substantially simplify our proofs}.
\edit{A more detailed outline of this section is provided in its introduction
    (and the same is true for the other longer sections, i.e., Sections~\ref{sec:theory} and Section~\ref{sec:empirical}).}

\edit{In \textbf{Section~\ref{sec:theory}}, we present the key theoretical contributions of the paper.}
First, we look at imperfect-information games and propose domain- and algorithm- independent notions of value functions and depth-limited games.
We also describe (the depth-limited versions of) various algorithmic problems such as game-value computation, equilibrium computation, and best response computation.
Our goal for each of these problems is that whenever we find a solution of the depth-limited version of the problem and plug it into the full game, it should fulfil the role of a (partial) solution to the non-depth-limited version of the problem.
For example, a depth-limited Nash equilibrium should coincide with a standard Nash equilibrium in all decision-points above the depth limit. However, even some promising choices of value functions can fail to work for some of the above problems. We thus describe a natural hierarchy of conditions on value functions and formally prove that each of them fulfils the above goal for a different class of computational problems. Moreover, we observe that with the proposed formalization, the two previously separate approaches to depth-limited solving of imperfect information games -- value functions \cite{DeepStack} and multivalued states \cite{brown2018depth} -- can be viewed as two instances of a single unifying framework.
After studying the fundamental properties of value functions, we discuss methods for representing value functions more efficiently. One part of this endeavour is encoding the functions’ input and output more compactly --- we show that value functions can be defined on either individual histories or information sets and that the two representations can be translated to each other. Moreover, we formally prove that value functions can be factorized based on public information (or common knowledge) and that this factorization cannot be further refined in general. The second part is approximating value functions by neural networks, which is likely to be easier if there is a unique approximation target. Unfortunately, we see that there can sometimes be multiple suitable value functions. Failing uniqueness, we investigate whether the set of suitable value functions is at least well-behaved -- i.e., convex. We prove that this is true for certain types of value functions and pose the general question as an open problem. However, we remain optimistic about value function approximation since the non-uniqueness did not prove to be a problem in practice.

While all of the above results are presented using the extensive-form game formalism, \textbf{Section~\ref{sec:fog}} explains how all of them apply the POSG- (partially-observable stochastic game; \cite{POSGs2004hansen}) and FOSG- (factored-observation stochastic game; \cite{FOG}) formalisms used in multiagent reinforcement learning.

\edit{The experimental part of the paper (\textbf{Section~\ref{sec:empirical}})} focuses on the depth-limited version of counterfactual regret minimization (CFR), an algorithm which seems particularly promising due to its recent successes \edit{in poker \cite{DeepStack,Libratus,Pluribus} and other domains \cite{pog,celli2019learning}.}
We show that any game represented as a FOSG admits a unified representation of inputs and outputs to the value function.
We demonstrate, in three domains with very different properties, that this representation can serve as a suitable encoding for a neural network. This encoding allows for an accurate approximation even with the simplest feed-forward architecture. We show that when this value network is plugged into depth-limited CFR, the algorithm produces strategies with low exploitability.
While doing so, we experiment with various loss functions in an attempt to identify those which serve best as a proxy for exploitability when used in conjunction with depth-limited CFR.
To our surprise, the algorithm’s performance is not very sensitive to this parameter.
We also investigate whether the trained value function is robust to differences between the training distribution and the one demanded by depth limited CFR. In all three domains, the trained value function generalizes well to unseen situations. Furthermore, we investigate the dependence of the quality of the learned value function on the amount of training data and key hyper-parameters.

\edit{Finally, we summarize the most-related existing works (Section~\ref{sec:related_work}) and present our conclusions (Section~\ref{sec:conclusion}).}
    \subsection{Novelty of the Theoretical Contribution}\label{sec:novelty}

\edit{Since the main contribution of Section~\ref{sec:background:values} is in formalizing and organizing content that is well-known or simple, the results presented there are \textit{not novel unless stated otherwise}.
The corresponding novel content is listed in the introduction to this section.
On the other hand, all content in Section~\ref{sec:theory} (and Appendix~\ref{sec:uniqueness}) is \textit{novel unless stated otherwise}.
Finally, an important novel part of the paper is the domain-independent encoding proposed in Section~\ref{sec:sub:encoding}.}
\section{Extensive-Form Games}\label{sec:EFGs}

In this section, we formally describe the EFG model used throughout the paper.
We make a slight deviation from the historically standard (but much less convenient) definition by assuming that information sets of each player are defined even over terminal states and states in which it is the opponent's turn.
This modification is consistent with \cite{brown2018depth} and \cite{FOG}.

A extensive-form game $G$ can be described by:
$\mc{H}$ -- a finite set of \defword{histories} (states, nodes), representing sequences of actions.
    We use $g \sqsubset h$ to denote the fact that $g$ is equal to or a prefix of $h$.
$\mc{Z}$ -- the set of terminal histories (those $z\in \mc H$ which are not a prefix of any other history).
%
$\mc N = \{1, \dots, N\}$ --- the \defword{player set}, where $c$ is a special player, called ``chance'' or ``nature''.
The \defword{player function} $\mc P : \mc H \setminus \mc Z \rightarrow \mc N \cup \{ c \}$ denotes which player acts in the given \edit{non-terminal} history.
$\mc A(h) := \mc A_p(h) := \{ a \, | \ ha \in \mc H \}$ is the set of \defword{actions} available to $p=\mc P(h)$ at $h$.\footnote{To simplify some of the formulations, we assume that the each $p'\neq \mc P(h)$ takes a \textit{noop} action at $h$ (which doesn't do anything, and isn't explicitly written into the history, i.e., $h \, \textnormal{noop} = h$).}
%
\defword{The strategy of chance} is a fixed probability distribution $\sigma_c$ over actions available at chance nodes (those $h$ where $\mc P(h) = c$).
The \defword{utility function} $u=(u_p)_{p\in \mc N}$ assigns to each terminal history $z$ a reward $u_p(z)\in \R$ received by each player upon reaching $z$.

The \defword{information-partitions} $\mc I = (\mc I_p)_{p\in \mc N}$, where each $\mc I_p$ is a partition of $\mc H$, capture the imperfect information of $G$.
    If $g, h \in \mc H$ belong to the same \defword{information set} (infoset) $I\in \mc I_p$ then $p$ cannot distinguish between them.
    For each $I \in \mc I_p$, the available actions $\mc A_p(I) := \mc A_p(h)$ are thus the same for each $h \in I$.
To help decompose the game, we consider the \defword{public tree} $\mc S$ which partitions $\mc H$ into \defword{public states}\footnote{\edit{Intuitively, public states partition the game based on the information that is public --- e.g., cards laying face up on the table or actions visible to all players.
}} $S \in \mc S$ which are closed\footnote{\edit{That is, $(\forall p \in \mc N) (\forall I\in \mc I_p) : I \subset S \lor I \cap S = \emptyset$.}} under membership within infosets.
We only consider games where players have \defword{perfect recall}, i.e., they remember their past actions and infosets visited so far.\footnote{\edit{Formally: denote by $s_p(h)$ the sequence of infosets encountered and actions taken by player $p$ along $h$. Player $p$ has perfect recall if we have $g,h \in I \in \mc I_p$ $\implies$ $s_p(g)=s_p(h)$.}}\footnote{Together with $\mc I_p$ covering full $\mc H$ and players formally taking a \textit{noop} action in histories where they are inactive, this has the desirable \cite{timeability} implication of enforcing \defword{timeability} of all games \edit{(informally, the infoset structure being compatible with the players having a sense of time passing \cite{timeability})}. For more details, see \cite{FOG}.)}
A \defword{behavioral strategy} $\sigma_p \in \Sigma_p$ of player $p$ assigns to each $I\in \mc I_p$ a probability distribution $\sigma_p(I)$ over actions $\mc A_p(I)$.
A tuple $\sigma = (\sigma_p)_{p\in \mc N}$ is called a \defword{strategy profile}.
\edit{By $u_p(\sigma)$, we denote the expected utility for player $p$ if all players play according to $\sigma$.}

The~profile $\sigma$ is an \defword{$\epsilon$-Nash equilibrium} if the benefit of switching to some alternative $\sigma'_p$ is limited by $\epsilon$, i.e., if 
$u_p(\sigma_p', \sigma_{-p}) \leq u_p(\sigma) + \epsilon$ (where $-p$ denotes all the players other than $p$).
When $\epsilon = 0$, the profile is called a Nash equilibrium (NE) \edit{and we write $\sigma \in \textnormal{NE}(G)$}.
\edit{In the remainder of the paper, we only consider}t \defword{two-player zero-sum games}, where $\mc N = \{1, 2\}$ and $u_2 = - u_1$.
It is a standard result that in two-player zero-sum games, all $\sigma^*\in \textrm{NE}(G)$  have the same utility, called \defword{game value} $u_p(\sigma^*) = \max_{\sigma_p \in \Sigma_p} \min_{\sigma_{-p} \in \Sigma_{-p}} u_p(\sigma) =: \gv{p}{G}$.
The \defword{exploitability} $\expl{}{}{\sigma}$ of $\sigma \in \Sigma$ is the average of exploitabilities $\expl{p}{}{\sigma_p}$, where $\expl{p}{}{\sigma_p} := \gv{p}{G} - \min_{\sigma_{-p} \in \Sigma_{-p}} u_p(\sigma_p,\sigma_{-p})$.
\section{Clarifying Key Concepts in Imperfect Information Games}\label{sec:background:values}


\edit{In this section, we describe the remaining prerequisites of this paper.
Unfortunately, while most of them are considered standard and well-known by the community around EFGs and CFR,
    they have often not been previously published or made fully formal,
    and some of the existing definitions cannot be applied to unreachable parts of the game
    (which prevents reasoning about counterfactual scenarios where one of the players changes their strategy).
Moreover, many of the concepts are ``not very accessible'' to new audience, largely because their definitions are scattered across many different conference papers with strict page limits.
The goal of this section is to remedy this situation by explaining all of these concepts in one place, in a way that would be fully formal and accessible to readers not yet fully familiar with the CFR literature.
}
    
\edit{Given the goal of this section, it is unsurprising that most of the content is not novel --- our contribution is in organizing it, finding the right formulations (which was sometimes surprisingly difficult), and coming up with the formal proofs (which was typically straightforward once we had the right formulations).
However, to our best knowledge, \textbf{several of the observations are novel, and might be of interest even to readers well-versed in the CFR literature}.\footnote{
        \edit{These results are not \textit{particularly} difficult to prove and some other researchers might have arrived at them independently.
        However, we remember not being aware of these results prior to starting the work on this paper, and we consider all of them useful and non-obvious.}
    }
    \textbf{These are:}
        (a) The observation that a number of properties of history- and infoset- values are not just a corollary of the definition, but in fact equivalent to it (the ``moreover'' parts of Lemma~\ref{lem:hist-value:characterization} and Theorem~\ref{thm:infoset-v-characterization}).
        (b) The observation that $p$'s reach probability of (info)sets should be defined as in eq. \ref{eq:rp-p-set}, not as a sum of $p$'s reach probabilities of histories in that (info)set.
        (c) Using an approach inspired by trembling-hand equilibria to extend beliefs to unreachable infosets (Definition~\ref{def:infoset-belief}).
        (d) The observation that once we know counterfactual values of leaf infosets, backpropagating them through the infoset tree is simple (and much simpler than backpropagating normal values; Theorem~\ref{thm:infoset-cfv-properties}).
}

We first describe the key concepts for histories (Section~\ref{sec:background:values:hist}).
We then extend these to sets of histories, and in particular to infosets (Section~\ref{sec:sub:infoset-reach}).
In Section~\ref{sec:cfr}, we describe counterfactual values and the counterfactual regret minimization algorithm, which plays an important role in subsequent sections.

\subsection{Expected Utilities of Histories}\label{sec:background:values:hist}

Before describing expected utilities, we need the concept of a \defword{reach probability} of a history.
\edit{This definition and others in this section rely on the intuition that we are playing the game $G$ using some strategy (in this case $\sigma$), we are currently in some situation in $G$ (here the root), and we are asking about the probability of some event in this playthrough of $G$ (here encountering $h$).
To avoid confusion, the less-formal definition (here eq. \ref{eq:hist-reach}) is always accompanied by its fully-formal equivalent (here eq. \ref{eq:hist-reach-formal}):
}

\begin{align}
\rp{}{\sigma}{h} \ 
    & :=
    \textbf{Pr}_\sigma \left[ \, h \textnormal{ reached during the course of the game} \right]
    \label{eq:hist-reach}
    \\
& :=
    \edit{ \Pi_{ga \sqsubset h} \ \sigma(g,a)} .
    \edit{\footnotemark}
    \label{eq:hist-reach-formal}
\end{align}
\footnotetext{
    Where \edit{$\sigma(g,a)$} stands for $\sigma_{\mc P(g)}(I,a)$, where $g \in I\in \mc I_{\mc P(g)}$ \edit{(resp. $\sigma_c(g,a)$ when $\mc P(g)=c)$}.
}

Reach probabilities can be decomposed into the player-$p$ component, and the corresponding $p$'s \defword{counterfactual reach probability} (i.e., the chance of reaching $h$ in the counterfactual (cf.) situation where $p$ aims to do so):
\begin{align}
\rp{p}{\sigma}{h} \ & := \textbf{Pr}_\sigma \left[ \, h \textnormal{ reached} \mid \textnormal{every $p'\neq p$ plays to reach $h$} \right] \\
& := \edit{ \Pi_{ga \sqsubset h, \, \mc P(g) = p} \, \sigma(g,a) , }\\
\rp{-p}{\sigma}{h} & := \textbf{Pr}_\sigma \left[ \, h \textnormal{ reached} \mid \textnormal{$p$ plays to reach $h$} \right] \\
& := \edit{ \Pi_{ga \sqsubset h, \, \mc P(g) \neq p} \, \sigma(g,a) . }
\end{align}
\edit{These definitions implies that for every $p$,}
\begin{equation}\label{eq:rp's:product}
\rp{}{\sigma}{h} = \rp{1}{\sigma}{h} \rp{2}{\sigma}{h} \rp{c}{\sigma}{h} = \rp{p}{\sigma}{h} \rp{-p}{\sigma}{h} .
\end{equation}

We say that a node $h$ is \defword{reachable}, resp. \defword{counterfactually reachable} by $p$, under $\sigma$ if $\rp{}{\sigma}{h} > 0$, resp. $\rp{-p}{\sigma}{h} > 0$.
When the reach probability is zero, $h$ is said to be \defword{unreachable}, resp. \defword{counterfactually unreachable}.
%
We extend reach probabilities to \textit{paths}\footnote{\edit{When $h$ isn't an extension of $g$ (or equal to it), we set $\rp{}{\sigma}{g,h} = 0$.}}, obtaining an analogue of \eqref{eq:rp's:product}:
\begin{align}\label{eq:rp:path}
\rp{}{\sigma}{g,h} & := \textbf{Pr}_\sigma \left[ \, h \textnormal{ reached } \mid \textnormal{current node is } g \, \right]
\edit{= \Pi_{g \sqsubset h', \, h'a \sqsubset h} \sigma(h',a)} ,
\\
\rp{}{\sigma}{g,h}
& = \edit{\rp{1}{\sigma}{g,h} \rp{2}{\sigma}{g,h} \rp{c}{\sigma}{g,h} = \rp{p}{\sigma}{g,h} \rp{-p}{\sigma}{g,h} .}
    \label{eq:rp-hist-product}
\end{align}

As a useful piece of terminology, we say that a set $H\subset\mc H$ is \defword{thin} if no two nodes $g,h\in H$, $g\neq h$, satisfy $g \sqsubset h$ (i.e., $H$ is an antichain w.r.t. $\sqsubset$).
Otherwise, $H$ is \defword{thick}.
A thin set $L \subset \mc H$ to which no $h \notin L$ can be added without making it thick (i.e., a maximal antichain w.r.t. $\sqsubset$) is called a \defword{slice} (through $\mc H$).
While we typically imagine that $L$ slices $\mc H$ ``somewhere in the middle'', the trivial examples of slices are the singleton $\{\textnormal{root}\}$ and the set of all leaves $\mc Z$.
However, slices do not need to be strictly ``horizontal'' --- that is, they can contain histories with different lengths.

\medskip

We now define the \defword{value of} $h \in \mc H$ (for $p \in \mc N$ under $\sigma$) and the corresponding \defword{action values} (when $h \notin \mc Z$) as\footnote{\edit{Since $\hv{p}{\sigma}{h}$ is the \defword{expected utility} under $\sigma$, it would also make sense to denote these values simply as $u_p^\sigma(h)$.
The main reason for our preference for $v$ and $q$ is its similarity to the notation used in reinforcement learning, which serves to bring attention to the similarity between the mathematical properties of these objects.}}
\begin{align}\label{eq:hist-value}
\hv{p}{\sigma}{h} \ \ \ & := \mathbf{E}_\sigma \left[ u_p(z) \mid z \in \mc Z, \textnormal{ current history is } h \right] \\
& := \sum_{z\in \mc Z} \rp{}{\sigma}{h,z} u_p(z) , \\
\hav{p}{\sigma}{h}{a} & := \mathbf{E}_\sigma \left[ u_p(z) \mid z\in \mc Z, \textnormal{ current history is } h, \textnormal{ $a$ taken at $h$} \right] \\
& := \sum_{z\in \mc Z} \rp{}{\sigma}{ha,z} u_p(z)
.\label{eq:act-hist-value}
\end{align}

Lemma~\ref{lem:hist-value:characterization} summarizes the properties of expected values of histories:
(1) states that, by definition, values are calculated as expectations over utilities of terminal states.
Moreover, by (2), this is equivalent to each history-value being the expectation of values over the history's children.
\edit{(2') restates (2) from $p$'s point of view, by saying that} these expectations can be expressed as weighted sums, where the weights are either $p$'s action probabilities or the counterfactual reach probabilities probabilities of reaching the child from $h$.
An important consequence of these properties is (3), which claims that values can also be computed as expectations over any \textit{slice} $L$ through the game tree (below the given history $h$\edit{, i.e., satisfying $\nexists g \in L \setminus \{h\} :  g \sqsubset h$}).

\begin{restatable}[Characterization of $\hvNP{p}{\sigma}$]{lemma}{histValueCharacterization}\label{lem:hist-value:characterization}
For any $p\in \mc N$ and $\sigma \in \Sigma$,
$v(\textnormal{root}) = u_p(\sigma)$ and
the values $\hvNP{p}{\sigma}$ and $\hav{p}{\sigma}{h}{a} := \hv{p}{\sigma}{ha}$ have the following properties
\begin{itemize}
\item[(1)]   $\hv{p}{\sigma}{h} = \mathbf{E}_\sigma \left[ u_p(z) \mid z \in \mc Z, \textnormal{ current history is } h \right]$ for every $h \in \mc H$.
\item[(2)] $\hvNP{p}{\sigma} = u_p$ on $\mc Z$ 	and on $\mc H \setminus \mc Z$, we have $\hv{p}{\sigma}{h} = \sum_{a \in \mc A(h)} \sigma(h,a) \hv{p}{\sigma}{ha}$.
\item[(2')] $\hvNP{p}{\sigma} = u_p$ on $\mc Z$ and on $\mc H \setminus \mc Z$, we have
	\begin{align*}
		\hv{p}{\sigma}{h} =
		\begin{cases}
			\sum_{a \in \mc A_p(h)} \sigma_p(h,a)\hav{p}{\sigma}{h}{a}		& \textnormal{ when } \mc P(h)=p \\
			\sum_{a\in \mc A(h)} \rp{-p}{\sigma}{h,ha} \hv{p}{\sigma}{ha}	& \textnormal{ when } \mc P(h) \neq p .
		\end{cases}
	\end{align*}
\item[(3)] $\hvNP{p}{\sigma} = u_p$ on $\mc Z$ and on $\mc H \setminus \mc Z$, we have $\hv{p}{\sigma}{h} = \sum_{h' \in L } \rp{}{\sigma}{h,h'} \hv{p}{\sigma}{h'}$ for every slice $L$ through $\mc H$ below $h$.
\end{itemize}
Moreover, each of these conditions can be used as an equivalent definition of $\hvNP{p}{\sigma}$ (i.e., it automatically implies all the others).
\end{restatable}

These properties and their equivalence are used -- typically implicitly -- in essentially all related proofs (e.g., they imply that it is correct to talk about the value of a history $h$ even when the strategy above $h$ is unknown).
While this result is by no means surprising, it provides an intuition for what properties we might get for infosets values.


\begin{proof}
\edit{First, applying (1) to $h= \textnormal{root}$ gives $v(\textnormal{root}) = u_p(\sigma)$.
Second, note that
(2) and (2')
are equivalent (since $\sigma(h,a)$ is equal to either $\sigma_p(h,a)$ or $\rp{-p}{\sigma}{h,ha}$, depending on $\mc P(h)$ and $\hv{p}{\sigma}{ha} = \hav{p}{\sigma}{h}{a}$ when $\mc P(h)$).}

\edit{Finally, for the remaining equivalences, consider the following lemma (whose proof is in the appendix, together with all other proofs not shown in the main text).}

\begin{restatable}{lemma}{treeLemma}\label{lem:tree-characterization}
\edit{Let $(T,\sqsubset)$ be a finite tree, $Z\subset T$ its leaves, $f : Z \to \R$, and $P : T^2 \to [0,1]$ a function s.t. $P(t,t) = 1$, $P(t,s) = 0$ when $\neg (t \sqsubset s)$, and $P(s,u) = P(s,t) P(t, u)$ when $s \sqsubset t \sqsubset u$.
Then the following are equivalent for $F : T \to \R$:
\begin{itemize}
    \item[(a)] $F(t) = \sum_{z \in Z} P(t,z) f(z)$ for $t \in T$,
    \item[(b)] $F(z) = f(z)$ on $Z$ and $F(s) = \sum_{t \in \ims{s}} P(s,t) F(t)$ for $s \in T \setminus Z$ (where $\ims{s}$ is the set of all immediate successors of $s$ in $T$),
    \item[(c)] $F(z) = f(z)$ on $Z$ and $F(s) = \sum_{t \in L} P(s,t) F(t)$ for $s \in T \setminus Z$, whenever $L$ is a slice through $T$ below $s$.
\end{itemize}
}
\end{restatable}

\edit{
\noindent The equivalences
(1) $\iff$ (2) $\iff$ (3)
are now an immediate corollary of the lemma, using $(T, \sqsubset) := (\mc H, \sqsubset)$, $P(h,h') = \rp{}{\sigma}{h,h'}$, $f(z) := u_p(z)$, and $F(h) := \hv{p}{\sigma}{h}$.
}
\end{proof}


\subsection{Generalization to Information Sets}

To talk about expected values of infosets, we first need to extend reach probabilities \edit{(eq. \ref{eq:hist-reach}-\ref{eq:rp-hist-product})} to infosets and define beliefs.
We start by defining reach probabilities of arbitrary subsets of $\mc H$ and proving that in the specific case of infosets, reach probabilities can be factored into the contributions of individual players.
Afterwards, we show players' beliefs can be defined even over infosets that are unreachable, which allows for computing reach probabilities for arbitrary paths through the infoset tree $\mc I_p$.
This enables us to prove that infoset values' behaviour is similar to that described in Lemma~\ref{lem:hist-value:characterization} for history values (Theorem~\ref{thm:infoset-v-characterization}).

\paragraph{Reach Probabilities and Their Factorization}\label{sec:sub:infoset-reach}

First of all, we see that the reach probability $\rp{}{\sigma}{H}$ of a general set $H\subset \mc H$ can be computed by summing the reach probabilities of $H$'s \defword{upper frontier} \cite{halpern2016upperFrontier}
\begin{align}
\rp{}{\sigma}{H}
\ & := \textbf{Pr}_\sigma \left[ \, H \textnormal{ reached during the course of the game} \right] \\
& \, := \sum \left\{ \rp{}{\sigma}{h} \mid h \in H \ \& \ (\nexists g \in H, \, g \neq h) : g \sqsubset h \right\} . \label{eq:rp's:as-sum}
\end{align}

\edit{\textbf{Assumption:} We can deal with reach probabilities of infosets and public states by always taking the sum over their upper frontier.
However, this would be somewhat tedious without bringing new insights.
Instead, we will thus \textbf{assume that all infosets and public states are thin} (i.e., equal to the upper frontier).}
Under this assumption, \eqref{eq:rp's:as-sum} simplifies to
    $\rp{}{\sigma}{I} = \sum_{h\in I} \rp{}{\sigma}{h}$ for $I \in \mc I_p$,
    resp. $\rp{}{\sigma}{S} = \sum_{h\in S} \rp{}{\sigma}{h}$ for $S\in \mc S$.

As far as \textit{intuitive} (yet also formal) definitions go, generalizing $p$'s reach probability and its counterfactual counterpart to sets is straightforward, but \edit{perhaps non-obvious}\footnote{
    \edit{If we naively defined $\rp{p}{\sigma}{H}$ as a sum $\sum_{h\in H} \rp{p}{\sigma}{h}$, the sum could easily end up strictly larger than $1$, and the resulting quantity would not be useful --- e.g., it wouldn't satisfy equation \ref{eq:rp}.}
}
--- we define them in terms of the corresponding players attempting to maximize the chance that the set is reached:
\begin{align}
\rp{-1}{\sigma}{H} & := \textbf{Pr}_\sigma \left[ \, H \textnormal{ reached} \mid \textnormal{P$1$ plays to reach $H$} \right] \\
& \, := \max_{\rho_1 \in \Sigma_1} \rp{}{\rho_1,\sigma_{-1}}{H} , \\
\rp{1}{\sigma}{H} & := \textbf{Pr}_\sigma \left[ \, H \textnormal{ reached} \mid \textnormal{both P$2$ and chance play to reach $H$} \right] \\
& \, := \max_{\rho_2 \in \Sigma_2}  \max_{\rho_c \in \Sigma_c} \rp{}{\sigma_1,\rho_1,\rho_c}{H}
    \label{eq:rp-p-set}
\end{align}
for player $1$, and similarly for player 2.
However, the resulting quantity only has desirable properties when considered over specific sets (e.g., infosets):

\begin{restatable}[Factorization of Infoset Reach Probabilities]{lemma}{infosetRP}\label{lem:infoset-reach}
For any $\sigma \in \Sigma$, $p\in \mc N$, and $I \in \mc I_p$, we have
\begin{align}
\rp{p}{\sigma}{I} \, & = \rp{p}{\sigma}{h} \textnormal{ for each } h \in I,
\label{eq:p's-rp}\\
\rp{-p}{\sigma}{I} & = \sum\nolimits_{h\in I} \rp{-p}{\sigma}{h} , \textnormal{ and}
\label{eq:p's-cf-rp}\\
\rp{}{\sigma}{I} \, & = \rp{p}{\sigma}{I} \rp{-p}{\sigma}{I} .
\label{eq:rp}
\end{align}
\end{restatable}

Note that the result is very much \textit{not} true in general --- e.g., when $I$ is an infoset \textit{of player 1}, the result doesn't hold for \textit{player 2's} probabilities.
Indeed, $\rp{2}{\sigma}{h}$ might be different for every $h\in I$, and the counterfactual reach probabilities $\rp{-2}{\sigma}{h}$, $h\in I$, might sum to a number larger than $1$.
As with histories, we say that a set $H \subset \mc H$ is \defword{reachable} (resp. \defword{unreachable}) under $\sigma$ when $\rp{}{\sigma}{H}$ is greater than zero (resp. equal to zero).
An infoset \textit{of $p$} is \defword{counterfactually reachable} (resp. unreachable) (by $p$)  when $\rp{-p}{\sigma}{I}$ is greater than (resp. equal to) zero.

\paragraph{Beliefs}

Lemma~\ref{lem:infoset-reach} allows us to introduce a robust notion of a belief over histories within an infoset.
When $I$ is \textit{reachable}, we define
\begin{align}\label{eq:cond-rp:informal}
\rp{}{\sigma}{h|I} := \textbf{Pr}_\sigma \left[ \, \textnormal{current history is $h$} \mid \textnormal{curr. infoset is $I$} \right]
:= \edit{ \frac{\rp{}{\sigma}{h}}{\rp{}{\sigma}{I}} \, }.
\end{align}
When $I\in \mc I_p$ is the current infoset of $p$, $\rp{}{\sigma}{\, \cdot \, |I}$ can be interpreted as $p$'s \defword{belief} about what the current state of the game is.
To extend this notion even to \textit{unreachable} infosets, we use an approach inspired by trembling hand equilibria, where strategies are injected with an infinitesimal amount of noise \cite{selten1974reexamination} (which makes $I$ reachable):
\begin{definition}[Generalized belief over an infoset]\label{def:infoset-belief}
For $I \in \mc I$ that is unreachable under $\sigma \in \Sigma$, the \defword{(generalized) belief over} $I$ is
$ \lim_{n\rightarrow \infty} \rp{}{(1-\frac{1}{n})\sigma + \frac{1}{n} \textnormal{unif}}{h|I}$, 
where $\textnormal{unif}$ denotes the uniformly random strategy.
\end{definition}

\noindent \textit{On the first approximation,} we can think of these beliefs as the beliefs we would obtain if all strategies that cause $I$ to be unreachable were replaced by the uniform strategy.
The limit serves to give priority to the histories that only require fewer deviations from $\sigma$ to become reachable.
While the choice of \textit{uniformly} random noise is an ad-hoc one, the resulting concept is nevertheless compatible with the notion of belief from \eqref{eq:cond-rp:informal}.

\begin{restatable}[Equivalent definitions of the infoset belief]{lemma}{equivalentInfosetDef}\label{lem:cond-rp}
Let $\sigma \in \Sigma$, $I \in \mc I_p$.
\edit{(1) The limit defining $\rp{}{\sigma}{h|I}$ always exists.}
(2) For cf. reachable $I$, $\rp{}{\sigma}{h|I} = \lim_n \rp{}{(1-\frac{1}{n})\sigma + \frac{1}{n} \textnormal{unif}}{h|I} = \frac{\rp{-p}{\sigma}{h}}{\rp{-p}{\sigma}{I}}$.
(3) For reachable $I$,
\begin{equation}
\rp{}{\sigma}{h|I}
= \lim_{n\rightarrow \infty} \rp{}{(1-\frac{1}{n})\sigma + \frac{1}{n} \textnormal{unif}}{h|I} 
= \frac{\rp{-p}{\sigma}{h}}{\rp{-p}{\sigma}{I}}
= \frac{\rp{}{\sigma}{h}}{\rp{}{\sigma}{I}}
.
\end{equation}
\end{restatable}

\noindent As a corollary, it is correct to write $\rp{}{\sigma}{h|I}$ without regard for $I$'s reachability:
When the infoset is reachable, the symbol can stand for $\frac{\rp{}{\sigma}{h}}{\rp{}{\sigma}{I}}$.
When it is only reachable counterfactually, it can still stand for $\frac{\rp{-p}{\sigma}{h}}{\rp{-p}{\sigma}{I}}$.
And when unreachable even counterfactually, we have to use the limit version.

\paragraph{Paths in the Infoset Tree}

The extension relation $\sqsubset$ (defined in Sec.~\ref{sec:EFGs}) can be translated from $\mc H$ to $\mc I_p$ (and $
\mc S$) by saying that $J \in \mc I_p$ is an extension of $I \in \mc I_p$, written as $J \sqsupset I$, if there are some histories $h \in J$, $g\in I$ s.t. $h$ extends $g$.
    (By perfect recall, this is equivalent to saying that every element of $J$ is an extension of some element of $I$.)
This turns $\mc I_p$ into a tree and allows us to talk about \textit{paths} and \textit{slices} through $\mc I_p$ \edit{(defined analogously to paragraph below eq. \ref{eq:rp-hist-product})} and distinguish between \textit{terminal} and \textit{non-terminal infosets} (i.e., $I\subset \mc Z$ and $I\subset \mc H \setminus \mc Z$).
For $I \in \mc I_p$,
\begin{equation}
    \edit{
        \ims{I} := \left\{ J \in \mc I_p \mid J \subset \{ ha \mid h\in I, a \in \mc A(I) \} \right\}
    }
\end{equation}
thus denotes the collection of \defword{immediate successors} of $I$ in $\mc I_p$.
For $\mc P(I)=p$,
\begin{equation}
    \edit{
        \ims{I,a} := \left\{ J \in \mc I_p \mid J \subset \{ ha \mid h\in I \} \right\}
    }
\end{equation}
denotes those immediate successors \edit{$J$ for which $a$ was chosen at $I$}. 

Since we already have the notion of belief $\rp{}{\sigma}{\, \cdot \, | I }$, we can provide both the intuitive definition of reach probability over a path in $\mc I_p$ and its calculation-friendly equivalent:

\begin{align}
\rp{}{\sigma}{I,J} \ & := \textbf{Pr}_\sigma \left[ \, J \textnormal{ reached} \mid \textnormal{current infoset is $I$}\right], \\
& \, = \sum\nolimits_{\edit{g}\in I} \rp{}{\sigma}{g|I}\sum\nolimits_{g\sqsubset h\in J} \rp{}{\sigma}{g,h} , \\
\rp{-p}{\sigma}{I,J} & := \textbf{Pr}_\sigma \left[ \, J \textnormal{ reached} \mid \textnormal{curr. infoset is $I$, $p$ plays to reach $J$}\right] \\
& \, = \sum\nolimits_{\edit{g}\in I} \rp{}{\sigma}{g|I}\sum\nolimits_{g\sqsubset h\in J} \rp{-p}{\sigma}{g,h} , \\
\rp{p}{\sigma}{I,J} \ & := \textbf{Pr}_\sigma \left[ \, J \textnormal{ reached} \mid \textnormal{curr. infoset is $I$, $-p$ play to reach $J$}\right] \\
& \, = \sum\nolimits_{\edit{g}\in I} \rp{}{\sigma}{g|I}\sum\nolimits_{g\sqsubset h\in J} \rp{p}{\sigma}{g,h} .
\end{align}
This guarantees that the reach probabilities of infosets can be decomposed, both in terms of the $p$-component and counterfactual component and in terms of splitting the path to any infoset into segments:

\begin{restatable}[Properties of Infoset Reach Probabilities]{lemma}{pathInfosetProperties}\label{lem:path-rp-props}
\edit{For any $\sigma \in \Sigma$ and $I \sqsubset J \sqsubset K$ in $\mc I_p$, we have:}
\begin{enumerate}[(1)]
    \item \edit{$\rp{}{\sigma}{I,J} = \lim_{n\rightarrow\infty} \frac{ \rp{}{\sigma^n}{J} }{ \rp{}{\sigma^n}{I} }$,
    $\rp{p}{\sigma}{I,J} = \lim_{n\rightarrow\infty} \frac{ \rp{p}{\sigma^n}{J} }{ \rp{p}{\sigma^n}{I} }$,} \\
    \edit{and
    $\rp{-p}{\sigma}{I,J} = \lim_{n\rightarrow\infty} \frac{ \rp{-p}{\sigma^n}{J} }{ \rp{-p}{\sigma^n}{I} }$
    (where $\sigma^n$ denotes $\frac{n-1}{n}\sigma + \frac{1}{n} \textnormal{unif}$).}
    \item \edit{$\rp{}{\sigma}{I,J} = \rp{p}{\sigma}{I,J} \rp{-p}{\sigma}{I,J}$.}
    \item \edit{$\rp{}{\sigma}{I,K} = \rp{}{\sigma}{I,J} \rp{}{\sigma}{J,K}$.}
\end{enumerate}
\end{restatable}

\noindent
\edit{As a result, the tree $\mc I_p$ satisfies the assumptions of Lemma~\ref{lem:tree-characterization}.}

\paragraph{Expected Utilities of Information Sets}

We now have all tools required to define the expected utilities of infosets and show they behave analogously to $\hv{p}{\sigma}{h}$:

\begin{definition}[Infoset value]\label{def:infoet-values}
Let $I\in \mc I_p$ and $\sigma \in \Sigma$.
The ($p$'s) \defword{value of} $I$ \defword{under} $\sigma$ is defined as
\begin{equation}
\Iv{}{\sigma}{I} := \Iv{p}{\sigma}{I} := \sum\nolimits_{h\in I} \rp{}{\sigma}{h|I} \hv{p}{\sigma}{h} .
\end{equation}
For non-terminal $I$ and $a \in \mc A_p(I)$, ($p$'s) value of taking $a$ at $I$ is
\begin{equation}
\Iav{}{\sigma}{I}{a}
:= \Iav{p}{\sigma}{I}{a}
 \edit{
 : = \sum\nolimits_{h\in I} \rp{}{\sigma}{h|I} \hav{p}{\sigma}{h}{a}
 }
 .
\end{equation}
\end{definition}

By the following lemma, the action values can be derived from infoset values.
As a result, we will mostly focus our analysis on the latter.
\edit{
\begin{restatable}{lemma}{actionVals}\label{lem:Q-vals}
For any $\sigma\in\Sigma$, $I\in\mc I_p$ s.t. $\mc P(I)=p$, and $a\in \mc A(I)$, we have
\begin{equation}
    \Iav{}{\sigma}{I}{a} = \sum\nolimits_{J \in \ims{I,a} } \rp{-p}{\sigma}{I,J} \Iv{}{\sigma}{J} .
\end{equation}
\end{restatable}
}



\noindent
\edit{Since $\Iv{}{\sigma}{I} = \sum_{h\in I} \rp{}{\sigma}{h|I} \hv{p}{\sigma}{h} = \sum_{h\in I} \frac{\rp{-p}{\sigma}{h}}{\rp{-p}{\sigma}{I}} \hv{p}{\sigma}{h}$ for counterfactually reachable $I$,}
the infoset values from Definition~\ref{def:infoet-values}
can be viewed as a generalization of the notion of counterfactual utilities from \cite{CFR} \edit{(whose definition does not work for counterfactually unreachable $I$)}.

As advertised, Theorem~\ref{thm:infoset-v-characterization} demonstrates that infoset values can be defined in several equivalent ways.
Specifically, it shows that
\edit{they can be viewed as weighted sums of history-values (1)
or as expected utilities calculated over the infoset tree (2-4).
The latter can be viewed either as the expectation over terminal states (2), over immediate successors in the infoset tree (3, 3'),
or over any slice through the infoset tree (below the given infoset) (4).}

\begin{restatable}[Characterization of $\IvNP{}{\sigma}$]{theorem}{infosetValueCharacterization}\label{thm:infoset-v-characterization}
Suppose that terminal infosets are always singleton.
Then for any $p\in \mc N$ and $\sigma \in \Sigma$,
we have $\Iv{p}{\sigma}{\textnormal{root}} = u_p(\sigma)$ and
the functions $\IvNP{}{\sigma} : \mc I_p \to \R$ and $\Iav{}{\sigma}{I}{a} := \sum_{\ims{I,a} } \rp{-p}{\sigma}{I,J} \Iv{}{\sigma}{J} $ have the following properties:
\begin{itemize}
\item[(1)] \edit{$\Iv{}{\sigma}{I} = \sum_{h\in I} \rp{}{\sigma}{h|I} \hv{p}{\sigma}{h} $.}
\item[(2)] \edit{$\Iv{}{\sigma}{I} = \sum_{z \in \mc Z} \rp{}{\sigma}{I,\{z\}} u_p(z) $.}
\item[(3)] $\IvNP{p}{\sigma} = u_p$ on $\mc Z$ and for non-terminal $I \in \mc I_p$, we have
    \begin{equation}
        \Iv{}{\sigma}{I} = \sum_{\edit{J\in}\ims{I}} \rp{}{\sigma}{I,J} \Iv{}{\sigma}{J} .
    \end{equation}
\item[(3')] $\IvNP{p}{\sigma} = u_p$ on $\mc Z$ and for non-terminal $I$, we have
	\begin{align*}
		\Iv{}{\sigma}{I} = 
		\begin{cases}
			\sum_{a \in \mc A_p(I)} \sigma_p(I,a)\Iav{}{\sigma}{I}{a} 		& \textnormal{ when $p$ acts in $I$} \\
			\sum_{J \in \ims{I}} \rp{-p}{\sigma}{I,J} \edit{\Iv{}{\sigma}{J}}	& \textnormal{ when $p$ doesn't act in $I$} .
		\end{cases}
	\end{align*}
\item[(4)] $\IvNP{p}{\sigma} = u_p$ on $\mc Z$ and for non-terminal $I \in \mc I_p$ \edit{and any slice $\mc L$ through $\mc I_p$ below $I$, we have 
$\Iv{}{\sigma}{I} = \sum\nolimits_{J \in \mc L } \rp{}{\sigma}{I,J} \Iv{}{\sigma}{J}$
.}
\end{itemize}
Moreover, each of these conditions can be used as an equivalent definition of $\IvNP{p}{\sigma}$ (i.e., it automatically implies all the other properties).
\end{restatable}

While the result assumes that terminal infosets are always singleton, this assumption is primarily a cosmetic one, to highlight the connection to Lemma~\ref{lem:hist-value:characterization}.
Indeed, if terminal infosets were of the more general form $Z\subset \mc Z$, we could equally well anchor $\IvNP{}{\sigma}$ using the function $U^\sigma_p(Z) := \sum_{z\in Z} \rp{}{\sigma}{h|Z} u_p(z)$.\footnote{Historically, many implementations of algorithms in EFGs ``ran on $\mc H$''. The results presented in this section can be particularly relevant when attempting to instead use structures where infosets (or even public states \cite{ReBeL}) play a prominent role. Since the infoset trees $\mc I_p$ are much smaller than $\mc H$, these have the potential to be more effective (particularly if the specific game allows for an efficient computation of $U^\sigma_p(Z)$).}

A useful corollary of Theorem~\ref{thm:infoset-v-characterization} is that \edit{to compute} $\Iv{p}{\sigma}{I}$, \edit{we only need to specify} $p$'s strategy below $I$ (but neither above it nor below its siblings) and $-p$'s strategy above and below the public state that contains $I$ (but not below the public state's siblings).

Finally, \cite{ReBeL} recently showed that $\Iv{}{\sigma}{I}$ can also be defined as a partial derivative (or rather, more precisely, a supergradient) of the expected utility of the public state which contains $I$ with respect to the reach probability of $I$.
This result thus presents another useful view of values functions, one that is orthogonal to the direction taken here.
We will discuss it further in Remark~\ref{rem:PBS-values}, once we have access to more related concepts.

\subsection{Counterfactual Values and CFR}\label{sec:sub:cfvs}

The tools developed in this section also allow us to summarize the properties of the \textit{counterfactual} values used in the CFR literature \cite{CFR}.
Since the algorithm we use in our experimental section is an extension of CFR, we use this opportunity to describe CFR's standard version.

\paragraph{Counterfactual Values}

\defword{Counterfactual values}, and the corresponding action-values, are defined as the non-counterfactual values weighted by the counterfactual reach probability:
\begin{align}
\hvCf{p}{\sigma}{h} & := \ \rp{-p}{\sigma}{h} \hv{p}{\sigma}{h} , \label{eq:cf-value}\\
\havCf{p}{\sigma}{h}{a} & := \ \hvCf{p}{\sigma}{ha}, \\
\IvCf{}{\sigma}{I} \ & := \ \IvCf{p,}{\sigma}{I} \ \ \ 
    := \sum\nolimits_{h\in I} \hvCf{p}{\sigma}{h} , \\
\IavCf{}{\sigma}{I}{a} & := \ \IavCf{p,}{\sigma}{I}{a}
    := \sum\nolimits_{h\in I} \havCf{p}{\sigma}{h}{a}
    .
\end{align}

\noindent The main advantage of the counterfactual values over the ``standard'' ones is that they never run into problems with unreachable states --- either $h$ (or $I$) is counterfactually reachable, and then its counterfactual value follows from the well-defined formula \eqref{eq:cf-value}, or it is cf. unreachable and its cf. value is $0$.
The downside is that these values have no clear intuitive interpretation (unlike $\Iv{}{\sigma}{I}$ being the expected utility conditional on being at $I$).

\edit{The following result highlights the important properties of $\IvCfNP{}{\sigma}$.
First, $\IvCf{p}{\sigma}{\textnormal{root}}$ coincides with $u_p(\sigma)$ --- in particular, if neither player can improve their $\IvCf{p}{\sigma}{\textnormal{root}}$, $\sigma$ is an equilibrium (1).
Second, $\IvCfNP{}{\sigma}$ can be obtained from $\IvNP{}{\sigma}$ using (2).
This implies (3) and (4): once we know some player's counterfactual values in leaves, backpropagating them only requires the knowledge of \textit{that player's} reach probabilities.
In particular, computing the cf. values of leaves is more difficult than computing their normal values but backpropagating these values is much easier.}


\begin{restatable}[Properties of $\IvCfNP{}{\sigma}$]{theorem}{CFVproperties}\label{thm:infoset-cfv-properties}
\edit{For any $\sigma \in \Sigma$, $p\in \mc N$, and non-terminal $I \in \mc I_p$, we have:
\begin{enumerate}[(1)]
    \item $\IvCf{p,}{\sigma}{\textnormal{root}} = u_p(\sigma)$.
    \item $\IvCf{}{\sigma}{I} = \rp{-p}{\sigma}{I} \Iv{}{\sigma}{I}$.
    \item For any slice through $\mc I_p$ below $I$, we have
        $\IvCf{}{\sigma}{I} = \sum_{J\in \mc L} \rp{p}{\sigma}{I,J} \IvCf{}{\sigma}{J}$.
    \item (a) For terminal $Z \in \mc I_p$, $\IvCf{}{\sigma}{Z} = \sum_{z\in Z} \rp{-p}{\sigma}{z} u_p(z)$.\\
        (b) When $\mc P(I)=p$, we have
            \begin{align*}
            \IvCf{}{\sigma}{I}
            = \sum_{a \in \mc A(I)} \sigma_p(I,a) \IavCf{}{\sigma}{I}{a}
            = \sum_{a \in \mc A(I)} \sigma_p(I,a) \sum_{J\in \ims{I,a}} \IvCf{}{\sigma}{J}
            .
            \end{align*}
            \\
        (c) When $\mc P(I) \neq p$, we have $\IvCf{}{\sigma}{I} = \sum_{J\in \ims{I}} \IvCf{}{\sigma}{J}$.
\end{enumerate}
}
\end{restatable}

\paragraph{Counterfactual Regret Minimization}\label{sec:cfr}

\edit{Informally speaking,} the counterfactual regret minimization algorithm (\defword{CFR}) \edit{uses} counterfactual values to compute an approximate equilibrium by iteratively traversing the game tree and minimizing regret at each information set \cite{CFR}.
%
\edit{Formally, the \defword{immediate counterfactual regret} is the difference between counterfactual value of an infoset and the highest cf. value achievable by changing the strategy only at that infoset:
\begin{equation}
R^\sigma_{\textnormal{cf}}(I)
:=
\max_{a\in\mc A(I)} R^\sigma_{\textnormal{cf}}(I,a)
:=
\max_{a\in\mc A(I)} {\ Q^{\sigma}_{\textnormal{cf}}(I,a) - V^\sigma_{\textnormal{cf}}(I)} .
\end{equation}
}





\noindent
\edit{CFR starts with uniformly random strategy $\sigma^1 $ and updates it using the \defword{regret matching} update rule with respect to immediate cf. regrets:
}
\begin{align}\label{eq:regret-matching}
{\sigma^{t+1}}(I,a) = 
\begin{cases}
	\frac{
	    \sum^{\edit{t}}_{\edit{k}=1}
	        R^{\sigma^{\edit{k}}}_{\edit{\textnormal{cf}}}(I,a)
        }{
            \sum^{\edit{t}}_{\edit{k}=1} \sum_{a\in \mc A(I)} R^{\sigma^{\edit{k}}}_{\edit{\textnormal{cf}}}(I,a)
        } 
	    & \textnormal{if} \ \  \sum^{\edit{t}}_{\edit{k}=1} \sum_{a\in \mc A(I)} R^{\sigma^{\edit{k}}}_{\edit{\textnormal{cf}}}(I,a) > 0 \edit{,}\\
	\frac{1}{\abs{A(I)}}
	    & \textnormal{otherwise.}
\end{cases}
\end{align}

\noindent
\edit{This causes the \defword{average strategy} $\bar \sigma$ to converge to an equilibrium \cite{CFR}:
\begin{equation}
    \bar \sigma (I,a) = \sum^T_{t=1} \frac{ P^{\sigma^t}_{\edit{p}}(I)}{\sum^T_{k=1} P^{\sigma^k}_{\edit{p}}(I)} \sigma^t(I,a) .
\end{equation}
}

\edit{While CFR typically employs regret matching, it is not the only valid option.
One alternative is the Exp3 algorithm \cite{auer2002nonstochastic}, which always produces fully-mixed strategies.
Adopting such algorithm is tempting since it makes the whole $\mc H$ is reachable and simplifies many theoretical considerations.
However, this benefit comes with two limitations.
First, it goes away the moment we artificially restrict the action space (as is routinely done, for example, in poker \cite{DeepStack}).
Second, practical implementations of CFR ignore the (counterfactually) unreachable parts of the game tree.
    Using fully-mixed strategies will therefore increase the time needed to run the algorithm.
}

\section{Theory}\label{sec:theory}

In this section, we present the theoretical contributions made by this paper.
Because of the section's length, we first give a brief summary of its contents.

We start by introducing depth-limited versions of many notions such as exploitability, best-response, and Nash equilibria (Section~\ref{sec:trunks}).
Afterwards, we define the abstract notion of depth-limited games --- these can be understood as game trees where terminal values depend not just on the leaf in which the game ends, but also on the strategy used to get to the leaf (Section~\ref{sec:sub:dl_games}).
To utilize this notion, we need to identify depth-limited games that are useful for finding equilibria of non-depth-limited games.
In Section~\ref{sec:enabling-dl}, we thus provide a sufficient condition for solutions of a depth-limited game to coincide with Nash equilibria of the original game (or, more precisely, with strategies that can be extended into such Nash equilibria).
Unfortunately, we also see that this criterion alone does not imply the existence of \textit{practical} methods for solving the depth-limited game.

To allow for such practical methods, we define optimal value functions as those that correspond to values of optimal extensions of the trunk strategy (Section~\ref{sec:opt_vfs}).
We make a distinction between (a) reachably-optimal, (b) counterfactually-optimal, and (c) universally-optimal value functions, based on whether the corresponding extension of the trunk-strategy performs well (a) in situations that arise under the trunk strategy, (b) in those that could arise if one player deviated from the trunk strategy, or (c) in all situations.
We look at \edit{the first two types} of optimality in detail, showing how to compute the corresponding value functions and discussing which depth-limited algorithms are enabled by them --- for a summary, see Table~\ref{tab:opt_properties-summary}.
In particular, we prove that counterfactually-optimal value functions are both sufficient and necessary to make depth-limited CFR work well (Proposition~\ref{prop:enable-DL-CFR}).

\begin{table}[tbh]
    \begin{center}
    \begin{tabular}{ c || c c c } 
    Optimality & reachable & counterfactual & universal \\
    \hline\hline
    Described in & Section\,\ref{sec:opt_vfs:reach} & Section\,\ref{sec:opt_vfs:cf} & Section\,\ref{sec:opt_vfs:univ} \\
    \hline
    Optimal at & $\rp{}{\trunkS}{I}>0$ & $\rp{-p}{\trunkS}{I}>0$ & all $I$\\
    \hline
    Suff. stats & joint range & range & \edit{not discussed}\\
    \hline
    How to & value- & CFR; post-processing & \edit{not discussed} \\
    compute? & solving & via best-response &  \\
    \hline
    Algorithms & IS-MCTS, & CFR, best response, & minimax, \\
    enabled     & subgame value & poss. fictitious play      & possibly others
    \end{tabular}
    \end{center}
    \caption{A summary of properties of different types of optimal value functions.}
    \label{tab:opt_properties-summary}
\end{table}

In Section~\ref{sec:noam}, we relax the notion of optimality to only require being optimal w.r.t. a subset of all strategies and observe that this still enables finding high-quality strategies.
Moreover, we observe that \cite{brown2018depth}'s multi-valued states can be understood as a specific instance of depth-limited solving that relies on this type of optimal value functions.
We argue that this connection deserves further attention.

%

Finally, we show how to represent value functions more compactly and prove that public states provide the minimum context that is still informative enough to enable the computation of optimal values (Section~\ref{sec:compact-representations}).
We also note that optimal values are not uniquely defined, which could prove important (and potentially inconvenient) when approximating them by neural nets (Section~\ref{sec:uniqueness}).
However, we do not currently see this topic as priority since we have not encountered any difficulties in practice (Section~\ref{sec:empirical}). 
    \subsection{Depth-Limited Methods}\label{sec:DL_games}

Depth-limited methods are a standard tool in perfect-information games\edit{, and several such methods exist in imperfect information games as well \cite{brown2018depth,CFR-D,DeepStack,wiggers2016structure}.
Using the background given in Section~\ref{sec:background:values}, we now present a framework that allows for unifying these approaches.}

We briefly talk about general partial strategies that are defined only on some subset of the game tree and discuss how to measure their quality. However, we quickly move to the specific case where the subset of the game tree is a so-called trunk --- i.e., a connected subset that contains the tree’s root. We define non-abstract value functions as mappings that take a fully-defined strategy and a node at the bottom of the trunk and return the nodes expected value under the given strategy (Section~\ref{sec:trunks}).

To abstract away the bottom part of the game, we define abstract value functions as mappings that output a real number for each trunk strategy and a node at the bottom of the trunk. A depth limited game is then defined analogously to a standard game, except that values of terminal nodes (at the bottom of the trunk) are determined by an abstract value function instead of a utility function. Unlike in \textit{standard} imperfect-information games and depth-limited \textit{perfect-information} games, the desirability of an outcome in a depth-limited imperfect-information game thus depends not just on which terminal node is generated, but also on which strategy was used to do so (Section~\ref{sec:sub:dl_games}).

In any depth-limited game, we can look for optimal strategies, best responses, exploitability, etc.
However, all these concepts are defined with respect to the given abstract value function --- if the depth-limited game is to be useful, the value function will need to capture information about the optimal play in the part of the game that it is intended to be abstracted away. But before analyzing optimal value functions, we need to be more specific about what it means for a depth-limited game to be useful.
Initially, it might seem that all that is needed is for strategies “optimal with respect to the abstract value function” to coincide with Nash equilibria of the original game.
Unfortunately, this alone isn't enough because there are value functions that satisfy this criterion while not admitting any realistic method for finding the optimal strategy.
(A simple \textit{artificial} example is a value function that a gives a binary penalty to any player whose strategy is exploitable\footnote{Formally, we set $\vv(h,\trunkS):=0$ when $\expl{1}{}{\trunkS_1}=\expl{2}{}{\trunkS_2}=0$, $\vv(h,\trunkS):= -1$ when $\expl{1}{}{\trunkS_1}>0$ and $\expl{2}{}{\trunkS_{2}}=0$, $\vv(h,\trunkS):= 1$ when $\expl{2}{}{\trunkS_2}>0$ and $\expl{1}{}{\trunkS_{1}}=0$, and $\vv(h,\trunkS):=0$ when $\expl{1}{}{\trunkS_1}, \expl{2}{}{\trunkS_2} > 0$.} --- the only way to find an equilibrium here is to randomly stumble upon it.
However, we also later provide a realistic example.)
Instead, we can talk about the usefulness in the context of specific calculations that we might wish to perform on the game: for example, does the given value function enable a depth limited version of algorithms like CFR or fictitious play? (That is, can we straightforwardly modify those algorithms to work in the depth-limited game and will their outputs converge to Nash equilibria?)
Does the value function enable finding the game value of subgames?
Or perhaps our interest is something other than finding Nash equilibria.
For example, to allow for exploiting specific opponents, it is important that the value function enables an analogue of the best-response operator.
Since the purpose of each calculation is different, the “enabling” concept is hard to capture formally. However, it can be formalized in all specific cases (Section~\ref{sec:enabling-dl})
 
\edit{At a} high level, this section shows how to combine depth-limited games with \textit{arbitrary} value functions, some of which might be of no practical use, while Section~\ref{sec:opt_vfs} analyzes \textit{optimal} value functions, giving depth-limited algorithms guarantees analogous to their original versions.

\subsubsection{Trunks and Partial Strategies}\label{sec:trunks}

Before motivating and defining depth-limited games, we introduce the notion of partial strategies, trunk strategies, and their exploitability.
If $p$'s (full) strategy is function defined on $\mc I_p$, their \textit{partial} strategy is defined on a subset of $\mc I_p$:

\begin{definition}[Partial strategy]
    A \defword{partial strategy} of player $p$ is a mapping that assigns a probability distribution $\sigma_p(I) \in \Delta(\mc A(I))$ to each infoset $I$ from some \textit{subset} $\mc J_p \subset \mc I_p$.
    \edit{Let $H \subset \mc H$ be a set closed under membership within infosets.\footnote{\edit{That is, we have $(\forall p \in \mc N)(\forall I \in \mc I_p) : I \cap H \neq \emptyset \implies I \subset H.$}.}}
    $\Sigma_p^{H}$ denotes the set of all $p$'s partial strategies defined on $\mc J_p = \left\{ I \in \mc I_p \mid I \subset H \right\}$.
    \edit{We set} $\Sigma^H := \Sigma^H_1 \times \Sigma^H_2$.
\end{definition}

When a partial strategy $\rho_p$ and a (possibly partial) strategy $\sigma_p$ satisfy (i) the domain of $\rho_p$ is a subset of the domain of $\sigma_p$ and (ii) $\sigma_p = \rho_p$ on the domain of $\rho_p$, we say that $\sigma_p$ \defword{extends} $\rho_p$ and write $\rho_p \subset \sigma_p$.\footnote{Recall that since partial strategies are formally subsets of $\mc I_p \times \Delta \mc A_p$, \edit{$\sigma_p$ is indeed an extension of $\rho_p$ iff $\sigma_p \supset \rho_p$.}}

\begin{definition}[Trunk]\label{def:trunk}
    A set $\trunk \subset \mc H \edit{ \setminus \mc Z }$ is called a \defword{trunk}\footnotemark if it is closed under parent nodes and \edit{membership within} public states.\footnote{The metaphor is that if $\mc H$ is a tree, then $\trunk$ is its trunk, the subgames rooted at some $S\subset \mc Z^\trunk$ are its branches, and terminal histories are its leaves. While trunks can and do come in various shapes, we typically imagine trunks \edit{of the form $\mc T = \{ h \in \mc H \setminus \mc Z \mid \textnormal{length}(h) \leq k \} $ for some $k$}. (Assuming that game-theoretical trees grow downwards, this is equivalent to $\mc T := $ ``the upper half of $\mc H$ that has been sliced in two horizontally''.)}
    By $\mc Z^\trunk := \{ ha \mid h\in \trunk,\, a\in \mc A(h), \, ha \notin \trunk \}$, we denote the leaves of $\trunk$.\footnote{By Definition~\ref{def:trunk}, leaves of a trunk $\trunk$ are \textit{just below} $\trunk$. For a trunk strategy profile $\trunkS$, the leaves of $\trunk$ are thus the first nodes where \edit{$\sigma^{\mc T}$} becomes undefined.}
	The elements of $\Sigma_p^\trunk$ are called \defword{trunk strategies}.
\end{definition}
\footnotetext{\edit{This definition formalizes the notion of trunk used in \cite{CFR-D}.}}

Note that the leaves of any trunk are a slice through $\mc H$ (in the sense of the definition below \edit{eq.~\ref{eq:rp-hist-product}}), allowing us to apply Lemma~\ref{lem:hist-value:characterization} and Theorem~\ref{thm:infoset-v-characterization}. In particular, the value of any history (or infoset) in $\trunk$ can be calculated by summing up the (appropriately weighted) values of its descendants in $\mc Z^\trunk$.

As in the full-game setting \edit{(Sec.~\ref{sec:EFGs})}, we can use exploitability to quantify the magnitude of mistakes that a partial strategy makes.
Exploitability of a partial strategy is thus defined as the exploitability of the strategy's least-exploitable\footnote{This is in line with the intuition that a strategy defined nowhere hasn't made any mistakes yet and has, therefore, exploitability zero.} extension to the full game $G$.
To simplify the discussion, we will mostly focus on trunk strategies.
However, a number of concepts we study (such as this one) could also be applied to the more general notion of partial strategies.

\begin{definition}[Trunk exploitability]
\label{def:trunk-exploitability}
The \defword{exploitability} of a trunk strategy $\sigma_p^\trunk$ is defined as
\begin{align*}
\expl{p}{}{\trunkS_p} := & \ 
\min \left\{ \expl{p}{}{\sigma_p} \mid 
\sigma_p^\trunk \subset \sigma_p \in \Sigma_p \right \} .
\label{eq:expl}
\end{align*}
\end{definition}

We can measure the quality of trunk-strategy \textit{profiles} $\trunkS = (\sigma_1^\trunk, \sigma_2^\trunk)$  via the average of $\expl{p}{}{\sigma_p^\trunk}$ for $p=1,2$.
We say that $\trunkS$ is an $\epsilon$-\defword{Nash equilibrium in the trunk}, denoted as $\trunkS \in \epsilon\textnormal{-NE}(G)|_\trunk$, if it can be extended into an $\epsilon$-NE of $G$.
Since Nash equilibria in $G$ are precisely the pairs of unexploitable strategies, Definition~\ref{def:trunk-exploitability} immediately implies that equilibria in the trunk are precisely the restrictions of full-game equilibria to $\trunk$:

\begin{lemma}\label{lem:trunk-expl}
A partial strategy has $\expl{p}{}{\sigma_p^\trunk} = 0$ if and only if it can be extended into a Nash equilibrium in $G$.
\end{lemma}

Definition~\ref{def:trunk-exploitability} also suggests that the exploitability of a partial strategy can be obtained by comparing the original value of the game $G$ with the value of the game $G(\sigma_p^\trunk)$ defined by ``forcing'' $p$ to play $\sigma_p^\trunk$ in $\trunk$.\footnote{Formally, the $G(\sigma_p^\trunk)$ is obtained by turning the decision points in $\sigma_p^\trunk$'s domain into chance nodes with probabilities $\sigma_p^\trunk(I)$. \edit{Since we can always convert the chance strategies in $\trunk$ back to players' strategies, we will sometimes abuse the notation slightly and treat strategies in $G(\sigma_p^\trunk)$ as extensions of $\sigma_p^\trunk$.}}
The following formula thus gives a practical recipe for computing trunk exploitability:

\begin{proposition}[Computing trunk exploitability]\label{prop:trunk_expl}
Exploitability of a trunk strategy can be computed as
\begin{align*}
\expl{p}{}{\sigma_p^\trunk} = & \ 
\gv{p}{G} - \gv{p}{G(\trunkS_p)} .
\end{align*}
\end{proposition}

\begin{proof}
For $\Sigma_p \ni \sigma_p \supset \sigma_p^\trunk$, we have $\expl{p}{}{\sigma_p} = \gv{p}{G} - \edit{\min}_{\sigma_{-p}} u_p(\sigma_p,\sigma_{-p})$.
Taking the \edit{minimum} over $\sigma_p$, we get
\begin{equation*}
\expl{p}{}{\sigma_p^\trunk} =
\edit{ \min_{\sigma_p \supset \trunkS_p} \expl{p}{}{\sigma_p} = \ }
\gv{p}{G} - \edit{\max_{\sigma_p \supset \trunkS_p}} \ \edit{\min_{\sigma_{-p} \supset \trunkS_{-p}}} u_p(\sigma_p,\sigma_{-p}) ,
\end{equation*}
where the last term is equal to $\gv{p}{G(\sigma_p^\trunk)}$.
\end{proof}

\subsubsection{Formal Definition of Depth-Limited Games}\label{sec:sub:dl_games}

In the previous section on partial strategies, we have seen how to measure the quality of a trunk strategy in terms of how well would the strategy perform when extended into the full game.
However, the trunk itself --- without the context of the full game --- is just a set of histories, \textit{not an optimization problem} that can be solved.
In this section, we will explain how to turn the trunk into a depth-limited \textit{game} --- i.e., something that \textit{can} be solved --- and discuss how its solutions relate to Nash equilibria in the trunk.

The central notion is that of a value function.
We will be particularly interested in value functions that correspond to history-values under some strategies in the full game.
However, when approximating such value functions (for example by a neural network), we might get functions which no longer correspond to any specific set of strategies, which is where the notion of \textit{abstract} value functions become useful:

\begin{definition}[Value function]\label{def:value_function}
Given a trunk $\trunk$, an \defword{(abstract) value function} for $\trunk$ is any function $ \vv : {\mathcal Z}^\trunk \times \Sigma^\trunk \to \R$.
\edit{We denote its values as $\vf{}{\trunkS}{h}$.}

When an extension $\sigma \in \Sigma$ of $\trunkS \in \Sigma^\trunk$ satisfies $\vf{}{\trunkS}{h} = \hv{1}{\sigma}{h}$ for every $h\in \mc Z^\trunk$, we say that $\vv^{\trunkS}$ \defword{corresponds to} $\sigma$.
\end{definition}

In practice, value functions will typically not require the full trunk-strategy as context --- for example, in perfect-information games, they require no context at all.
Nor is it strictly necessary that the first component of a value-function's input is a \textit{history} rather than an infoset.
However, this basic case considered above is both simplest and most general.
We thus present the main results in this framework and return to the topic of more compact representations of $\vv$ in Section~\ref{sec:compact-representations}.

\begin{figure}
    \centering
    \includegraphics[width=0.6\textwidth]{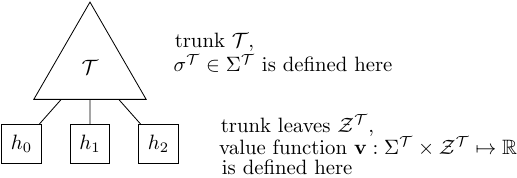}
    \caption{A depth-limited version of some game.
    A trunk strategy is defined in the trunk $\trunk$.
    Trunk leaves $\mc Z^{\trunk}$ are \textit{just below} the trunk, i.e., \textit{not} a part of it.
    For a fixed trunk strategy $\trunkS$, the value function $\vf{}{\trunkS}{\cdot}$ assigns a real number to each history in $\mc Z^{\trunk}$.
    }
    \label{fig:trunkvf}
\end{figure}

\begin{definition}[Depth-limited game]\label{def:dl_game}
For a two-player zero-sum EFG, a depth-limited game is formed by a pair $(\trunk, \vv)$, where $\trunk$ is a trunk and $\vv$ \edit{an abstract value function}.
\end{definition}

Figure~\ref{fig:trunkvf} shows an example of a depth-limited game.
%
%
A depth-limited game is \textit{not} an instance of an EFG, since $\vv$ behaves differently than utilities do (\edit{i.e., the value of a leaf additionally depends on the strategy used to reach that leaf}).
However, all the traditional definitions can be re-used in $(\trunk, \vv)$:

\begin{notation}[Depth-limited game concepts]\label{not:DL-concepts}
Let $(\trunk, \vv)$ be a depth-limited game.
\emph{Expected utility in $(\trunk, \vv)$} is defined as
\begin{equation}\label{eq:DL-u}
\DLu{p}{\trunkS} := \sum_{h \in \mathcal Z^\trunk} \rp{}{\trunkS}{h} \vf{p}{\trunkS}{h} ,
\end{equation}
where $\vfNP{1}{} := \vv$ and $\vfNP{2}{} := - \vv$.
A trunk strategy $\sigma_p^\trunk \in \Sigma_p^\trunk$ is a \emph{best response in $(\trunk, \vv)$} to $\sigma_{-p}^\trunk \in \Sigma_{-p}^\trunk$ if it maximizes the expected utility in $(\trunk, \vv)$, i.e., if $\sigma_p^\trunk \in \argmax_{\tilde \sigma_p^\trunk} \, \DLu{p}{ \tilde \sigma_p^\trunk , \sigma_{-p}^\trunk}$.
A trunk strategy profile is a \emph{solution\footnote{Recall that ``Nash equilibria in the trunk'', $\textnormal{NE}(G)|_\trunk$, are defined as restrictions of full-game NE to $\trunk$. To avoid a possible confusion between $\textnormal{NE}(G)|_\trunk$ and ``Nash equilibria'' of $(\mc T,\vv)$, we will reserve the term \textit{solution} to depth-limited games and use \textit{Nash equilibrium} to talk about restrictions of $\textnormal{NE}(G)$.}
of $(\trunk, \vv)$} if each $\sigma_p^\trunk$ is a best response in $(\trunk, \vv)$ to $\sigma_{-p}^\trunk$.
(Similarly, we could talk about $\epsilon$-approximation solutions.)
The (max-min) value of $(\trunk, \vv)$ for player $p$ is defined as $\gv{p}{\trunk,\vv} := \max_{\trunkS_p} \min_{\trunkS_{-p}} \DLu{p}{\trunkS}$ (which stands for ``game value'', from reasons that will become clear later).
\emph{Exploitability in $(\trunk, \vv)$} of $\sigma_p^\trunk$  is defined as
$\expl{p}{v}{\sigma_p^\trunk}
:= \gv{p}{(\trunk,\vv}\edit{)} - \min_{\sigma_{-p}^\trunk} \DLu{p}{\sigma_p^\trunk,\sigma_{-p}^\trunk}$.
Finally, we use the upper index $\vv$ to refer to $(\trunk,\vv)$-variants of the various notions of value introduced in Section~\ref{sec:background:values}; for example, we the value of a history in $(\trunk,\vv)$ is $\DLhv{p}{\trunkS}{h} := \sum_{z \in \mc Z^\trunk} \rp{}{\trunkS}{h,z}\vf{p}{\trunkS}{z}$.
\end{notation}

\medskip

Until we have a reason to believe otherwise, we should not automatically assume that these ``in $(\trunk,\vv)$'' variants of classical concepts behave as their traditional counterparts.
Indeed, an arbitrary function $f : \Sigma^\trunk \to \R$ can be turned into an abstract value function via $\vv_f(h,\trunkS) := f(\trunkS)$, which trivially makes the expected value of any strategy in $(\mc T,\vv_f)$ equal to $u^{\vv_f}_1(\trunkS) = f(\trunkS)$.
This implies, for example, that not all $\vv$-s will satisfy $\gv{2}{\trunk,\vv} =  - \gv{1}{\trunk,\vv}$ (since not all functions satisfy $\max \min f = \min \max f$).
Nevertheless, we will later show that ``non-abstract'' value functions (and possibly others) mostly behave as we might intuitively expect.

\subsubsection{Enabling Depth-Limited Algorithms}\label{sec:enabling-dl}

\edit{Depth-limited games and value functions are ultimately only useful to the extent that they allow us gain insights about the original game.
One might intuitively expect that a value function $\vv$ is useful if and only if} the solutions of $(\trunk, \vv)$ coincide with the equilibria of the full game (Definition~\ref{def:equil_preservation} below).
\edit{However, we will see (in Example~\ref{ex:Oliehoek}) that a depth-limited game can be of essentially no practical use despite satisfying this criterion.
This suggests that rather than asking \textit{whether} $\vv$ is useful or not, we need a more nuanced approach:
The algorithms for \textit{full} EFGs naturally come with corresponding depth-limited variants.
However, a depth-limited variant of an algorithm might work as intended with some value functions but not others.
As a result, we propose to view a value function's usefulness in terms of the class of algorithms that it enables, i.e., that work as intended when coupled with it.
We now describe these ideas in more detail and illustrate them on depth-limited minimax and CFR.
}


\edit{We start with the naive approach of checking for a correspondence between the solutions of $(\trunk, \vv)$ and the equilibria of the full game.}
Recall that any trunk $\trunk$ corresponds to two types of solution concepts:
First, since $\trunk$ is a part of the full game $G$, we can look at the trunk-equilibria $\textnormal{NE}(G)|_{\trunk}$ given by Definition~\ref{def:trunk-exploitability} --- i.e., the strategies that can be extended to the full game in a non-exploitable manner.
Second, endowing $\trunk$ with a value function $\vv$ turns $(\trunk,\vv)$ into a depth-limited game whose solutions (described in Notation~\ref{not:DL-concepts}) are the strategies optimal w.r.t. $\vv$.
The \edit{latter} can be computed solely based on $\vv$ (while possibly being of no use at all), while the \edit{former} are the strategies that we are after.
Intuitively, one might thus expect that all we need for $(\trunk,\vv)$ to be useful is for these two notions to coincide:

\begin{definition}[Equilibria preservation]\label{def:equil_preservation}
$
(\trunk,\vv)$ \defword{preserves equilibria} of $G$ if for every $\trunkS \in \Sigma^\trunk$,
$\trunkS \textnormal{ is a solution of } (\trunk,\vv)$
if and only if $\trunkS \in \textnormal{NE}(G)|_\trunk$.
\end{definition}

\noindent A simple sufficient condition for \edit{equilibrium preservation is that} the expected utility of any strategy $\trunkS$ in $(\trunk,\vv)$ coincides with value of the so-called \defword{value-solving subgame}\footnote{
    Analogously to $G(\trunkS_p)$, $G(\trunkS)$ is obtained by fixing each player's trunk strategy in $G$ to $\trunkS\!$. The name value-solving subgame comes from the fact that solving this subgame will result in a strategy with the correct values (however, the strategy might be highly exploitable \cite{CFR-D}).
} $G(\trunkS)$:

\begin{restatable}{lemma}{EqPreservation}\label{lem:NE_preservation}
Suppose that a value function satisfies $\DLu{p}{\trunkS} = \gv{p}{G(\trunkS)}$ for every $\trunkS \in \Sigma^\trunk$.
Then $\vv$ preserves Nash equilibria of $G$.
\end{restatable}

Unfortunately, equilibrium preservation is not sufficient to efficiently solve depth limited games:

\begin{example}[The correct utility isn't all you need]\label{ex:Oliehoek}
Suppose we have a blackbox that takes a trunk strategy $\trunkS$ and returns a single scalar $\textnormal{BBox}(\trunkS) := \gv{1}{G(\trunkS)}$.\footnote{The value function $V$ studied by \cite{wiggers2016structure} in partially-observable stochastic games is of this type.}
This blackbox is undeniably useful, as it tells us the expected utility of the trunk strategy's ``optimal extension''.
Additionally, setting $\vf{}{\trunkS}{h} := \textnormal{BBox}(\trunkS)$ for each $h$ produces a value function which preserves the equilibria of $G$, since it (trivially) satisfies $\DLu{1}{\trunkS} = \gv{1}{G(\trunkS)}$.
However, combining this value function with algorithms that search through the game tree will be useless, since it provides no information about the individual leaves.
\end{example}

\edit{Now that we have the advertised negative result, we explain what is meant by ``a value function $\vv$ enabling a depth-limited version of an algorithm $A$''.}

\edit{First, before making a depth-limited modification of $A$, we need to \textbf{formulate $A$ in a way that cleanly decomposes the calculation into the trunk part} (to remain unchanged) \textbf{and the bottom part} (to be replaced by $\vv$).
For example, consider the problem of finding the value $\gv{1}{G}$ of a two-player perfect-information zero-sum game $G$.
The formula $\gv{1}{G} := \max_{\sigma_1 \in \Sigma_1} \max_{\sigma_2 \in \Sigma_2} u_1(\sigma_1, \sigma_2)$ produces the correct answer, but it is not obviously amenable to a trunk-bottom decomposition.
On the other hand, the recursive calculation $\gv{1}{G} = v(\textnormal{root})$, $v(h) := $ the maximum/minimum of child values (depending on which player acts at $h$), $v(z) := u_1(z)$ for $z\in \mathcal Z$ \textit{is} obviously decomposable --- we simply replace the recursive call $v(h)$ by a call to the value function $\vv$ when $h$ is at the bottom of the trunk.
Note that for some algorithms, it is unclear (at least to the authors) how they could be usefully decomposed at all --- an example is solving an EFG by a sequence-form linear program \cite{shoham2008multiagent}.}

\edit{Second, a \textbf{depth-limited algorithm can only be coupled with value functions that do not require more context than the algorithm internally produces}.
For example, \textit{any} depth-limited algorithm can be coupled with $\vv$ that doesn't require any context beyond the current history $h$ (e.g., the minimax value function $v$ from the above paragraph).
On the other hand, some value functions can be much more demanding --- for example, the one from Example~\ref{ex:Oliehoek} requires the full trunk strategy $\trunkS$ as context.
Practical algorithms will likely fall somewhere between these two extremes.
For example, in Section~\ref{sec:compactness_and_uniqueness}, we will see that depth-limited CFR (from Example~\ref{ex:DL-CFR} below) can be evaluated one public state $S$ at a time, requiring each player's reach probabilities of infosets in $S$.}

\edit{Finally, \textbf{for $\vv$ to be useful for $A$,} is not sufficient that $A$ runs when coupled with $\vv$ --- instead, \textbf{$A$ coupled with $\vv$ needs to achieve the purpose intended for DL-$A$}, which typically amounts to having guarantees analogous to $A$.
For example, in perfect-information games, the minimax backpropgation algorithm can be used to obtain the minimax strategy, so our intent for DL-minimax might be to find the trunk-portion of this strategy.
For DL-minimax to work as we intended, it is thus sufficient to couple it with the minimax value function described above.
However, other value functions might also suffice, including any $\vv$ that values $ha$ higher than $hb$ (where $a, b \in \mc A(h)$) if and only if the minimax value of $ha$ is higher than the minimax value of $hb$.
Note that an algorithm can have multiple purposes, which might pose different requirements on $\vv$.
For example, minimax can also be used to obtain the minimax value of the game, but DL-minimax coupled with $\vv$ that merely preserves the action-value order would typically fail at this.}

\edit{When speaking informally, we use \defword{$\vv$ enables DL-$A$} to mean that all three conditions above hold.
While defining this notion formally for a general $A$ would be cumbersome, doing so for specific $A$ and ``intent for $A$'' should be straightforward.
For example, we could say that $\vv$ enables DL-minimax iff $\vv$ only depends on the current history (i.e., $\forall h \in \mc Z^\trunk \, \forall \trunkS$, $\rho^\trunk : \vf{p}{\trunkS}{h} = \vf{p}{\rho^\trunk}{h}$) and the resulting DL-minimax outputs the trunk-portion of a minimax strategy in the full game.}

\edit{We conclude this section by describing a depth-limited version of CFR, which will be of central interest throughout the paper.}

\begin{example}[DL-CFR]\label{ex:DL-CFR}
\edit{To define depth-limited CFR, we first need to formulate CFR in a way that is amenable to a trunk-bottom decomposition.
Recall that standard CFR (Section~\ref{sec:cfr}) works by iteratively applying regret matching (eq. \ref{eq:regret-matching}) at each infoset $I$, with respect to regrets}
\begin{align*}
\IavCf{}{\sigma^t}{I}{a} - \IvCf{}{\sigma^t}{I} ,
\end{align*}
\edit{where the values $\IvCf{}{\sigma^t}{I}$ are computed by summing the terminal utilities in leaves \textit{of the full game $G$}:}
\begin{align}\label{eq:cfr-vals}
\IvCf{}{\sigma^t}{I}
 = \sum_{h\in I} \frac{\rp{-p}{\sigma^t}{h}}{\rp{-p}{\sigma^t}{I}} \hv{p}{\sigma^t}{h} , \ 
\textnormal{ where }
\hv{p}{\sigma^t}{h}
= \sum_{z \in \mc Z} \rp{}{\sigma^t}{h,z} u_p(z) 
\end{align}
\edit{(and similarly for $\IavCf{}{\sigma^t}{I}{a}$).
To make \eqref{eq:cfr-vals} amenable to decomposition, we express it in terms of the values of the leaves \textit{of the trunk}:}
\begin{align}
\hv{p}{\sigma^t}{h}
& = \sum_{z \in \mc Z} \rp{}{\sigma^t}{h,z} u_p(z) 
 = \sum_{z' \in \mc Z^\trunk } \sum_{z \in \mc Z} \rp{}{\sigma^t}{h,z'} \rp{}{\sigma^t}{z',z} u_p(z) \\
& = \sum_{z' \in \mc Z^\trunk } \rp{}{\sigma^t}{h,z'} \hv{p}{\sigma^t}{z'} . \label{eq:CFR-rephrase}
\end{align}
\edit{Plugging $\vv$ into \eqref{eq:CFR-rephrase}, the computation only needs to reach the \textit{trunk} leaves:}
\begin{align*}
\DLIvCf{}{\sigma^{t,\trunk}}{I}
	:=
	\sum_{h\in I}
		\frac{\rp{-p}{\sigma^{t,\trunk}}{h}}{\rp{-p}{\sigma^{t,\trunk}}{I}}
		\DLhv{p}{\sigma^{t,\trunk}}{h}
	,
	\textnormal{ where }
	\DLhv{p}{\sigma^{t,\trunk}}{h}
	\sum_{z' \in \mc Z^\trunk }
		\! \!
		\rp{}{\sigma^t}{h,z'} \vf{p}{\sigma^t}{z'}
	\label{eq:DL-infoset-val}
\end{align*}
\edit{(and similarly for $\DLIavCf{}{\sigma^t}{I}{a}$).}
\noindent
\edit{Finally, we define DL-CFR as an algorithm that works identically to standard CFR (Section~\ref{sec:cfr}), except it only computes strategy in the trunk and performs regret matching updates with respect to $\DLIavCf{}{\sigma^t}{I}{a} - \DLIvCf{}{\sigma^{t,\trunk}}{I}$ instead of $\IavCf{}{\sigma^t}{I}{a} - \IvCf{}{\sigma^t}{I}$.}
\noindent
\edit{If we wanted to formalize ``$\vv$ enabling DL-CFR'', we could interpret it, for example, as ``the exploitability of DL-CFR's average (trunk) strategy goes to $0$ as the number of iterations increases''.
}
\end{example}

In the next section, we investigate a class of value functions which are in some sense optimal, allowing DL-CFR and other algorithms to work as intended.

    \subsection{Optimal Value Functions}\label{sec:opt_vfs}

In the previous section, we argued that not all abstract value functions convey a sufficient amount of information about the game. So, which value functions \textit{are} useful?
We start by discussing the obvious and elegant answer (previously adopted by \cite{wiggers2016structure} and \cite[Thm.\,1-3]{ReBeL}):
look at value functions that correspond to Nash equilibria in subgames (Section~\ref{sec:opt_vfs:reach}).
We initially believed that this is where the paper’s story ends.
Unfortunately, this is not the case.
If our goal was only to compute the game-values of subgames \edit{(as in \cite{wiggers2016structure})}, these value functions would be sufficient (Proposition~\ref{prop:enable-DL-u}).
However, Example~\ref{ex:DL-CFR-fail} shows that these value functions can fail to be useful for practical algorithms such as CFR or fictitious play.

Unfortunately, this is not the case, as this answer turns out to be insufficient for many purposes.
However, such value functions can fail to be useful for practical algorithms such as CFR or fictitious play (Example~\ref{ex:DL-CFR-fail}).
Fortunately, value functions can be useful for these purposes if they satisfy additional properties (Section~\ref{sec:opt_vfs:cf}).\footnote{
However, such functions do suffice for these purposes if they satisfy additional properties (see Section~\ref{sec:opt_vfs:cf}).
    One way to satisfy these extra properties is to solve subgames by CFR (in a particular manner), the value functions used in the implementation of ReBeL \cite{ReBeL} are likely to be of this stronger type.}

To give a better answer, we introduce the notion of optimality with respect to a collection of infosets (Section~\ref{sec:opt_vfs:defs}).
Informally speaking, an extension of a trunk strategy is optimal with respect to an infoset if the infoset's owner cannot improve their expected utility at that infoset by changing their strategy in the bottom of the game. A strategy’s extension is better if it is optimal with respect to a bigger collection of infosets.
And a value function is better if it corresponds to better extensions.
In this terminology, the initial wrong answer proposed to use value functions that are optimal only with respect to infosets that are reachable (i.e., have non-zero probability of being encountered under the given trunk strategy).

The \textit{most practical} alternative answer we have identified is to use a stronger notion of \textit{counterfactual optimality}, where players maximize infoset values for all infosets that their opponent’s strategy allows them to enter.
(That is, those that are reachable by the given player in the counterfactual scenario where the player decides to do so; see Section~\ref{sec:opt_vfs:cf}.)
We show that obtaining value functions of this type is not difficult --- we can \edit{either use CFR to find the Nash equilibria of subgames or} apply a simple post-processing step to Nash equilibria found by an arbitrary method.
\edit{(The former is done in \cite{ReBeL}.)}
Moreover, they enable a depth-limited variant of CFR, best response, and likely also fictitious play.
We also describe a simple example where depth-limited CFR and fictitious play both fail when combined with a value function that is reachably optimal but not counterfactually optimal.

\edit{Finally, we also briefly discuss an even stronger notion of universally-optimal value functions which maximize infoset values in \textit{all} infosets, including those reachable only when both the player and their opponent deviate from their strategy.}


\subsubsection{Optimality Criterion}\label{sec:opt_vfs:defs}

In this section, we describe the optimal value functions promised above.
The starting point is the notion of optimality with respect to a single infoset, which states that the infoset's owner cannot improve the infoset's expected utility by switching to a different strategy:
\begin{equation}\label{eq:optimal_infoset_value}
	\Iv{p}{\sigma}{I}
	= \max_{\rho_p \in \Sigma_p} \Iv{p}{\rho_p,\sigma_{-p}}{I}
	=: \Iv{p}{*,\sigma_{-p}}{I}
	\textnormal{, where } I \in \mc I_p.
\end{equation}

Before proceeding with the formal definitions, Remark~\ref{rem:infoset-opt} provides further intuition regarding the usage of this criterion.
\begin{remark}[Understanding infoset optimality]\label{rem:infoset-opt}
Firstly, we see that the maximum in \eqref{eq:optimal_infoset_value} is taken over all strategies of $p$.
However, recall that as far as $p$'s strategy is concerned, $\Iv{p}{\rho_p,\sigma_{-p}}{I}$ only depends on the restriction of $\rho_p$ to the part $\Succ{I} := \{ J \in \mc I_p \mid J \sqsupset I \}$ of $p$'s infoset tree that lies below $I$ (including $I$; this follows from Theorem~\ref{thm:infoset-v-characterization}).
As a result, it doesn't matter whether the maximum is taken over the whole $\Sigma_p$, over $\Sigma_p^{\Succ{I}}$, or over any set inbetween.

Second, which infosets should be required to satisfy \eqref{eq:optimal_infoset_value}?
Since our goal is to design value functions for depth-limited solving, we can start by restricting the condition to the infosets that lie in $\mc Z^\trunk$.
Among all possible extensions $\sigma$ of $\trunkS$, it is -- \textit{all else being equal} -- better if $\sigma$ is optimal with respect to as many $I$'s as possible.
Indeed, we can view $\vf{}{\trunkS}{h}$ as providing information about $h$ --- if $h$ is unreachable under $\trunkS$, we might nevertheless find uses for any pieces of ``off-policy data'' about it.
Another useful metaphor is to view game-solving as following a gradient of strategy quality --- if the extension $\sigma$ is optimal w.r.t. more infosets, the gradient will convey more information and we can learn faster.
However, the assumption of ``all else being equal'' does not hold in practice.
Indeed, the goal will typically be to obtain the \textit{easiest-to-find} extension $\sigma \supset \trunkS$ which is \textit{informative enough} for our method of choice.
\end{remark}

The following definition gives a general ``blueprint'' for various versions of optimality and introduces the three types that will be the most relevant for us.
Informally speaking, it simply states that reachably (resp. counterfactually, resp. universally) optimal \textit{extensions of a trunk strategy} are those for which $\Iv{p}{\sigma}{I} = \Iv{p}{*,\sigma_{-p}}{I}$ holds in all reachable (resp. counterfactually-reachable, resp. all) infosets\footnote{The different versions of reachability are defined in Section~\ref{sec:sub:infoset-reach}.} at the bottom of the trunk.

\begin{definition}[Optimal extensions]\label{def:optimal-extension}
Let $\mc J \subset \{ I \in \mc I \mid I \subset \mc Z^\trunk \}$ be collection of leaf-infosets for a trunk $\trunk$ and $\trunkS \in \Sigma^\trunk$.
An extension $\sigma$ of $\trunkS$ is said to be \defword{optimal} on $\mc J$, denoted $\sigma \in \OptE{\trunkS}{\mc J}$, if
	$\Iv{}{\sigma}{I} = \Iv{}{*,\sigma_{-p}}{I}$ holds for every $I \in \mc J$.

The extension is said to be \defword{reachably-optimal} if it is optimal on $\mc J = \{ I \subset \mc Z^\trunk \mid I \in \mc I_p \textnormal{ reachable} \}$.
Analogously, we define \defword{counterfactually-optimal} and \defword{universally-optimal} extensions as those corresponding to $\mc J = \{ I \subset \mc Z^\trunk \mid I \in \mc I_p \textnormal{ counterfactually reachable} \}$, resp. $\mc J = $ all infosets in $\mc Z^\trunk$.
%
\end{definition}

As we indicated earlier, any notion of optimality of extensions can be derived from the corresponding notion of optimality of values.
Informally speaking, \defword{reachably-optimal value functions} are those that correspond to reachably-optimal extensions \emph{on reachable infosets} (but can take arbitrary values elsewhere), \defword{counterfactually-optimal value functions} correspond to cf.-optimal extensions \textit{on cf.-reachable infosets} (but have arbitrary values elsewhere), and \defword{universally-optimal value functions} are those that corresponds to universally-optimal extensions (on the whole $\mc Z^\trunk$):

\begin{definition}[Optimal value functions]\label{def:optimal-vf}
Let $\trunk$ be a trunk and $\vec{\mc J} = (\mc J^{\trunkS})_{\Sigma^\trunk}$,  $\mc J^{\trunkS} \subset \{ I \in \mc I \mid I \subset \mc Z^\trunk \}$.
A value function $\vv : \mc Z^\trunk \times \Sigma^\trunk \to \R$ is \defword{$\vec{\mc J}$-optimal} if it ``corresponds to extensions that are optimal on $\bigcup \mc J^{\trunkS}$'' in the following sense:
\begin{align*}
\left( \forall \trunkS \in \Sigma^\trunk \right) \left( \exists \sigma \in \OptE{\trunkS}{\mc J^{\trunkS}} \right) \left( \forall h \in \bigcup \mc J^{\trunkS} \right) : \vf{}{\trunkS}{h} = \hv{1}{\sigma}{h} .
\end{align*}
\end{definition}


In Sections~\ref{sec:opt_vfs:reach}-\ref{sec:opt_vfs:univ}, we investigate these different variants of optimality in the order of increasing strictness.
We explain how to compute the (first two types of) value functions, lists some of the depth-limited algorithms they enable, and explain some of their shortcomings.

\subsubsection{Reachably-Optimal Value Functions}\label{sec:opt_vfs:reach}

Reachably-optimal value functions are the easiest to obtain, but also only ``enable'' fewer depth-limited computations than the other two types of optimal value functions.
The following result shows that reachably-optimal values can be computed by fixing the trunk strategy and solving the remainder of the game:

\begin{restatable}[Computing reachably-optimal values]{proposition}{ComputingReachOptVf}\label{prop:comp-reach-optimal}
Suppose that for each $\trunkS$, there is some $\edit{\sigma \in}\, \textnormal{NE}(G(\trunkS))$ s.t. $\vf{}{\trunkS}{h} = \hv{1}{\edit{\sigma}}{h}$ holds for all $h \in \mc Z^\trunk$.
Then $\vv$ is a reachably optimal value function.
\end{restatable}

The proof goes by showing that if an extension of $\trunkS$ isn't reachably optimal, one of the players can gain by deviating from it.
Note that in the other direction, it can also be shown that all reachably-optimal value functions are of this form, except for the fact that they can be defined arbitrarily on the unreachable parts of $\mc Z^\trunk$.


Proposition~\ref{prop:enable-DL-u} shows that as long as we are only after computing the \textit{utility} of $\trunkS$'s best extension (i.e., the subgame value), reachably-optimal value functions are sufficient.

\begin{restatable}[Enabling utility calculation]{proposition}{EnableDLu}\label{prop:enable-DL-u}
Any reachably-optimal value function satisfies
\begin{align}
 \DLu{p}{\trunkS}
& = \sum_{h \in \mathcal Z^\trunk} \rp{}{\trunkS}{h} \vf{p}{\trunkS}{h}
= \gv{p}{G(\trunkS)} \label{eq:DL-u_prop} .
\end{align}
\end{restatable}

\noindent
This can be shown by combining Theorem~\ref{thm:infoset-v-characterization} with the fact that expected utility of a strategy only depends on the values of reachable histories.
In particular, this proposition demonstrates that for any trunk-strategy, reachably-optimal value functions enable us to compute the expected utility (at the root) obtained if both players play ``optimally'' in the remainder of the game. (Because no matter which version of optimality we choose, this number will be equal to $\gv{p}{G(\trunkS)}$.)
Together with Lemma~\ref{lem:NE_preservation}, Proposition~\ref{prop:enable-DL-u} immediately yields the following corollary:

\begin{theorem}[Equilibrium preservation]\label{thm:reach_opt_preserve_NE}
Reachably-optimal value functions preserve Nash equilibria.
\end{theorem}

By Theorem~\ref{thm:reach_opt_preserve_NE}, it is also true that the equilibria of the corresponding depth-limited game $(\trunk,\vv)$ are precisely the equilibria of $G$ restricted to $\trunk$.
Unfortunately, this is not the same as offering a practically viable method for finding these equilibria. Indeed, as shown by Example~\ref{ex:DL-CFR-fail}, reachably-optimal value functions might fail to enable depth-limited variants of more efficient solution methods such as CFR (and fictitious play, as argued later):

\begin{example}[Reachably-optimal value functions do not enable DL-CFR]\label{ex:DL-CFR-fail}
Let $G$ be a two-round simultaneous-move (but otherwise perfect-information) game that works as follows:
The first round looks like matching pennies\footnote{In matching pennies, both players can select either Heads or Tails. If the choices match, player 2 pays one point to player 1. Otherwise, player 1 pays one point to player 2.}, except that the utility for heads-heads is $\pm (1+\epsilon)$ instead of $1$ (this is to break the symmetry while running CFR).
Only player 1 acts in the second round and has a choice between `doing nothing' ($dn$; no change to utilities) or `going mad' ($gm$; transfer 1000 utility to the opponent).
The only Nash equilibrium of $G$ is to `do nothing' in the second round and play something very close to the uniform strategy in the first round.

Consider the trunk $\trunk := $ the first round, for which each information set just below $\trunk$ corresponds to an element of $\mc Z^\trunk = \{ (H,H), (H,T), (T,H), (T,T) \}$.
For any trunk strategy $\trunkS$, the only extension that satisfies $\Iv{p}{\sigma}{I} = \Iv{p}{*,\sigma_{-p}}{I}$ for \textit{every} $I$ (independently of its reachability) is the $\sigma$ that `does nothing' in the second round of the game, independently of what has happened in the trunk.
However, if we only require optimality in reachable states, $\sigma$ is suddenly allowed to `go mad' in states that are only reachable counterfactually.
In particular, the following value function is reachably optimal:
\begin{align*}
\vf{}{\trunkS}{h} & := u_1(h,gm) \textnormal{ whenever } \rp{1}{\trunkS}{h} = 0, \\
\vf{}{\trunkS}{h} & := u_1(h,dn) \textnormal{ otherwise.}
\end{align*}
In other words, $\vv$ corresponds to player 1 always believing that they would go mad if they deviated from their current trunk strategy.

We claim that running DL-CFR on $(\trunk,\vv)$ produces a strategy where player 1 chooses H with (near) certainty --- i.e., one that is very far from being a Nash equilibrium.
To see this, note that DL-CFR starts out with a uniform trunk strategy, for which $\vv$ predicts the `do nothing' action everywhere in the bottom.
The first CFR update thus works as if we were running CFR on the one-round game.
Since the utility for $(H,H)$ is slightly above 1, the maximizing player will play pure $H$ in the second iteration, while the minimizing player 2 will play pure $T$.
Since $(H,T)$ is an optimal outcome for player $2$, they will only deviate from playing $T$ if player 1 deviates from $H$.
However, given how $\vv$ is defined, this will never happen. Indeed, while player 1 sees that the $(H,T)$ outcome yields -1 utility, they believe that playing $(T,T)$ would result in $u_1(T,T,gm) = -999$ utility.
The average strategy of DL-CFR in $(\trunk,\vv)$ will thus converge to the highly-exploitable strategy $(H,T)$.
\end{example}

However, despite not being sufficiently informative for CFR, reachably-optimal value functions will nevertheless be suitable for other algorithms.
For example, they are a natural candidate for obtaining a depth-limited variant of \textit{information-set MCTS} (IS-MCTS) \cite{ISMCTS}.
The standard IS-MCTS algorithm works by running some MCTS algorithm on $\mc H$ while sharing statistics between all histories within every infoset.
(The algorithm is far from state of the art and lacks theoretical guarantees, so it is best used as a cheap baseline solution.)
The corresponding \defword{depth-limited IS-MCTS} algorithm can thus be obtained by stopping the MCTS recursion in the \textit{trunk} leaves and using the values $\vf{}{(\cdot)}{h}$ corresponding to the current strategy\footnote{To have the current strategy defined in the whole trunk, including in places that MCTS hasn't yet explored, we can assume that the strategy is initialized to be uniformly random.}.

We could also consider \defword{evolutionary methods} for playing imperfect information games: for example, we could have a population of agents play each other and keep the successful ones, modifying them slightly before the next iteration.
All that is needed to execute a depth-limited version of this method is to obtain the correct expected utilities at the root.
By Proposition~\ref{prop:enable-DL-u}, reachably-optimal value functions are sufficient for the task.

\subsubsection{Counterfactually-Optimal Value Functions}\label{sec:opt_vfs:cf}

We now turn to the stronger notion of \textit{counterfactual} optimality.
Informally speaking, counterfactually-optimal value functions give a ``correct answer'' to the question ``so, this information set \textit{that I didn't play into}, what would have happened if I \textit{did} play there?''. (Which was not the case for \textit{reachably}-optimal value functions.)
As we will see, this ensures that using such functions together with depth-limited CFR produces provably correct strategies.
We also show how cf.-optimal value functions enable the depth-limited computation of best-response and explain how this could be useful for enabling depth-limited fictitious play.

As observed in \cite{Neil_thesis}, counterfactual optimality can be achieved by \textit{post-processing} the Nash equilibria of value-solving subgames \edit{(described below Def.~\ref{def:equil_preservation})}.
This can be done via ``counterfactual best-response'', i.e., a recursively calculated pure strategy $cbr_p \in \Sigma_p$ that takes some action from $\argmax_{a\in\mc A(I)} \Iav{}{cbr_p,\sigma_{-p}}{I}{a}$ in each $I\in \mc I_p$ that is counterfactually reachable by $p$.

\begin{restatable}[Computing counterfactually-optimal values]{proposition}{CompCfOptVf}\label{prop:computing-cf-opt}
Suppose that for each $\trunkS$, a value function $\vv$ is of the form $\vf{}{\trunkS}{h} = \hv{1}{\mu}{h}$, where $\mu\in \Sigma$ is obtained by
\begin{itemize}
\item starting with some $\sigma \in \textnormal{NE}(G(\trunkS))$,
\item \edit{for both $p = 1,2$,} going through all $I\subset \mc Z^\trunk$, $I \in \mc I_p$, that are counterfactually reachable by $p$ but not reachable,
\item and replacing $\sigma_p$ by $cbr_p(\sigma_{-p})$ on such infosets and their descendants.
\end{itemize}
Then $\vv$ is counterfactually optimal.
\end{restatable}

A particular consequence (of the proof) is that every reachably-optimal value function can be made counterfactually optimal by ``fixing'' the values in infosets that are only reachable counterfactually.
\edit{In practice, we can also obtain counterfactually-optimal value functions by running CFR on the bottom of the game.\footnote{\edit{More precisely, by applying CFR to the full game $G$ but keeping the strategies fixed to $\sigma^t(I) := \trunkS(I)$ for all $I\subset \trunk$ and $t$ (instead of updating them via regret matching as usual).}}
This produces a NE of $G(\trunkS)$ in which the values in all counterfactually reachable infosets are already maximal \cite{CFR}, so the post-processing step will be unnecessary.}

\paragraph{Depth-Limited CFR}

In Example~\ref{ex:DL-CFR}, we have described the depth-limited variant of CFR.
With counterfactually-optimal value functions, we are now ready to prove that DL-CFR converges to a Nash equilibrium of the game.

Informally, CFR works by finding a strategy in which both players maximize their counterfactual values $\rp{-p}{\sigma}{I} \Iv{p}{\sigma}{I}$, which is equivalent to maximizing the values $\Iv{p}{\sigma}{I}$ in counteractually reachable infosets.
In comparison, DL-CFR works by
(i) explicitly minimizing the (cf.) regret in the trunk and
(ii) implicitly minimizing the (cf.) regret in the remainder of the game (by using a value function that corresponds to a zero-regret strategy).
One way to prove the correctness of DL-CFR is thus to formalize this idea and mimick the standard CFR proof.
However, since the existing CFR-D algorithm\footnote{Recall that the D in CFR-D stands for ``decomposition'' (rather than anything to do with depth-limited solving) \cite{CFR-D}. Indeed, CFR-D works by decomposing the whole bottom of the game into many smaller subgames. However, this does require first splitting the game into the trunk and the bottom, which makes many of the ideas relevant for depth-limited solving.} \cite{CFR-D} also relies on a similar idea, we can show that our DL-CFR approach fits the CFR-D formalism, which guarantees that the existing CFR-D results apply to DL-CFR as well:

\begin{restatable}[Enabling DL-CFR]{proposition}{EnablingDLCFR}\label{prop:enable-DL-CFR}
Let $(\trunk,\vv)$ be a depth-limited game corresponding to a counterfactually-optimal value function. Then:
\begin{enumerate}[(1)]
\item DL-CFR can be viewed an instance of CFR-D and inherits its guarantees.
\item In particular, the strategy $\overline \sigma^{\trunk, t}$ produced after $t$ iterations of DL-CFR satisfies $\expl{}{}{\overline \sigma^{\trunk, t}} \overset{t\to \infty}{\longrightarrow} 0$.
\end{enumerate}
In general, the same is not true for reachably-optimal value functions.
\end{restatable}

\paragraph{Depth-Limited Best-Response and Fictitious Play}

We now show that counterfactually optimal value functions also enable an efficient computation of best response.
In Notation~\ref{not:DL-concepts}, we have defined a best response to $\sigma_{-p}^\trunk$ in $(\trunk,\vv)$ as the strategy $\sigma^{\mc  T}_p \in \Sigma^\trunk_p$ that maximizes the expression
\begin{equation*}
\DLu{p}{\mu_p^\trunk, \sigma_{-p}^\trunk} = \sum_{I \subset \mc Z^\trunk} \rp{}{\mu_p^\trunk, \sigma_{-p}^\trunk}{h} \vf{p}{\mu_p^\trunk, \sigma_{-p}^\trunk}{h} .
\end{equation*}
However, this type of best response is difficult to calculate, because the optimal course of action in the bottom part of the game depends on the trunk strategy, which in turn depends on the bottom part.\footnote{Proposition~\ref{prop:enable-DL-CFR} implies that this could be done by fixing the opponent's trunk strategy to $\sigma_{-p}^\trunk$ and solving the resulting 1-player depth-limited game by DL-CFR.}
We could instead use a more efficient but less rigorous approach which pretends that modifying the trunk strategy has no influence on the output of the value function:

\begin{definition}[Naive best-response]
A trunk strategy $\trunkS_p$ is a \defword{naive best-response} to a trunk strategy \emph{profile} $(\rho^\trunk_p,\trunkS_{-p})$ in $(\trunk,\vv)$ if it satisfies
\begin{equation*}
\trunkS_p \in \argmax_{\mu_p^\trunk} \sum_{h \in \mc Z^\trunk} \rp{}{\mu_p^\trunk, \sigma_{-p}^\trunk}{h} \vf{p}{\rho^\trunk_p,\trunkS_{-p}}{h} .
\end{equation*}
\end{definition}

\noindent Because of its dependency on $\rho_p^\trunk$, a naive best-response might not be a true best response.
However, for counterfactually optimal values, this concept nevertheless preserves a crucial property of standard best responses:

\begin{restatable}[NE as mutual naive best-responses]{proposition}{NaiveBR}
If $\vv$ is a a counterfactually optimal value function, then any $\trunkS \in \Sigma^\trunk$ where each $\sigma_p^\trunk$ is a naive best-response to $\trunkS$ in $(\trunk,\vv)$ can be extended into a Nash equilibrium in $G$.
\end{restatable}

\noindent As with DL-CFR, the result doesn't hold if $\vv$ is only \textit{reachably} optimal (as witnessed using the counterexample constructed in Example~\ref{ex:DL-CFR-fail}).

Another algorithm that might be particularly suited for depth-limited modification is fictitious play (FP) \cite{heinrich2015fictitious}, which can be formulated as a process that keeps a growing list of strategies, where each newly added strategy is a best response to the average of the strategies identified thus far (and that returns this average when terminated).
Replacing best response by naive best response will therefore produce \defword{depth-limited fictitious play}, which should work analogously to standard FP.
Unfortunately, there are games in which fictitious play fails to find a strong strategy even when given infinite time.
From this reason, we defer the analysis of its depth-limited variants to a future work.
However, we believe that to the extent that FP does have some performance guarantees, DL-FP should be combined with counterfactually optimal value functions to preserve them.

\subsubsection{Universally-Optimal Value Functions}\label{sec:opt_vfs:univ}

\edit{Universal optimality is the strongest of the tree variants of optimality introduced in Definition~\ref{def:optimal-vf} --- universally-optimal value functions correspond to strategies whose values are optimal for \textit{all} infosets in $\mc Z^\trunk$, including those that are completely unreachable under the given trunk strategy.
While this notion often unnecessarily strong,
it allows us to make an important connection between our analysis and value functions in perfect information games.
Indeed, recall that perfect information games have a canonical a value function --- the one that assigns to each state its minimax value.
Since reachably-optimal and counterfactually-optimal values can be arbitrary in (counterfactually) unreachable parts of $\mc Z^\trunk$, neither of them coincides with minimax values when applied to perfect information games.
In contrast, universally-optimal value functions \textit{do} successfully generalize minimax values since they are optimal everywhere.
However, since counterfactual optimality is sufficient for the purposes of this paper, we will not discuss universal optimality in more detail.}
\subsection{Efficient Computation of Optimal Value Functions}\label{sec:compactness_and_uniqueness}

Our analysis so far has focused on theoretical properties of optimal value functions, leaving aside the surrounding computational considerations.
In this section, we show how to relax and generalize some of the formal definitions to make dealing with optimal value functions more practical.
We also analyze properties of these functions that are relevant to their approximation.

One formal limitation of our earlier definitions is that optimal value functions require the player’s infoset values to be at least as high as values under \textit{all} other strategies (Section~\ref{sec:noam}).
This is both difficult to verify and unnecessary --- in practice, we might do \textit{equally} well by relaxing the assumption to only include some representative subset of strategies (e.g., pure undominated strategies).
By pruning the considered strategy set even further, we can make \textit{trade-offs} between computational costs and theoretical guarantees.
We show that this approach can be implemented in a way that reduces to the existing ``multivalued states'' method described in \cite{brown2018depth}.
This exposes a previously unrecognized connection between two coexisting depth-limited approaches to solving imperfect information games, \cite{DeepStack} and \cite{brown2018depth}, and opens up the potential for their cross-fertilization. 

A second formal limitation is that value functions take the \textit{whole} trunk strategy as an input and return a separate value \textit{for each history} at the bottom of the trunk (Section~\ref{sec:compact-representations}).
To decrease the size of the input and localize the computation of values,
we show that public belief states provide a sufficient context for computing optimal values.
Moreover, we show that information about anything less than a public state might fail at this task.
We also argue that for practical purposes, the output should be aggregated over infosets, as is often done in the recent literature \cite{DeepStack,ReBeL}.

Thirdly, we remark that in practice, it will be desirable to use value functions that are only $\epsilon$-optimal, typically in conjunction with neural networks.
For this purpose, it would be desirable to have a unique approximation target.
However, it turns out that optimal value functions are not uniquely defined.
We thus argue that the next best thing to hope for is that the set of such functions is convex.
We show that this is true for reachably-optimal value functions (this is both straightforward and already known).
As an open problem, we ask whether the same is true for counterfactual and universal optimality --- we provide some evidence both for and against this claim.
Since these details might be of interest to fewer readers, they are presented in \ref{sec:uniqueness}.



\subsubsection{Alternative Computation of Optimal Value Functions\texorpdfstring{\\}{}and the Relation to Multi-Valued States}\label{sec:noam}

\edit{In this section, we show an alternative method of computing value functions, connecting the value-function approach to depth-limited search with the multi-valued states introduced in \cite{brown2018depth}.}

\edit{To motivate the approach, let us inspect the condition which appears in the definition of optimality (Definition~\ref{def:optimal-extension}).
To count as a (reachably) optimal extension of a trunk strategy $\trunkS$, a strategy $\sigma$ needs to satisfy the following (for every reachable trunk-infoset of $p$, $p=1,2$):}
\begin{align*}
     (\forall \rho_p \in \Sigma_p) : \Iv{}{\sigma_p, \sigma_{-p}}{I} \geq \Iv{}{\rho_p,\sigma_{-p}}{I} .
\end{align*}
\edit{However, taken literally, this condition requires comparing $\sigma_p$ to all other strategies $\rho_p \in \Sigma_p$, and there are uncountably many of those.
Instead, we can consider a weaker -- but often sufficient -- requirement that $\sigma$ performs at least as well as any strategy from some \textit{portfolio} of strategies $\portfolio \subset \Sigma^{\mc H \setminus \trunk}$.
The following definition captures this idea formally:}

\begin{definition}[Optimality w.r.t. a portfolio]\label{def:optimality:generalized-strat}
\edit{Let $\trunkS \in \Sigma^\trunk$ and $\portfolio = \portfolio_1 \times \portfolio_2 \subset \Sigma^{\mc H \setminus \trunk}$.
An extension $\sigma$ of $\trunkS$ is said to be reachably \defword{optimal with respect to} $\portfolio$ if for every reachable trunk-leaf infoset $I \in \mc I_p$,
\begin{enumerate}[(i)]
    \item $\sigma$ below $I$ is realizable by $\portfolio$ (i.e., the restriction of $\sigma_p$ to $\{ J \in \mc I_p \mid J \sqsupset I \}$ is a convex combination of elements of $\portfolio_p$) and
    \item for every $\rho^\downarrow_p \in \portfolio_p$, $\Iv{}{\sigma_p, \sigma_{-p}}{I} \geq \Iv{}{\rho^\downarrow_p,\sigma_{-p}}{I}$.\footnote{\edit{
        Recall that by Theorem~\ref{thm:infoset-v-characterization}, $\Iv{}{\rho^\downarrow_p,\sigma_{-p}}{I}$ is well-defined even for a partial strategy $\rho^\downarrow_p$.}
    }
\end{enumerate}
A value function $\vv$ is reachably optimal w.r.t. $\portfolio$ if for each $\trunkS$, $\vf{p}{\trunkS}{\cdot} = \hv{p}{\sigma}{\cdot}$ for some extension $\sigma$ of $\trunkS$ that is reachably optimal w.r.t. $\portfolio$.
Counterfactual and universal optimality w.r.t. $\portfolio$ is defined analogously.}
\end{definition}


\edit{In the extreme case where $\portfolio$ contains \textit{all} strategies, the notion trivially coincides with reachable optimality.
However, we might get a high-quality value function even with a much smaller portfolio:}
\edit{To start with, we can reduce $\portfolio$ without hurting the quality of $\vv$ at all.
Indeed, Proposition~\ref{thm:portfolio-and-vf} provides a theoretical lower bound on the degree of such reduction, and a much more extensive (yet still lossless) reduction might be possible in practice:}

\begin{restatable}[Inspired by \cite{brown2018depth}]{proposition}{pureUndomPreserveNE}
    \label{thm:portfolio-and-vf}
\edit{Suppose that $\vv$ is reachably\footnote{
    A limitation of Proposition~\ref{thm:portfolio-and-vf} is its non-constructiveness --- it talks about the solutions of $(\trunk, \vv)$ without explaining how to obtain them.
    Fortunately, if we further assumed that $\vv$ is \textit{counterfactually} optimal (w.r.t. $\portfolio$),
    we could prove that it enables depth-limited variants of constructive methods such as CFR
    (analogously to how adding counterfactual optimality on top of Theorem~\ref{thm:reach_opt_preserve_NE} gives Proposition~\ref{prop:enable-DL-CFR}).
    For brevity, we refrain from discussing this modification in more detail.
}
optimal w.r.t. a portfolio that contains (the trunk restrictions of) all pure undominated strategies.
Then $\vv$ preserves the equilibria\footnote{
    \edit{Recall that, by Definition~\ref{def:equil_preservation}, $\vv$ is said to preserve equilibria of $G$ if any solution of $(\trunk, \vv)$ is a restriction of some NE of $G$ to $\trunk$, and vice versa.}
}
of $G$.}
\end{restatable}

\noindent
\edit{Further reduction in the portfolio's size can come at \textit{some} cost to the quality of $\vv$.
However, we might be able to reduce $\portfolio$ down to a handful of strategies while only harming $\vv$ a little.
For example, \cite{brown2018depth} constructed a competitive agent for heads-up no-limit Texas hold'em poker using less than ten strategies\footnote{
    By Theorem~\ref{thm:portfolio-game} below, the approach from \cite{brown2018depth} is equivalent to the one in Proposition~\ref{thm:portfolio-and-vf}.
}.}

\medskip

\edit{A practical method of computing optimal values can be devised by considering what we might call a ``partial normal-form representation'' of the game.
The underlying idea is that instead of taking \textit{all} actions sequentially, the players can only play sequentially until they reach the bottom of the trunk, at which point they simultaneously\footnote{Recall that the simultaneous decision is equivalent to player one choosing first and player two choosing second, without learning the first player's choice.} announce the strategy they would use in the bottom part of the game.
The game then immediately terminates, giving each player the expected utility corresponding to combination of the selected strategies.
When the choice of the bottom strategy is restricted to the portfolio, this corresponds to the following game:}

\begin{definition}[Inspired by \cite{brown2018depth}]\label{def:portfolio-game}
\edit{For a trunk $\trunk$ and portfolio $\portfolio$, $G(\trunk, \portfolio)$ is a game which proceeds identically to $G$, up to the point where it reaches some trunk-leaf $z \in \mc Z^\trunk$.
The information sets on $\mc Z^\trunk$ are the same as in $G$.
In $z \in \mc Z^\trunk$, the players simultaneously select some $\sigma^\downarrow_p \in \portfolio_p$.
The resulting history $z\sigma^\downarrow_1\sigma^\downarrow_2$ is terminal in $G(\trunk, \portfolio)$ and yields utility $u_p(z\sigma^\downarrow_1\sigma^\downarrow_2) := \hv{p}{\sigma^\downarrow_1,\sigma^\downarrow_2}{z}$.}
\end{definition}

\edit{As advertised, the equilibria of the game $G(\trunk, \portfolio)$ coincide with the solutions of the depth-limited game $(\trunk, \vv)$ whose value function corresponds to $\portfolio$:}

\begin{restatable}{theorem}{portfolioGame}\label{thm:portfolio-game}
\edit{For any $\trunk$, $\portfolio$, and $\vv$ that is reachably optimal w.r.t. $\portfolio$, a trunk strategy $\trunkS$ is solution of $(\trunk, \vv)$ if and only it is in $\textnormal{NE}(G(\trunk,\portfolio))|_{\trunk}$.}
\end{restatable}


%


\edit{Finally, Theorem~\ref{thm:portfolio-game} has an important implication for the existing research on depth-limited solving (DLS):
In the past, one line of DLS research was based on value functions (primarily building on \cite{DeepStack}).
Around the same time, \cite{brown2018depth} suggested a second approach called multi-valued states.
In our terminology, this approach consists of using the game $G(\trunk, \portfolio)$ with $\portfolio_1 := $ all pure strategies (resp. a small set of heuristic strategies in practice) and $\portfolio_2$ which consists of a single ``blueprint strategy'' (typically obtained by solving an abstraction of $G$).
Initially, these approaches might seem incompatible or unrelated --- they certainly did to the authors of this text.
However, with Theorem~\ref{thm:portfolio-game}, we see that they constitute two different methods of approximating the same value function.
Moreover, now that the connection has been made explicit, we can make progress by combining the ideas behind the two approaches.
For example, when approximating a value function by a neural network, we can use the games $G(\trunk, \portfolio)$ to generate the training data.}
\subsubsection{Compact Representation of Optimal Value Functions}\label{sec:compact-representations}

In this section, we discuss how to represent the input and output of optimal value functions more compactly.
\edit{These improvements are particularly useful in practical applications, where value functions are often approximated by neural networks.}

\paragraph{Input}

The input can be compressed in two steps: first, we can ``flatten'' the trunk strategy by only considering the corresponding reach probabilities.
Second, we can ``narrow it down'' by only considering reach probabilities that are relevant for computing the value of the specific history.
The first part is formally captured by the following definition (inspired by \cite{wiggers2016structure}) and its \edit{simple corollary}:

\begin{definition}[Sufficient statistic]\label{def:suff_stat}
A quantity $s : \Sigma^\trunk \times \mathcal Z^\trunk \to \mathcal S$
is a \defword{sufficient statistic} for $\vv$ \edit{(resp. for $\vv$ on $Z\subset \mc Z^\trunk$)} if there is some $\tilde \vv : \mathcal Z^\trunk \times \mathcal S \edit{\to \R}$ s.t. we have 
$\vf{}{\trunkS}{h} = {\tilde{\mathbf{v}}}^{s(\trunkS,h)}(h)$ for each $h \in \mc Z^\trunk$ \edit{(resp. $h \in Z$)} and $\trunkS$.
\end{definition}

\begin{restatable}[Sufficient statistics for optimal value functions]{proposition}{SufficientStats}
\label{prop:range_suffices}
Let $\trunk$ be a trunk and $\trunkS \in \Sigma^\trunk$.
\begin{enumerate}[(1)]
\item The (joint) reach probabilities $(\rp{}{\trunkS}{h})_{h\in \mc Z^\trunk}$ provide a sufficient statistic for computing some reachably-optimal $\vv$.
\item The factored reach probabilities $( \rp{1}{\trunkS}{h}, \rp{2}{\trunkS}{h} )_{h\in \mc Z^\trunk}$ provide a sufficient statistic for computing some counter\-fact\-ually-op\-ti\-mal $\vv$.

In particular, it suffices to keep $(\rp{1}{\trunkS}{I})_{\mc Z^\trunk \supset I \in \mc I_1}$ and $(\rp{2}{\trunkS}{I})_{\mc Z^\trunk \supset I \in \mc I_2}$.
\end{enumerate}
\end{restatable}

Moreover, the decomposition of $\mc Z^\trunk$ into public states allows us to compute value functions in a more \textit{localized} manner, by looking at a single subgame at a time.
Formally, we have the following \edit{(well-known but previously unpublished)} result:

\begin{restatable}[Localization by public states]{proposition}{Localization}\label{prop:public-state-sufficiency}
For any public state $S \edit{\subset} \mc Z^\trunk$:
\begin{enumerate}[(i)]
\item Both $(\rp{}{\trunkS}{h})_{h\in S}$ and $(P^{\trunkS}(h|S))_{h\in S}$ provide a sufficient statistic for computing some reachably-optimal value function $\vf{}{\trunkS}{\cdot}$ on $S$.\label{case:normalized-range}
\item $(P^{\trunkS}_1(I))_{S\supset I \in \mc I_1}$ and $(P^{\trunkS}_2(I))_{S\supset I \in \mc I_2}$ together provide a sufficient statistic for computing some counterfactually-optimal $\vf{}{\trunkS}{\cdot}$ on $S$.\label{case:range}
\end{enumerate}
\end{restatable}

We give an abstract proof of this result, showing that the definitions of appropriate optimal value functions can be rephrased to only depend on the statistics listed in the corresponding case of this proposition.
This implies that it must be possible to compute the desired value function using those statistics only.
Inspired by the poker literature, we refer to these statistics as range:

\begin{definition}[Range]\label{def:range}
For a (possibly partial) strategy $\sigma$ and public state $S$, the \defword{range} at $S$ (corresponding to $\sigma$) is defined as 
\begin{equation}
\range{\sigma}{S} := \left( \left( \rp{p}{\sigma}{I} \right)_{S \supset I \in \mc I_p} \right)_{p=1,2} .
\end{equation}
\end{definition}

\noindent Similarly, we can talk about \defword{joint range} $(\rp{}{\sigma}{h})_{h\in S}$ and \defword{normalized joint range} $(\rp{}{\sigma}{h|S})_{h\in S}$.
Moreover, \cite{ReBeL} recently introduced the related notion of a \defword{public belief state}, which is essentially the pair formed by a public state and a range at that state.

The following theorem shows that considering ranges at \textit{public states} is not an ad-hoc choice, but in fact the canonical one.
It states that changing the reach probability of a single history might change the value of any other history within the same public state.
(Formally, the result uses \textit{common-knowledge} public states, which \cite{MCCR} defines as the \textit{smallest} subsets of $\mc H$ closed under infosets of both players.)

\begin{restatable}[Public state minimality]{theorem}{wholeRangeDep}\label{thm:wholeRangeDep}
Let $\trunk$ be a trunk, $\mc Z^\trunk$ its leaves, $S \subset \mc Z^\trunk$ a common-knowledge public state, and $h_0$, $g \in S$.
Suppose that trunk strategies $\trunkS$ and ${\mu}^\trunk$ render the same non-zero reach probabilities at $S$, except that $\rp{}{\trunkS}{g} \neq  \rp{}{\mu^\trunk}{g} = 0$.

Then there exists some game $G$, s.t. $\trunk$ is a trunk in $G$ and $\mc Z^\trunk$ the corresponding leaves, for which both $G(\trunkS)$ and $G(\mu^\trunk)$ each have a unique Nash equilibrium, $\sigma$ and $\mu$, and these satisfy $\hv{1}{\sigma}{h_0} \neq \hv{1}{\mu}{h_0}$.
\end{restatable}

\noindent While this result doesn't rule out more compact sufficient statistics for \textit{specific} games or trunk strategies, it does say that \textit{any} trunk is part of some game where anything less than data over whole public states will fail to be sufficient for \textit{many} strategies.
This yields the following corollary:

\begin{corollary}\label{cor:publ-st-minimality}
Ranges over anything less than (com\-mon-knowledge) public states might fail to be a sufficient statistic for computing (reachably-optimal or better) value functions.
\end{corollary}

\begin{figure}
\centering
\forestset{
    matrix_game/.style = {align=center,
        rectangle,
        draw,
        tier=MG
    }
}
\begin{forest}
[root, chance, s sep+=0.28\nodesize,
    [$h_0$,pl1,name=0 [1 0\\0 1, matrix_game ] ]
    [,pl1 [0] ]
    [$h_1$,pl1,name=1 [1 0\\0 1, matrix_game ] ]
    [$h_2$,chance,name=2 [1 0\\0 1, matrix_game ] ]
    [$h_3$,pl2,name=3 [1 0\\0 1, matrix_game ] ]
    [$h_4$,pl2,name=4 [1 0\\0 1, matrix_game ] ]
    [$g$,chance,name=5 [0\\1, matrix_game ] ]
]
\node [pl1_cl_infoset, fit=(\corners{0})(\corners{1})] {};
\node [infoset, augmented, fit=(2)(\corners{3})] {};
\node [infoset, augmented, fit=(\corners{4})(5)] {};
%
\node [infoset, opponent, augmented, fit=(\corners{1})(2),
    inner xsep = 0.1\nodesize] {};
\node [infoset, opponent, fit=(\corners{3})(\corners{4}),
    inner xsep = 0.1\nodesize] {};
\draw [line_infoset] (0!1) to (1!1);
\draw [line_infoset] (2!1) to (3!1);
\draw [line_infoset] (4!1) to (5!1);
\draw [line_infoset, opponent, bend right] (1!1.south) to (2!1.south);
\draw [line_infoset, opponent, bend right] (3!1.south) to (4!1.south);
\node[xshift=7em, yshift=-1em, align=right] {choose according to\\$\rp{}{\trunkS}{h}/P^{\trunkS}(S)$};
\end{forest}
\caption{An example of a game from Theorem~\ref{thm:wholeRangeDep}.}
\label{fig:wholeRangeDep}
\end{figure}

\paragraph{Output}

With the results obtained so far, we see that many value functions can be evaluated independently in each public state and values of all histories within each $S$ can be computed all at once.
In other words, we can view them as mappings of the type
\begin{equation*}
\vv : ( \textnormal{public state $S$} , \textnormal{range at $S$} ) \mapsto \textnormal{vector indexed by $h \in S$} .
\end{equation*}
However, public states typically have many more histories than information states (roughly quadratically), so the vector $\left( \vf{}{\textnormal{range}}{h} \right)_{h\in S}$ will have a large dimension.
Since many algorithms run on a per-infoset basis anyway (e.g., CFR does), we can aggregate the individual history values into infoset values $\Vf{p}{(\cdot)}{I}$, and view value functions as mappings
\begin{equation*}
\VV : ( \textnormal{public state $S$} , \textnormal{range at $S$} ) \mapsto \textnormal{vector indexed by $I \in \mc I$, $I\subset S$} , 
\end{equation*}
where $\Vf{}{(\cdot)}{I} = \sum_{h\in I} \rp{}{(\cdot)}{h|I}\vf{p}{(\cdot)}{h}$.
Depending on the specific application, we might replace the output by the format that is the most appropriate (e.g., counterfactual values of infosets for DL-CFR).
Since the resulting vectors will have a much lower dimension than the history-based representation, representing value functions in this manner makes them easier to approximate (e.g., by using a neural network).

\begin{remark}[Recovering values of histories]\label{rem:infoset-vals-to-h-vals}
One might argue that some algorithms are typically implemented over histories, so infoset-based value functions might fail to preserve enough information to run such algorithms.
However, many such algorithms in fact only depend values of infosets, rather than histories.
We can thus take the infoset values, $\Vf{p}{\trunkS}{I}$, and decompose them into some ``fake'' history values $(\hat{\mathbf{v}}^{\trunkS}(h))_{h\in I}$.
As long as we ensure that $\sum_{h\in I} \rp{}{\trunkS}{h|I}\hat{\mathbf{v}}^{(\cdot)}(h) = \Vf{p}{\trunkS}{I}$ holds for all infosets $\mc Z^\trunk \supset I \in \mc I_p$, these fake values will result in identical infoset values for $p$ in the whole trunk (Theorem~\ref{thm:infoset-v-characterization}).
This condition can be satisfied trivially by setting $\hat{\mathbf{v}}^{(\cdot)}(h_0) := \rp{}{\trunkS}{h_0|I}^{-1} \Vf{p}{\trunkS}{I}$ for an arbitrary cf. reachable $h_0 \in I$ and zeroing out the rest.
This is a useful trick that allows us to combine infoset-based value functions with history-based implementations (in our case, depth-limited CFR).
\end{remark}

\begin{remark}[Values of public belief state]\label{rem:PBS-values}
\cite[Theorem~1]{ReBeL} shows that infoset values coincide with supergradients of values of public belief states (which can be defined as $v^\sigma_p(S,r) = \sum_{S \supset I \in \mc I_p} \rp{}{\sigma}{I\mid S} \vf{p}{\sigma}{I}$).
This means that instead of computing values for all infosets for a single range, we also have the option to only compute a single value for the whole public state but do so for various different ranges, such that we can approximate the partial derivatives with respect to reach probabilities of individual infosets (as these correspond to infoset values).

Taken to the extreme, this result can be extended to an equivalence between values of individual infosets and a single value for the whole bottom part of the game.
(This is because the proof of Theorem~1 in \cite{ReBeL} does not rely on the minimality of public states and, therefore, goes through even when the public state spans the whole width of the game tree.)
In this paper, we chose to use the one-value-per-infoset representation because it seemed more suitable for discussing the differences between different types of optimality.
\end{remark}
\section{Value Functions in Partially-Observable Stochastic Games }\label{sec:fog}

In this section, we summarize the connection between the EFG model and the POSG model (traditionally used in multiagent RL).
In particular, we \edit{observe that all of our results apply to POSGs as well,
and that some of the existing POSG results are relevant (or even directly applicable) to value-functions in EFGs.
We also list the most relevant results from the POSG literature.}

Recall that the standard MARL model for zero-sum\footnote{While value-functions in other types of POSGSs --- e.g., in Dec-POMDPs \cite{oliehoek2006dec} --- are sometimes superficially similar to values in our setting, assumptions often made in these domains (e.g., joint training) give the resulting concepts vastly different properties. As such, their discussion would be outside of the scope of the present paper.} games is a partially-observable stochastic game (POSG), where players take actions which causes them to transition to a new state, obtain some reward, and receive some observation \cite{POSGs2004hansen}.
\edit{Another important difference between EFGs and POSGs is that while EFGs are tree-structured, POSGs can contain cycles.}

\edit{Reasoning about general POSGs is difficult because we soon encounter nested beliefs --- each player needing to maintain a belief over the current state of the game, a belief over the opponent's beliefs, a belief over the opponent's beliefs over the player's beliefs, and so on \cite{dermed2013value}.
One way of countering these difficulties is to work with a restricted class of POSGs.
This approach is certainly viable, and there are several works that analyze value functions in these settings \cite{cole2001dynamic,dermed2013value,horak_public,horak2017heuristic}.
However, nested beliefs are \textit{inherent} to many domains of interest (including poker),
so we also need some approach for tackling them.
This is where the connection between POSGs and EFGs becomes relevant.}

\edit{As shown in \cite{oliehoek2006dec,FOG}, EFGs and POSGs should not be viewed as two unrelated models.
Instead, EFGs are \textit{derived} objects that can be obtained by ``unrolling'' some underlying POSG, such that every POSG $G$ has a tree-structured EFG representation $E$.
While $E$ might be larger than the underlying POSG, it avoids nested beliefs by allowing players to use strategies defined on infosets (of which there are only finitely many) rather than on the difficult-to-handle nested beliefs.
As a result, many concepts used for analyzing POSGs, such as policies, their expected rewards, and solution concepts, are often defined on the EFG representation\footnote{
    \edit{Since the notation in this area does not seem to be settled, we should, strictly speaking, say that these concepts are defined on the EFG representation of $G$, \textit{or on some object with very similar properties}.}
}
of $G$ rather than on $G$ itself \cite{oliehoek2012decentralized}.
Most importantly for our work, this is also true of value functions \cite{wiggers2016structure,buffet2020bellman,delage2021hsvi},
which means that all of the present paper's results also apply to POSGs.}

\edit{In the opposite direction, some of the results from the POSG literature are also highly relevant for value functions in EFGs \cite{oliehoek2013sufficient,dibangoye2016optimally}.
A particularly relevant work is \cite{wiggers2016structure} which, translated into our terminology, studies the depth-limited expected-utility function $\left( \trunkS_1, \trunkS_2 \right) \mapsto \DLu{p}{\trunkS_1, \trunkS_2}$ (which the paper refers to as value function).
It shows that for any reachably-optimal value function $\vv$, $\DLu{p}{\, \cdot \,, \, \cdot \,}$ is concave-convex and has a unique max-min value.}

\edit{Finally, one downside of POSGs is that, by default, an EFG representation of a \textit{POSG} does not define (non-trivial) public states,
which are essential for localized computation of value functions (Proposition~\ref{prop:public-state-sufficiency}, Theorem~\ref{thm:wholeRangeDep}).
To reap the full benefit of the EFG representation, we can start with a minor extension of the POSG model, called factored-observation stochastic games (FOSGs) \cite{FOG}.
Indeed, each observation in a FOSG $G$ is explicitly decomposed into a private part and a public part, which naturally endows the corresponding EFG $E$ with a public partition --- each public state $S$ in $E$ can be identified with a sequence $\vec o_\textnormal{pub}$ of public observations in $G$.
Similarly, each information set can be identified with a pair $(\vec o_\textnormal{pub}, \vec o_{\textnormal{priv}(p)})$, where $\vec o_{\textnormal{priv}(p)}$ is a sequence of private observations (and actions) of player $p$.}

\edit{An important benefit of the FOSG-EFG correspondence is that identifying public states and information sets with sequences of observations gives them additional structure.
To the extent that similar game states (in terms of what the players observe) have similar values, this structure might allow for better generalization when approximating value functions by neural networks.
For this reason, all games in our experiments (Section~\ref{sec:empirical}) are initially specified using the FOSG model.}

\section{Experimental Evaluation}\label{sec:empirical}

    The goal of this section is to complement the theoretical contribution (Sections~\ref{sec:theory} and \ref{sec:fog}) by an analysis of the practical side of value functions --- approximating value functions by value networks and using them for depth-limited solving.
    In particular, we train a value network that would, in the limit of infinite resources, act as a counterfactually optimal value function (Definition~\ref{def:optimal-vf}) and use it in conjunction with a depth-limited version of CFR (Example~\ref{ex:DL-CFR}).
    We call the resulting algorithm \DLCFR{}, and describe it in detail in Section~\ref{sec:exp_setup_v2}.
    From results such as \cite{DeepStack,pog,Supremus,DeepRole}, we already know that approaches based on CFR and value functions can achieve impressive results.
    Instead of showing that the algorithm can solve large domains, our experiments thus focus on medium-sized domains where it is possible to thoroughly investigate the algorithm's behaviour and compute exact exploitabilities.
    We consider three domains with qualitatively different properties:
    Leduc hold'em poker (LH), a standard benchmark for imperfect information games,
    and (similarly-sized variants of) imperfect information goofspiel (GS) and imperfect information oshi-zumo (OZ).
    
    We start by describing the experimental setup and a universal encoding that avoids the need for hand-crafting domain features (Section~\ref{sec:exp_setup_v2}).
    Afterwards, we investigate several questions:
    First of all, since most past results only evaluated CFR and its variants on poker and poker-like domains (which have quite specific properties) \cite{PSCFR}, we should verify whether the depth-limited CFR approach empirically works beyond poker.
    This turns out to be the case --- as a by-product of our other experiments, we see that it is possible to train value networks that achieve low validation loss in all three domains and that coupling these value networks with \DLCFR{} produces strategies with low exploitability (Section~\ref{sec:sub:loss-vs_expl}).
    Second, we study the impact of the value network's quality on the strength of the resulting depth-limited solver, concluding that \DLCFR{} with a near-optimal value network successfully finds an $\epsilon$-Nash equilibrium strategy (Section~\ref{sec:sub:loss-vs_expl}).
    In particular, validation Huber loss of 0.001 was sufficient to achieve exploitability below $0.01$.
    Third, we look at how well the network generalizes to previously unseen situations (Section~\ref{sec:sub:generalization}).
    We observe that \DLCFR{} achieves a low exploitability despite encountering inputs significantly different from those present in the training data.
    Moreover, we show that the value network generalizes to unseen public states when there is enough public states to generalize from, suggesting that the FOSG encoding is suitable for value networks.
    Finally, since \DLCFR{} with a perfect value network would behave identically to CFR-D \cite{CFR-D}, we qualitatively compare the two algorithms (Section~\ref{sec:sub:cfrd-comparison}).
    We see that the value network achieves low loss on the inputs encountered in CFR-D and that both algorithms produce similar strategies on the tested domains. 
    
    In summary, our experiments demonstrate that the value function approach to depth-limited solving is viable in domains other than poker and that value functions can be approximated using a universal, domain-independent encoding and architecture.

\subsection{Experimental Setup}\label{sec:exp_setup_v2}

We now describe the algorithm used for depth-limited solving, its value network, the experimental domains, the process used for generating the corresponding training data, and the encoding of this data.
    
    \subsubsection{Depth-limited Solver}
    We use \DLCFR{} --- a depth-limited version of CFR+ (Example~\ref{ex:DL-CFR}, \cite{tammelin2014cfr+}) coupled with a value network.
    Each iteration, the algorithm computes the new strategy in the trunk and collects the ranges for all public states at the depth limit.
    Each public state $S$ and range is encoded (in a way described below) and used as a query to the value network, which outputs a vector of values, one for each infoset in $S$.
    Since our implementation of CFR+ runs on histories rather than on infosets, we then use the simple trick from Remark~\ref{rem:infoset-vals-to-h-vals} to convert the infoset values to history values.
    These values are then back-propagated by CFR+ as usual.
    When using \DLCFR{}, we always run it for 1\,000 iterations.
    The implementation of \DLCFR is available online at \cite{gtlib2}.
    
    \subsubsection{Training Data}\label{sec:sub:training-data}
    The training data is produced by the following process.
    First, we generate a random trunk strategy at each infoset in the trunk by choosing either a fully mixed strategy drawn from a uniform probability distribution 90\% of the time or a pure strategy for a randomly chosen action 10\% of the time.
    We then fix this trunk strategy and solve the bottom of the game using 1\,000 iterations of CFR+.
    In Section \ref{sec:opt_vfs:cf}, we referred to this technique as \defword{value solving} and showed that it produces counterfactually (near-)optimal values, so \DLCFR{} should -- with a perfect network -- converge to a Nash equilibrium (by Proposition~\ref{prop:enable-DL-CFR}).
    Finally, we create a single training data point out of each public state $S$ at the depth-limit --- the input is formed by $S$ and the corresponding range at $S$ (Definition~\ref{def:range}), the output is formed by the vector of counterfactual infoset-values (Section~\ref{sec:sub:cfvs}).
    (We explain this encoding in more detail, but only after describing the experimental domains.)
    As a result, each trunk strategy yields as many training data points as there are infosets at the depth limit (see Table~\ref{tab:domains_description2}).
    We use 90\% of the data for training and 10\% for validation.
    
    Since obtaining training data can be costly,
    we investigated what is the minimum amount of training data needed for satisfactory performance on unseen inputs.
    We used training sets of increasing size from 1\,000 up 50\,000 (step size 1\,000) depending on the domain, and computed the validation loss on a fixed data set.
    In all three domains, between below 10\,000 samples were sufficient for achieving a validation Huber loss of 0.001. We however opted to use a larger amount of data to run our main experiments on to analyze \DLCFR{} using a well-trained value network (exact numbers in Table~\ref{table:finalarch2}).
    Note that the necessary amount of data might grow significantly with increasing domain size (for example, DeepStack \cite{DeepStack} used $10$ million random trunk strategies).
    In such larger domains, it might, therefore, be advantageous to use a combination of
        (a) a more involved training process (such as the one used in \cite{ReBeL}) that puts more weight on inputs that are more likely to be relevant for the algorithm and
        (b) an approach like continual resolving \cite{DeepStack,MCCR}, which splits the game at multiple depths.

    \subsubsection{Value Network}
    All of the networks use a standard fully-connected feed-forward network, rectified linear units for the hidden layers and linear activation for the output layer, 3-6 layers, and layer-width 5-10x the input size (see Table~\ref{table:finalarch2}).
    They were trained for 1\,000 epochs by using the Adam optimizer \cite{adam} while minimizing Huber loss.
    The reasoning behind the choice of loss function is further explored in Section~\ref{sec:sub:loss-vs_expl}.
    Before settling on these choices, we experimented with various other options and hyperparameter values; for more details, see Appendix~\ref{app:sec:loss} and \ref{sec:app:hyperparams}.

    \begin{table}[ht]
        \begin{center}
         \begin{tabular}{|c c c c c|} 
         \hline
         Domain & Layers & Neurons & Training Strategies & Tr. Datapoints \\ 
         \hline\hline
         GS & 5 & 500 & 2\,000 & 18\,000 \\ 
         \hline
         OZ & 4 & 400 &  2\,000 & 34\,000 \\
         \hline
         LH & 6 & 200 & 812 & 70\,644\\
         \hline
        \end{tabular}
        \caption{The final value network architectures and the amounts of training data.
        Recall that each trunk strategy yields as many training samples as there are public states at the depth limit.}
        \label{table:finalarch2}
        \end{center}
    \end{table}
    
    \begin{table}[ht]
        \centering
        \begin{tabular}{|c|c|c|c|}
        \hline
             Game Property & Goofspiel & Oshi-zumo & Leduc hold'em\\
             \hline
             private actions & Yes & Yes & No \\
             early terminal nodes & No & Yes & Yes\\
             constant-size infosets & No & No & Yes\\
             rounds & 5 & 8 & 4\\
             total infosets & 5\,000 & 1\,600 & 4\,000\\
             total histories & 56\,000 & 20\,000 & 61\,000\\
             depth-limit (rounds) & 3 & 4 & 2 \\
             publ. states at depth-limit & 9 & 17 & 87 \\
             \hline
        \end{tabular}
        \caption{
            Properties of the domains used for empirical evaluation.
        }
        \label{tab:domains_description2}
    \end{table}

    \subsubsection{Experimental Domains}
    
    We now give a high-level description of the domains used in the experiments.
    For a more detailed description, see Appendix~\ref{sec:app:domains}.
    The first domain we use is Leduc hold'em (LH) \cite{southey2012bayes} --- a standard small variant of poker with a deck of six cards, two rounds of betting, and fixed bet sizes.
    Since poker is an imperfect information game with rather specific properties \cite{PSCFR}, we also use two other domains which are qualitatively different from it (and from each other).
    One of these games is imperfect information oshi-zumo (OZ) with board size 3 and a budget size 8.
    In this game, each player controls a wrestler and spends some amount of energy (technically called ``coins'') each round in an attempt to pushing the opponent towards the edge of the board.
    In the imperfect information variant of the game, the players learn which of them spent more each round, but not what the exact amount was.
    The last domain we use is a 5-card variant of imperfect information goofspiel (GS).
    In this game, a deck of cards $\{1, \dots, 5\}$ is auctioned off card by card, with each player trying to maximize the sum of card-values they win.
    However, instead of money, each player has their own deck $\{1, \dots, 5\}$ that they for betting.
    The different properties of all three domains are summarized in Table~\ref{tab:domains_description2}.
    For the purpose of depth-limited solving, we split each of the domains in a trunk and bottom after half of the rounds has passed.
    When calculating exploitability, we normalize all utilities to make the results comparable.

    \subsubsection{Domain-independent Encoding}\label{sec:sub:encoding}
    
    On the mathematical level, each input to the network is a pair, a public state $S$ and a range $r$ at $S$ (a vector of reach probabilities, one for each infoset at $S$; Definition~\ref{def:range}) and the corresponding output is a vector of counterfactual values (with the same shape as $r$). We now explain how \DLCFR{} encodes these objects when communicating with its value network.
    The FOSG representation identifies each public state $S$ with a sequence of public observations $\vec o_{\textnormal{pub}} = (o_1, \dots, o_n)$, where each $o_i$ belongs to some set of possible observations.
    Since this set is discrete and rather small in all of our domains (and in many others), we can enumerate it, use a one-hot vector to encode each $o_i$, and concatenate these vectors to obtain an encoding of the full sequence $(o_1, \dots, o_n)$.
    For example, there are three possible public outcomes of each round of goofspiel, depending on whether the currently-auctioned card is given either to player one, player two, or discarded (when the players make identical bids),
    so the corresponding observations would be encoded as $[1, 0, 0]$, $[0, 1, 0$], resp. $[0, 0, 1]$.
    If player one wins the first round, loses the second round, and the players are currently bidding for the third card, the corresponding public state would be encoded as $[1, 0, 0, 0, 0, 1]$.
    
    The encoding of ranges also relies on the FOSG representation.
    In a FOSG,
        each infoset $I \in \mc I_p$ is identified with a pair $(\vec o_\pub, \vec o_\priv )$,
        where $\vec o_\pub$ corresponds to the public set $S$ that contains $I$
        and $\vec o_\priv$ is some sequence of $p$'s private observations and actions.
    Since the publicly observable actions (e.g., bets in Leduc hold'em) are already a part of $\vec o_\pub$,
        we can assume that $\vec o_\priv$ only contains $p$'s private observations and their non-public actions (e.g., bid sizes in imperfect-information goofspiel and oshi-zumo).
    The key insight is that \textit{in the context of a specific public state}, we only need the private sequence $\vec o_\priv$ to specify an infoset.
    Consequently, we construct a universal encoding that works \textit{all} public states,
        by fixing some enumeration of the set of all sequences $\vec o_{\textnormal{priv(p)}}$ that can appear in \textit{some} public state at the depth limit.
    (For example,
        in our variant of goofspiel, there are no private observations but all actions --- i.e., using up one of the cards $\{1, \dots, 5\}$ --- are private.
        The set of all possible private sequences $\vec o_\priv$ at the start of the third round can, therefore, be identified with $\{ (c_1, c_2) \mid c_i = 1, \dots, 5, \, c_1 \neq c_2 \}$.)
    With this enumeration, the $p$'s part of a range $r$ at a public state $S$ can be encoded as a vector $(r_1, \dots r_n)$, where $r_i$ is the reach probability of the infoset that corresponds to the $i$-th private sequence $\vec o_\priv$.
    If this $\vec o_\priv$ does not correspond to any infoset at the given public state, we set $r_i$ to zero.
    (For example, our goofspiel scenario $S$ where player one wins the first round and looses the second is incompatible with any sequence $\vec o_\priv$ where they bid $1$ -- smallest possible amount -- in the first round or $5$ -- largest possible amount -- in the second round.)
    To encode the full range, we concatenate the part of player 1 with the part of player 2.
    Since counterfactual values have the same structure as ranges, we use the same method for their encoding (or rather, decoding).
    Overall, a major contribution of this encoding is that it works for \textit{any} domain described as a FOSG.
    Its main limitation (which we leave for future work) is that the width of the network's input layer scales with the number of private observation sequences.
    Fortunately, this number is manageable in the majority of games studied in the CFR literature so far --- e.g., in all variants of poker including no-limit Texas hold'em \cite{DeepStack}, in liar's dice \cite{burch2012efficient}, and in all domains considered in this paper.

\subsection{Value Network's Impact on Depth-limited Solving}\label{sec:sub:loss-vs_expl}

While we ultimately care about the exploitability of \DLCFR{}'s strategy, computing it is costly (Proposition~\ref{prop:trunk_expl}), so we would like to know whether the value network's quality is a reliable proxy for exploitability and if so, how low validation loss should we aim for.

Before analyzing the main question, we first look at a one aspect value network training that hasn't been addressed by previous work:
which loss function should we use for training and validation (the main candidates being Huber, $l_1$, and $l_\infty$ \cite{DeepStack}).
The details are presented in Appendix~\ref{app:sec:loss}.
In summary, for each domain and loss function combination, we trained value networks to minimize the given loss while calculating the validation loss for \textit{all three} loss functions.
We observed that the distinction between the losses is not significant.
We therefore choose Huber as it has been used in prior work \cite{DeepStack}.
We also performed an in-depth analysis of custom loss functions (Appendix~\ref{app:sec:loss}) before concluding that standard losses are sufficient.
As a result, we train all value networks by minimizing the Huber loss while measuring all three losses (Huber, $l_1$, $l_\infty$) on validation data and reporting the one that seems the most intuitive for the given experiment.

\begin{figure}[tb]
    \centering
    \includegraphics[width=\linewidth]{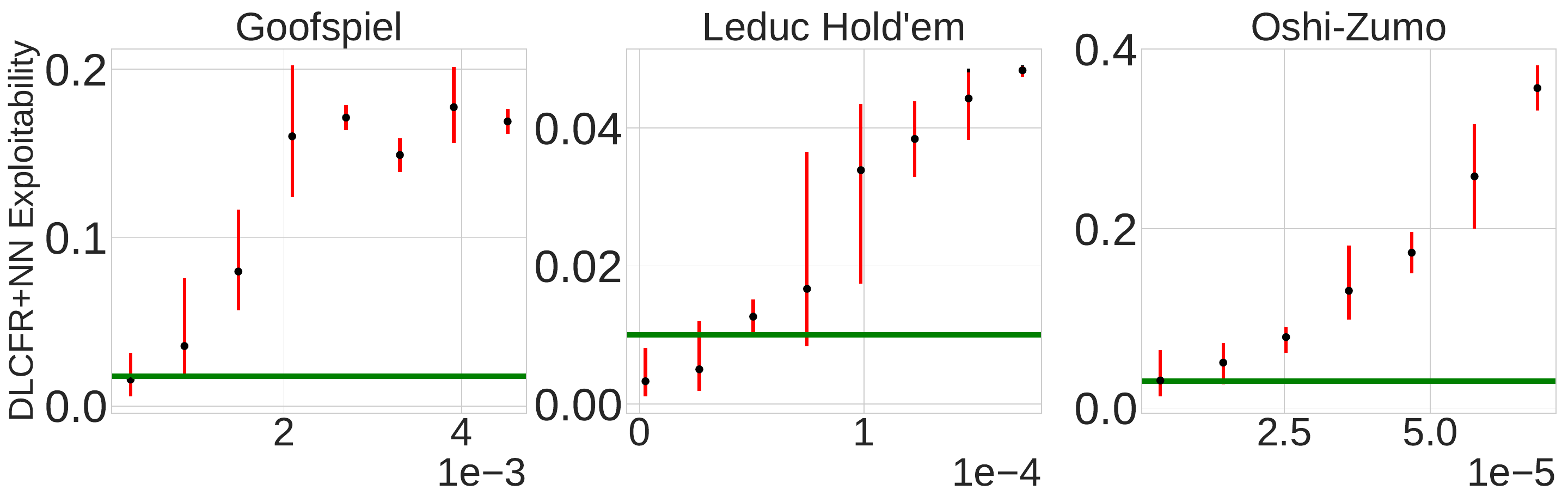}
    \caption{The relationships between the value network's Huber validation error (x-axis) and on the exploitability of \DLCFR{}.
    The figures display aggregate data from training runs with 100 different initializations.
    We grouped the data into eight buckets based on the validation loss.
    For each bucket, we display the interval between minimum and maximum exploitability (the red line) and the average exploitability (black dot) achieved with the corresponding value networks.
    The green line indicates the exploitability of CFR+ (1\,000 iterations, same as \DLCFR{}).
    The exploitabilities of \DLCFR{} with a constant value function (0.5 in GS, 0.7 in OZ, 0.2 in LH) are not shown as they would not fit into the graph.
    }
    \label{fig:expl_vs_error2}
\end{figure}

To investigate the relationship between network loss and exploitability, we ran 100 training runs of the value network (with different initializations) and continually saved checkpoints of the current weights and the corresponding validation losses.
We used each checkpoint for \DLCFR{} (1\,000 iterations), measured the exploitability of the resulting strategy, and plotted the aggregate data in Figure~\ref{fig:expl_vs_error2}.
As a baseline, we computed the exploitability of the standard CFR+ after 1\,000 iterations.
As a sanity check, we measured the exploitability of \DLCFR{} with a constant value function (that ignores the input and predicts value $0$ for every infoset) --- this always resulted in high exploitability (0.5 in GS, 0.7 in OZ, 0.2 in LH).
In all cases, we see a strong connection between validation loss and the resulting exploitability.
The networks are able to reach training Huber loss below 0.001 and exploitability below 0.01 --- in other words, \DLCFR{} performs on par with CFR+ that runs on the full tree.

\subsection{Generalizing to Unseen Inputs}\label{sec:sub:generalization}
To ensure that the value network isn't merely memorizing all data, we investigate its ability to generalize to unseen situations.
Recall that since the validation losses are low, the network must be able to do \textit{some} amount of generalization.
Moreover, as shown in Appendix~\ref{sec:app:val_cfrd}, the validation error remains low (in the sense of corresponding to an acceptable exploitability according to Figure~\ref{fig:expl_vs_error2}) even when we validate on inputs that would appear in CFR-D (i.e., on inputs \DLCFR{} would ask for if its value network was perfect).
However, we can also look at the network's generalization ability in more detail.

\begin{figure}[t]
    \centering
    \includegraphics[width=\linewidth]{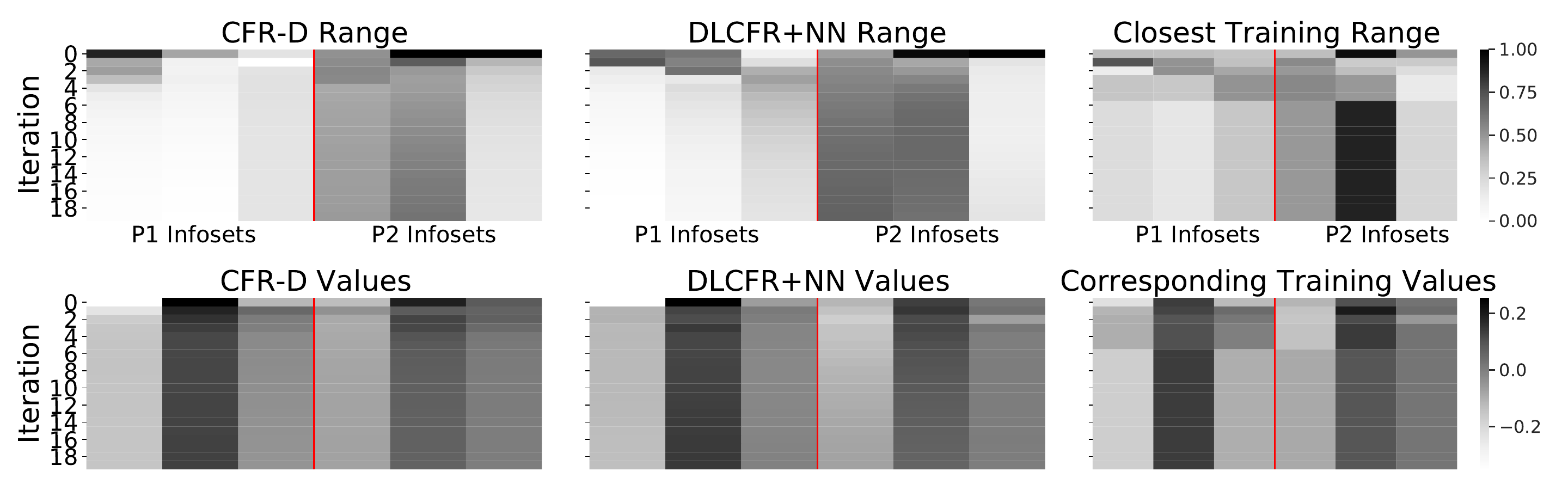}
    \caption{
        Qualitative comparison of CFR-D, \DLCFR{}, and training data in a specific public state in Leduc hold'em.
        Each cell corresponds to a reach probability (resp. counterfactual value) of one infoset in the given iteration.
        The values in the left column depict the values computed by CFR-D's resolving,
        \DLCFR{} value network's predictions are in the middle,
        and the training data points closest to the \DLCFR{} data are on the right.
    }
    \label{fig:heatmap_main2}
\end{figure}

First, we investigate how different the training ranges (Definition~\ref{def:range}) are from those used by \DLCFR{}.
To do so, we look at a single run of the algorithm in each domain.
Focusing on a single public state, we look at each input requested by \DLCFR{} and compare it with the training data point that is closest to it in terms of Euclidean distance
(see the middle and right columns of Figure~\ref{fig:heatmap_main2}, resp. Appendix~\ref{sec:app:heatmaps}, for a visualization of the data in LH, resp. GS and OZ).
In all cases, we see that even the closest training data points are significantly different from those requested by \DLCFR{}.

Second, we evaluate the value network's ability to generalize to parts of the game not encountered in the training.
We do this by withholding all data about a single public state from the training data, training the network on the remaining data, and computing validation loss on the withheld data.
We repeat this calculation for all public states at the depth limit, and report the results for Leduc hold'em in Figure~\ref{fig:pubstatecrossval_main2} and for the other domains in Appendix~\ref{sec:app:pubstatecrossval}.
The generalization is very poor in goofspiel, reasonable in oshi-zumo, and very strong in Leduc hold'em.
Since these domains have 9, 17, resp. 87 public states at the depth limit, this is suggests (though isn't a hard proof) that the encoding is suitable for value networks --- at least in our domains, it allows for generalization to the extent that there is enough data to generalize from.
\begin{figure}[tb]
        \centering
        \includegraphics[width=0.8\linewidth]{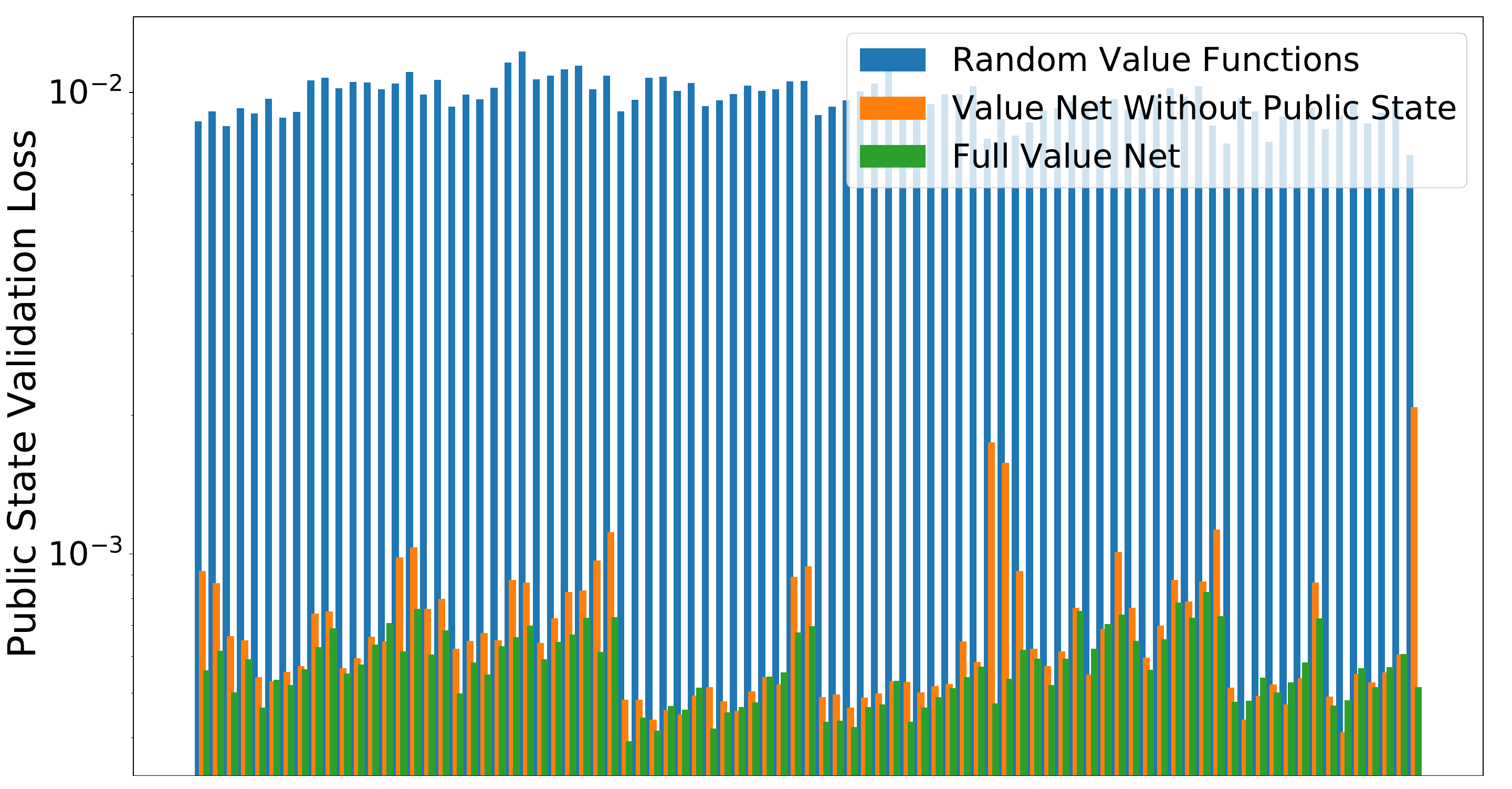}
        \caption{
            We show $l_\infty$ validation errors over withheld public states in Leduc hold'em.
            Each orange bar corresponds to the $l_\infty$ validation loss of a value network trained on all data except for the public state $S$ with given index.
            For comparison, we show the loss (at $S$) of the value network trained on all data (green)
                and the average loss (at $S$) of 10 randomly initialized value networks.
        }
        \label{fig:pubstatecrossval_main2}
    \end{figure}

\subsection{Qualitative Comparison to CFR-D}\label{sec:sub:cfrd-comparison}
    With a perfect value network, \DLCFR{} would behave identically to CFR-D.
    As our last experiment, we investigate whether this similarity also appears in practice.
    We run CFR-D on all three domains, record the corresponding ranges of public states at the bottom of the trunk, and compare them to ranges that appear in \DLCFR{}. Additionally, we show the corresponding values (subgame solutions) which have been computed by both algorithms in the bottom row. 
    The results for Leduc hold'em poker are depicted in the left and middle columns of Figure~\ref{fig:heatmap_main2};
    the data for the other two domains can be found in Appendix~\ref{sec:app:heatmaps}.
    We see that -- at least in all of our domains -- the two algorithms encounter very similar ranges and converge to similar strategies.

\section{Discussion}\label{sec:discussion}
    In this section, we summarize the most-relevant conceptual and theoretical results from related work and present our conclusions.

\subsection{Related Work}\label{sec:related_work}

\edit{The two results that inspired the paper the most are CFR-D \cite{CFR-D} and DeepStack \cite{DeepStack}.
Roughly speaking, a portion of this paper might be viewed as a domain- and algorithm- independent extension of CFR-D and DeepStack (minus the parts about continual resolving and evaluating on a large domain):}

\edit{In CFR-D, the letter D stands for ``decomposition''.
However, the algorithm can be understood as a form of depth-limited solving, first that we are aware of in imperfect-information games.
The key differences between CFR-D and the present paper are:
(1) CFR-D always solves the trunk using CFR.
(2) Upon reaching the bottom of the trunk, CFR-D solves the subgames by CFR or some other game-solving algorithm (instead of evaluating them using a value function).
(3) CFR-D is only evaluated on Leduc hold'em.
The paper uses the notion of trunk and subgames but does not define them formally.
It also points out that for CFR-D to work, the subgame-strategies need to be mutual \textit{counterfactual} best responses, not merely mutual best responses (i.e., Nash equilibria).
This observation inspired us to differentiate between reachable and counterfactual optimality and prompted our search for (counter-) Example~\ref{ex:DL-CFR-fail}.}

\edit{DeepStack has three relevant contributions:
(1) it builds on top of CFR-D, but rather than solving subgames by CFR, it trains a value network to imitate CFR's behaviour on subgames.
(2) It introduces continual resolving --- iteratively applying depth-limited solving in online play, and thus splitting the game on multiple depth levels.
(3) It demonstrates that with (1), (2), and certain poker-specific tricks, it is possible to achieve superhuman performance in full-sized poker (two-player no-limit Texas hold 'em).
The key limitation of DeepStack is that both the value-network architecture and the overall algorithm is specific to poker and CFR.
Additionally, due to imitating CFR-D, the value-network's training target is only defined in terms of CFR.}

\edit{A second related line of work, which seemingly does not have anything to do with value functions, revolves around multivalued-states \cite{brown2018depth}:
Where \cite{DeepStack} replaces each trunk-leaf history by a call to a value function, \cite{brown2018depth} replaces each trunk-leaf by a single decision of which strategy to use for the remainder of the game.
In Section~\ref{sec:noam}, we show that multivalued states can be understood as a computationally-convenient implementation of a particular type of an optimal value function (Theorem~\ref{thm:portfolio-game}).}

\edit{A third line of closely related work is the study of value functions in partially-observable stochastic games, discussed further in Section~\ref{sec:fog}.
Two results worth highlighting here are:
(1) By \cite{FOG, oliehoek2006dec}, the value functions studied in the present paper also apply to POSGs.
(2) The paper \cite{wiggers2016structure} proves that (what we would call) reachably-optimal values aggregated over the whole bottom of the trunk are concave-convex and have a unique max-min value.}

\edit{In summary, the value-function approach \cite{DeepStack} and the multivalued-states approach \cite{brown2018depth} are two ways of obtaining a similar result and can likely be combined.
Both of these can be applied to POSGs (and FOSGs \cite{FOG}), and some of the POSG results have implications for both approaches.}

\edit{There are also several recent relevant results (which only came out after this paper has been written, or around the same time).
The papers \cite{zhang2020small,zhang2021finding} extend alpha-beta pruning to imperfect-information games, and some of its definitions are relevant to our Section~\ref{sec:DL_games}.
The paper introducing ReBeL \cite{ReBeL} shows that infoset-values can be viewed as derivatives of values aggregated over public states and describes an online value-network training procedure similar to AlphaZero \cite{AlphaZero}.
Similarly, Player of Games \cite{pog} is an extension of AlphaZero and DeepStack \cite{DeepStack} which performs competitively in both perfect- and imperfect-information games (chess, Go, poker, and stratego).}

\edit{Finally, as the terminology suggests, the values considered in this paper are connected to the $V$- and $Q$-values used in (multiagent) (reinforcement) learning (MARL).
However, since both our setting and typical MARL settings need to be significantly simplified to overlap, the results in the two lines of work do not inform each other as much.
Indeed, the frequent differences are:
    (a) MARL often assumes general-sum rewards or $N\geq 3$ players,
    (b) we assume a finite-horizon, no discount factor, and rewards only in terminal states,
    (c) MARL often assumes simultaneous moves but otherwise perfect information (i.e., Markov games \cite{littman1994markov}), and hence uses stationary strategies.
However, where the two settings intersect, the defined value functions coincide.
For example, in MDPs (with a finite horizon, no discount factor, rewards in terminal states only, and a tree-structured state-space), universally-optimal value functions coincide with $v^*(s)$ \cite{sutton2018reinforcement}.
Similarly, universally-optimal value functions in two-player zero-sum Markov games (with the same qualifications) coincide with the minimax state values \cite{littman1994markov} and can be used to define the $Q$-functions used in \cite{NashQ, CEQ}.}
    \subsection{Conclusion}\label{sec:conclusion}

In this paper, we gave an accessible description of basic notions used in the CFR literature (Section~\ref{sec:background:values}) and introduced a number of concepts that enable reasoning about depth-limited games and value functions (Section~\ref{sec:DL_games}).
We proved that different degree of value-function optimality is required for different calculations and explained how to obtain the key types of value functions (Sections~\ref{sec:opt_vfs:reach}-\ref{sec:opt_vfs:univ}).
Additionally, we proved that public belief states provide the necessary and sufficient context for computing value functions (Theorem~\ref{thm:wholeRangeDep}).
Our description allows viewing Deepstack's value functions \cite{DeepStack} and Brown~et al.'s multivalued states \cite{brown2018depth} as two instances of a single unifying framework (Theorem~\ref{thm:portfolio-game}).
Moreover, the results also apply to partially-observable stochastic games and their recent extension, factored-observation stochastic games \cite{FOG} (Section~\ref{sec:fog}).
The theory shows that depth-limited solving is applicable to arbitrary domains and various algorithms. However, due to its success in recent years, our experimental evaluation focused on CFR.
We showed that adopting the FOSG formalism allows for a simple domain-independent encoding
which can be used for input and output of a value function (Section~\ref{sec:sub:encoding}).
We demonstrated the suitability of this encoding by showing that the resulting value network can generalize to unseen public states (Section~\ref{sec:sub:generalization}).
In three distinct domains, we used this encoding to train a simple feed-forward neural network that approximates an optimal value function.
We then implemented a depth-limited version of CFR that utilizes this network.
We performed an extensive experimental evaluation of this setup.
Most importantly, we confirmed that the value network's error is strongly correlated with the exploitability of the strategy found by the corresponding \DLCFR{} (Figure~\ref{fig:expl_vs_error2}),
    achieving performance that is as good as that of CFR with access to the full game.

Overall, we have shown that depth-limited solving is a viable and robust option for a range of imperfect-information games beyond poker.


%

\section*{Acknowledgements}
This work was both supported by the Czech science foundation grant no. 18-27483Y
and RCI grant CZ.02.1.01/0.0/0.0/16 019/0000765.


\clearpage
\appendix
\section{Non-Uniqueness of Value Functions}\label{sec:uniqueness}

Having explained how to compute and apply optimal value functions, it remains to ask about their uniqueness properties.
Unfortunately, the answers we offer in this section are often negative:
    there might be value functions that are all optimal in the same sense, yet prescribe different values.\footnote{Despite this, the fact remains that using any such value function will lead to good outcomes; the complication lies in the need to be consistent in this choice \cite{SoundSearch}.}
We will see that if one aims to use value functions in conjunction with function-approximation techniques, the natural follow-up question is whether the set of optimal value functions is at least convex.
We were unable to establish a conclusive result in either direction, but we have found some evidence that the answer might be positive.
The remainder of this section is devoted to giving further details and counterexamples on this topic.

\begin{proposition}[Optimal values aren't unique]
There exists $G$ and $\trunk$ for which there are multiple distinct universally-optimal value functions.
This holds even after infoset-aggregation $\Vf{p}{\trunkS}{I} := \sum_{h\in I} \rp{}{\trunkS}{h|I}\vf{p}{\trunkS}{h}$.
\end{proposition}

\noindent\textit{Proof.}
The sequential representation of matching pennies demonstrates this well.
\begin{wrapfigure}{r}{0.29\linewidth}
\vspace{-3mm}
\begin{forest}
[$\emptyset$,pl1, l sep-=0.5\nodesize,
        [H,pl2,name=a,
                [0] [1]
        ]
        [T,pl2,name=b,
                [1]
                [0]
        ]
]
\begin{pgfonlayer}{bg}
\draw [line_infoset, opponent, bend right] (!1) to (!2);
\end{pgfonlayer}
\node [infoset, augmented, fit=(\corners{a})] {};
\node [infoset, augmented, fit=(\corners{b})] {};
\end{forest}
\vspace{-5mm}
\end{wrapfigure}
Suppose the trunk $\trunk$ only consists of the root node where player 1 makes their decision and let $\trunkS$ be the uniform strategy.
Given the uniform trunk strategy, \textit{any} strategy that player 2 selects in the bottom of the game is going to extend $\trunkS$ optimally (even universally-optimally).
However, different player 2 strategies will result in different tradeoffs between $\hv{p}{(\cdot)}{\textnormal{heads}}$ and $\hv{p}{(\cdot)}{\textnormal{tails}}$, thus proving the first part of the proposition.
Since the singletons $\{\textnormal{heads}\}$ and $\{\textnormal{tails}\}$ each constitute a single infoset of player one, the infoset values are not unique either.\footnote{One might rightfully complain that this counterexample is trivial. However, if optimal values aren't unique in such a simple case, we should expect the situation to be even worse in more complicated games. In particular, we could easily construct a game where player one has ambiguous infoset values even for infosets where they act.}
\qed

\medskip

\noindent In light of Remark~\ref{rem:PBS-values}, this non-uniqueness should come as no surprise: \cite[Theorem 1]{ReBeL} establishes that infoset values correspond to supergradients of public belief state (PBS) values.
But since the PBS value function is concave but not necessarily differentiable, the supergradients might not be unique.
As a result, infoset values do not have to be unique either.

Some hope for infoset-value uniqueness stems from the fact that for a given trunk strategy $\trunkS$ and opponent strategy $\sigma_{-p} \supset \trunkS_{-p}$, all optimal responses $\sigma_p \supset \trunkS_p$ must lead to same values infoset values of $\mc Z^\trunk \supset I \in \mc I_p$ (trivially, since if $\Iv{p}{\sigma}{I}$ is equal to $\max_{\tilde \sigma_p \in \Sigma_p} \Iv{p}{\tilde \sigma_p, \sigma_{-p}}{I}$, it can only depend on $\sigma_{-p}$).
If one player has only a single optimal extension of the trunk strategy, the other player will thus have unique optimal infoset values.

Given that optimal values aren't unique, we might instead ask whether the corresponding set of functions is convex.
To see why this would be relevant, suppose that one half of a training set corresponds to an optimal value function $\vv$ and the other half to an optimal value function $\vv' \neq \vv$. Since a neural network trained on this dataset is likely to converge to $\frac{1}{2}(\vv + \vv')$, it would be advantageous if this value function was likewise optimal.
Since we so far failed to either prove or disprove this question, we pose it as an open problem for future work:

\begin{problem}\label{prob:VFs-convex?}
Is the set of (counterfactually) optimal value functions convex?
\end{problem}

\noindent We were able to prove a related result regarding the corresponding set of optimal extensions:

\begin{restatable}{proposition}{OEnonConvex}\label{prop:OE_isnt_convex}
The set $\OptE{\trunkS}{\textnormal{reach.}}$ of reachably-optimal extensions is always convex.
The same might not hold for $\OptE{\trunkS}{\textnormal{cf.}}$ and $\OptE{\trunkS}{\textnormal{univ.}}$.
\end{restatable}


\begin{figure}[ht]
\centering
\begin{forest}
[$r$,pl1,name=r
    [$c_1$,chance,name=L
        [$c_3$,chance,name=L2
            [$h'$,pl2,name=h2,tier=H
                [1][1]
            ]
        ]
    ]
    [$c_2$,chance,name=R
        [$g$,pl1,name=g
            [$h$,pl2,name=h,tier=H
                [1][-1]
            ]
            [1]
        ]
        [$g'$,pl1,name=g2
            [1][0]    
        ]
    ]
]
\node [pl1_cl_infoset, fit=(\corners{g})(\corners{g2})] {I};
\node [pl2_cl_infoset, fit=(\corners{h})(\corners{h2})] {J};
\node at ($(L)!0.5!(R)!0.2!(r)$) {$\trunk$};
\node at ($(L)!0.5!(r)+(-0.6em,0.6em)$) {1};
\node at ($(R)!0.5!(r)+(+0.6em,0.6em)$) {0};
\draw[darkgray,dashed,line width = 1.5pt] ($(L.south)+(-0.55\nodesize,-0.2\nodesize)$)--($(R.south)+(1.3\nodesize,-0.2\nodesize)$);
\end{forest}
\caption{A game with non-convex set of optimal extensions of $\trunkS$. Circles denote chance nodes with uniform strategy. The ``direction'' of the triangles denotes whether the node belongs to a maximizer (tip up) or minimizer (tip down).}
\label{fig:non_convex}
\end{figure}

\begin{proof}
The reachably-optimal part of the proposition follows from the fact that reachably-optimal extensions of $\trunkS$ are precisely the Nash equilibria of $G(\trunkS)$ (see the proof of Proposition~\ref{prop:comp-reach-optimal}) and the set of Nash equilibria is always convex.

We now construct a counterexample for the negative part of the proposition.
Let $G$ be the game from Figure~\ref{fig:non_convex}, where all chance strategies are uniform, $\trunk $ consists of the topmost three nodes $r$, $c_1$, and $c_2$, and strategy $\trunkS$ of player one (the maximizer) at $r$ is to always play left.

Firstly, note that independently of how we extend $\trunkS$ into a $\sigma$ in the whole game (i.e., by defining $\sigma_1(I)$ and $\sigma_2(J)$), we will necessarily have $\Iv{2}{\sigma}{J} = 1$.\footnote{This is assuming that both players share utility, with player one/two trying to maximize/minimize it.} Indeed, this is because $\Iv{2}{\sigma}{J}$ is the average of $\hv{2}{\sigma}{h'}$ and $\hv{2}{\sigma}{h}$, weighted by the reach probabilities of $h$ and $h'$, and the first of these probabilities is always zero.
This is \edit{the lowest value (or, rather, the only value)} that player 2 can get (given player one's strategy at $r$)
It follows that to see whether $\sigma$ is an optimal extension, we only need to verify the maximality condition for player one at $I$ (i.e., verify that $\Iv{1}{\sigma_1,\sigma_2}{I} = \max_{\rho_1|_I \in \Sigma_1} \Iv{1}{\rho_1|_I,\sigma_2}{I}$).
\edit{If the condition holds, the extension is universally optimal.
If it doesn't hold, the extension will be reachably optimal (since $I$ isn't reachable) but not counterfactually optimal (since $I$ \textit{is} counterfactually reachable).}

First, note that no matter which strategy player two picks, we always have
\begin{align*}
    \max_{\rho_1|_I \in \Sigma_1} \Iv{1}{\rho_1,\sigma_2}{I}
    \geq \Iv{1}{(0,1),\sigma_2}{I}
    = \frac{1}{2} \cdot 1 + \frac{1}{2} \cdot 0
    = 0.5
    .
\end{align*}
\edit{Moreover, the optimal strategy at $I$ depends the strategy player two selects at $J$.
If $\sigma_2(J)$ is s.t. $\hv{1}{\sigma_2}{h} = 0$, any strategy at $I$ will be optimal (since all strategies will yield $\Iv{1}{\cdot, \sigma_2}{I} = 0.5$).
If $\sigma_2(J)$ is s.t. $\hv{1}{\sigma_2}{h} < 0$, the only optimal strategy at $I$ will be $(0,1)$ (otherwise we would get $\Iv{1}{\cdot, \sigma_2}{I} < 0.5$).
Analogously, when $\sigma_2(J)$ is s.t. $\hv{1}{\sigma_2}{h} > 0$, the only optimal strategy at $I$ will be $(1,0)$ (which yields $\Iv{1}{\cdot, \sigma_2}{I} > 0.5$).}

We define \edit{$\sigma^A_2(J) = (\frac{1}{2},\frac{1}{2}),$ $\sigma^A_1(I)=(\frac{1}{2},\frac{1}{2})$ and $\sigma^B(J) = (0,1)$, $\sigma^A(I)=(0,1)$.}
\edit{By the above observation, both $\sigma^A$ and $\sigma^B$ are universally-optimal extensions of $\trunkS$.}
However, consider the strategy $\sigma^M := \frac{1}{2}\sigma^A + \frac{1}{2}\sigma^B$ (i.e., the $\frac{1}{2}$-convex combination of the two strategies).
\edit{We have $\sigma^M_1(I)=(\frac{1}{2},\frac{1}{2})$ and $\sigma^M_2(J)=(\frac{1}{4},\frac{3}{4})$.
By the above observation, this strategy is \textit{not} counterfactually optimal.}
\end{proof}

However, the ``slices'' of the set of optimal extensions (for a fixed strategy of one of the players) are convex:

\begin{restatable}{proposition}{SliceConvex}\label{prop:V_function_well_def}\label{prop:OE_slice_convex}
For $\trunkS$ and fixed $\sigma_{-p} \supset \trunkS_{-p}$, the set of extensions $\sigma_p \supset \sigma_p^\trunk$ for which $\sigma=(\sigma^\trunk_p,\sigma^\trunk_{-p})$ is universally (resp. counterfactually) optimal is convex.
\end{restatable}

Proposition~\ref{prop:V_function_well_def} provides some evidence that the answer to Problem~\ref{prob:VFs-convex?} might be positive.
And while the matter is by no means settled, this would correspond to the fact that we have, so far, not encountered any practical issues with neural networks generating value functions which would lead to exploitable strategies when used for depth-limited solving.

\begin{proof}
Let $\lambda \in [0,1]$ and suppose that the strategy profiles $(\rho_p,\sigma_{-p})$ and $(\nu_p,\sigma_{-p})$ are both universally-optimal extensions of $\trunkS \in \Sigma^\trunk$.
Clearly, the convex combination $\mu_p := \lambda \rho_p + (1-\lambda) \nu_p$ also extends $\trunkS_p$.
To prove the proposition, it remains to show that the strategy profile $(\mu_p, \sigma_{-p})$ is a universally-optimal extension of $\trunkS$.

Firstly, recall that the convex combination of strategies is defined in such a way that the mapping $\sigma_p \mapsto \hv{i}{\sigma_p, \sigma_{-p}}{h}$, $i\in \{1,2\}$, is convex for any fixed $\sigma_{-p}$ and $h$.
It follows that for any $\mc Z^\trunk \supset I \in \mc I_i$, the functions $\sigma_p \mapsto \Iv{i}{\sigma}{I} = \sum_{h\in I} \rp{}{\sigma}{h|I} \hv{i}{\sigma_p,\sigma_{-p}}{h}$ are likewise convex for any (since the beliefs $\rp{}{\sigma}{h|I}$ only depend on $\trunkS$).
By the assumptions on $\rho_p$ and $\nu_p$, we have $\Iv{i}{\rho_p,\sigma_{-p}}{I} = \Iv{i}{\nu_p,\sigma_{-p}}{I} = \max_{\sigma_p \in \Sigma_p} \Iv{i}{\sigma_p,\sigma_{-p}}{I}$ for all $\mc Z^\trunk \supset I \in \mc I_i$, $i=1,2$.
This concludes the proof, since the convexity of $\Iv{i}{(\cdot), \sigma_{-p}}{I}$ implies that $\mu_p$ satisfies the optimality condition for all $I\subset \mc Z^\trunk$.
(The proof in the counterfactually-optimal case is analogous.)
\end{proof}
\section{Proofs}\label{sec:proofs}

In this section, we give the proofs of the theoretical results presented in the main text.

\edit{
    \treeLemma*
}

\begin{proof}
$(c) \equiv (b)$:
Note that in $(c)$, the sum could equivalently be over $L' := \{ t \in L \mid s \sqsubset t \}$ (as $P(s,t) = 0$ for the remaining $t \in L$).
$(c)$ trivially implies $(b)$ by choosing $L := $ any slice containing $\ims(s)$.
In the opposite direction, let $L$ be a slice below $s$, $L'$ as above, and denote $L_0 := \{ s \}$, $L_{i+1} := (L \cap L_i) \cup \bigcup \{ \ims{t} \mid t \in L_i \setminus L \}$.
For $k :=$ $\max_{t\in L}$ length($t$) - length($s$), we have $L_k = L'$.
If $F$ satisfies $(b)$, it (trivially) satisfies $F(s)=\sum_{t\in L_0} P(s,t) F(t)$.
Suppose that $(c)$ holds for $L_i$.
Using the assumption from $(b)$ on any $t \in L_i \setminus L'$, we get that $(c)$ holds for $L_{i+1}$.
This shows that $(c)$ holds for $L_k = L'$, thus proving $(c)$.

$(a) \implies (b)$:
Since $P(z,z) = 1$, $F$ from $(a)$ coincides with $f$ on $Z$.
Moreover, for any $s \in T \setminus Z$, we have
\begin{align}
    F(s)
    & = \sum_{z \in Z} P(s,z) f(z)
    = \sum_{t \in \ims{s}} \sum_{t \sqsubset z \in Z} P(s, z) f(z) \\
    & = \sum_{t \in \ims{s}} \sum_{t \sqsubset z \in Z} P(s, t) P(t,z) f(z) \\
    & = \sum_{t \in \ims{s}} P(s, t) \sum_{t \sqsubset z \in Z} P(t,z) f(z)
    = \sum_{t \in \ims{s}} F(t)
    ,
\end{align}
which proves $(b)$.

$(b) \implies (a)$:
When $F$ coincides with $f$, $F$ satisfies $(a)$ (i.e., $F(t) = \sum_{z \in Z} P(t,z) f(z)$) for $t\in Z$.
When $F$ satisfies $(a)$ for all $t\in \ims(s)$, we have
\begin{align}
    F(s)
    & = \sum_{t \in \ims{s}} P(s,t) F(t)
    = \sum_{t \in \ims{s}} P(s,t) \sum_{z \in Z} P(t,z) f(z) \\
    & = \sum_{t \in \ims{s}} P(s,t) \sum_{t \sqsubset z \in Z} P(t,z) f(z) \\
    & = \sum_{t \in \ims{s}} \sum_{t \sqsubset z \in Z} P(s,t) P(t,z) f(z) \\
    & = \sum_{t \in \ims{s}} \sum_{t \sqsubset z \in Z} P(s,z) f(z)
    = \sum_{z \in Z} P(s,z) f(z)
    .
\end{align}
Thus, by backwards induction, $F$ satisfies $(a)$ for all $t\in T$.
\end{proof}

\infosetRP*

\begin{proof}
\edit{Let
$I\in \mc I_p$.
Since $I$ is assumed to be thin, it is equal to its upper frontier and we have $\rp{}{\mu}{I} = \sum_{h\in I} \rp{}{\mu}{h}$ for any $\mu \in \Sigma$.
By perfect recall, the sequence of actions taken by $p$ on the way to $h$ is the same for every $h\in I$, and likewise with the sequence of infosets encountered.
As a result, the number $\rp{p}{\mu}{h}$ is the same for every $h\in I$.
Together with the product formula \eqref{eq:rp's:product}, this gives the first part of the lemma:}
\begin{align}
    \rp{-p}{\sigma}{I}
    &
    = \max_{\rho_p \in \Sigma_p} \rp{}{\rho_p, \sigma_{-p}}{I}
    = \max_{\rho_p \in \Sigma_p} \sum_{h\in I} \rp{}{\rho_p, \sigma_{-p}}{h} \\
    &
    = \max_{\rho_p \in \Sigma_p} \sum_{h\in I} \rp{p}{\rho_p, \sigma_{-p}}{h} \rp{-p}{\rho_p, \sigma_{-p}}{h} \\
    &
    = \max_{\rho_p \in \Sigma_p} \rp{p}{\rho_p, \sigma_{-p}}{h_0} \sum_{h\in I} \rp{-p}{\sigma}{h} 
    = \sum_{h\in I} \rp{-p}{\sigma}{h} 
\end{align}
\edit{(where $h_0$ is an arbitrary element of $I$).
The last equation holds because the maximum is realized by any strategy where $p$ always selects, on the path to $I$, the action leading towards $I$ (which leads to $\rp{p}{\rho_p, \sigma_{-p}}{h}=1$).
}

\edit{For the second equality, we have}
\begin{align}
    \rp{p}{\sigma}{I}
    &
    = \rp{p}{\sigma}{h_0} \max_{\rho_{-p} \in \Sigma_{-p}} \max_{\rho_c \in \Sigma_c} \sum_{h\in I} \rp{-p}{\sigma_p, \rho_{-p}, \rho_c}{h} 
    \geq \rp{p}{\sigma}{h_0} 
\end{align}
\edit{
(witnessed by picking a strategy for chance and $-p$ that plays to reach $h_0$).
In the opposite direction, $\rp{p}{\sigma}{I}$ cannot be strictly higher than $\rp{p}{\sigma}{h_0}$ because --- no matter what chance and player $-p$ do --- there is a $1 - \rp{p}{\sigma}{h_0}$ chance that player $p$ will take some action that leads away from $h_0$, and hence (by perfect recall) away from $I$.
}

\edit{Finally, we use first two equations to get Equation \eqref{eq:rp}:
\begin{align}
\rp{}{\sigma}{I}
& = \sum_{h\in I} \rp{}{\sigma}{h}
= \sum_{h\in I} \rp{p}{\sigma}{h} \rp{-p}{\sigma}{h}
= \sum_{h\in I} \rp{p}{\sigma}{I} \rp{-p}{\sigma}{h}
\\
& = \rp{p}{\sigma}{I} \sum_{h\in I} \rp{-p}{\sigma}{h}
= \rp{p}{\sigma}{I} \rp{-p}{\sigma}{I} .
\end{align}
}
\end{proof}

\equivalentInfosetDef*

\begin{proof}
\edit{
(1):
For $x \in (0,1]$, denote $\sigma^x := (1-x) \sigma + x \textnormal{unif}$.
Our goal is to show that the limit $\rp{}{\sigma^x}{h} / \sum_{g\in I} \rp{}{\sigma^x}{I}$ exists.
For every $g\in I$, $\rp{}{\sigma^x}{g}$ is the product of $\sigma^x(k,a) = (1-x)\sigma(k,a) + x/|\mc A(k)|$, $ka \sqsubset g$.
If $n = n_g$ of the $\sigma(k,a)$-s is non-zero, the product will be equal to $C x^n (1 + x \varphi(x) )$, where $C=C_g>0$ is some constant (the product of non-zero $\sigma(k,a)$-s and $1/|\mc A(k)|$-s for those $k$ where $\sigma(k,a)=0$) and $\varphi = \varphi_g$ is some polynomial of $x$.
By denoting $m := \min_{g\in I} n_g$, we can further rewrite the product as $x^m x \psi_g(x)$ whenever $n_g > m$.
We then have
}
\begin{align}
\frac{\rp{}{\sigma^x}{h}}{\sum_{g\in I} \rp{}{\sigma^x}{h}}
= \frac{x^m}{x^m} \cdot
\frac{ C_h x^{n_h - m} (1 + x\varphi_g(x)) }{ \sum_{g, n_g = m} C_g (1 + x \varphi_g(x)) + x \sum_{g, n_g > m} \psi_g(x) }
.
\end{align}
As $x$ tends to $0$, the limit of this expression is either $0$ (when $n_h > m$) or $\frac{C_h}{\sum_{g, n_g > m} C_g}$ (when $n_h = m$).
This proves (1).

(2):
Recall that for $I\in \mc I_p$, $\rp{p}{\sigma}{g} =: \rp{p}{\sigma}{I}$ is the same for all $g\in I$.
We thus have
\begin{align}
\frac{\rp{}{\sigma^x}{h}}{\rp{}{\sigma^x}{I}}
= \frac{ \rp{p}{\sigma^x}{h} \rp{-p}{\sigma^x}{h} }{ \rp{p}{\sigma^x}{I} \rp{-p}{\sigma^x}{I} }
= \frac{\rp{-p}{\sigma^x}{h}}{\rp{-p}{\sigma^x}{I}} .
\end{align}
As $x\rightarrow 0$, we have $\rp{p}{\sigma^x}{h} \longrightarrow \rp{-p}{\sigma}{h}$ and $\rp{p}{\sigma^x}{I} \longrightarrow \rp{-p}{\sigma}{I}$.
Since $I$ being counterfactually reachable means that $\rp{-p}{\sigma^x}{I} >0$, we get  $\frac{\rp{-p}{\sigma^x}{h}}{\rp{-p}{\sigma^x}{I}} \longrightarrow \frac{\rp{-p}{\sigma}{h}}{\rp{-p}{\sigma}{I}}$ as $x\rightarrow 0$.

(3):
When $\rp{}{\sigma}{I} > 0$, we even get $\frac{\rp{}{\sigma^x}{h}}{\rp{}{\sigma^x}{I}} \longrightarrow \frac{\rp{}{\sigma}{h}}{\rp{}{\sigma}{I}}$.
\end{proof}

\pathInfosetProperties*

\begin{proof}
\edit{Whenever we manipulate limits in this proof, we should verify that this manipulation is correct and all the limits exist.
To avoid repeating the calculations done in Lemma~\ref{lem:cond-rp}, we skip this part of the proof.}

(1):
For $n\in \mathbb N$, we have $\rp{}{\sigma}{I,J} = $
\begin{align*}
    & = \sum_{g\in I} \rp{}{\sigma}{g|I} \sum_{g \sqsubset \in J} \rp{}{\sigma}{g,h}
    = \sum_{g\in I}
        \lim_{n\to \infty}
            \frac{\rp{}{\sigma^n}{g}}{\rp{}{\sigma^n}{I}} \sum_{g \sqsubset h \in J} \rp{}{\sigma}{g,h} \\
    & = \lim_{n\to \infty}
        \sum_{g\in I}
            \frac{\rp{}{\sigma^n}{g}}{\rp{}{\sigma^n}{I}} \sum_{g \sqsubset h \in J} \rp{}{\sigma}{g,h} \\
    & = \lim_{n\to \infty}
        \sum_{g\in I}
            \frac{\rp{}{\sigma^n}{g}}{\rp{}{\sigma^n}{I}}
            \left(
                \sum_{g \sqsubset h \in J}
                    \left(
                        \rp{}{\sigma^n}{g,h}  + \rp{}{\sigma}{g,h} - \rp{}{\sigma^n}{g,h}
                    \right)
        \right) \\
    & = \lim_{n\to \infty}
        \left(
        \sum_{g\in I}
            \sum_{g \sqsubset h \in J}
                \frac{\rp{}{\sigma^n}{g}}{\rp{}{\sigma^n}{I}} \rp{}{\sigma^n}{g,h}
        \right)
        + \sum_{g\in I}
            \left[
            \left(
                \lim_{n\to\infty}
                    \frac{\rp{}{\sigma^n}{g}}{\rp{}{\sigma^n}{I}}
            \right)
            \left( \sum_{g \sqsubset h \in J} 0 \right) \right]\\
    & = \lim_{n\to \infty}
                \frac{\rp{}{\sigma^n}{I}}{\rp{}{\sigma^n}{I}}
    .
\end{align*}
The proofs of the other two identities are analogous.

(2):
By (1), we have
\begin{align*}
\rp{}{\sigma}{I,J}
\overset{n\to\infty}{\longleftarrow}
    & \frac{\rp{}{\sigma^n}{J}}{\rp{}{\sigma^n}{I}}
= \frac{
        \rp{p}{\sigma^n}{J}\rp{-p}{\sigma^n}{J}
    }{
    \rp{p}{\sigma^n}{I}\rp{-p}{\sigma^n}{I}
    } =
    \\
& = \frac{ \rp{p}{\sigma^n}{J} }{ \rp{p}{\sigma^n}{I} }
    \frac{ \rp{-p}{\sigma^n}{J} }{ \rp{-p}{\sigma^n}{I} }
\overset{n\to\infty}{\longrightarrow}
\rp{p}{\sigma}{I,J} \rp{-p}{\sigma}{I,J} .
\end{align*}

(3):
By (1), we have
\begin{align*}
\rp{}{\sigma}{I,K}
\overset{n\to\infty}{\longleftarrow}
    \frac{\rp{}{\sigma^n}{K}}{\rp{}{\sigma^n}{I}}
= \frac{
        \rp{}{\sigma^n}{K}\rp{}{\sigma^n}{J}
    }{
    \rp{}{\sigma^n}{J}\rp{}{\sigma^n}{I}
    }
\overset{n\to\infty}{\longrightarrow}
\rp{}{\sigma}{I,J} \rp{}{\sigma}{J,K} .
\end{align*}
\end{proof}

\actionVals*

\begin{proof}
Let $\sigma$, $I$, and $a$ be as in the statement of the lemma.
We will expand the right-hand side of the equation to get $\Iav{}{\sigma}{I}{a}$.
For any $J \in \ims{I,a}$, denote by $J'$ the subset of $I$ for which $\{ ha \mid h \in J' \} = J$.
First, for any $g \in I$, there is either no $h \in J$ s.t. $g \sqsubset h$ (when $g \notin J'$) or exactly one such $h$ ($h=ga$, when $g \in J'$), in which case $\rp{-p}{\sigma}{g,h} = 1$.
As a result, the definition of $\rp{-p}{\sigma}{I,J}$ gives
\begin{align}
    \rp{-p}{\sigma}{I,J}
    = \sum_{g\in I} \rp{}{\sigma}{g|I}\sum_{g\sqsubset h\in J} \rp{-p}{\sigma}{g,h}
    = \sum_{g\in J'} \rp{}{\sigma}{g|I}
    = \rp{}{\sigma}{J' | I} .
\end{align}
To expand $\Iv{}{\sigma}{J}$, note first that for any $ga \in J$, we have $\rp{}{\sigma}{ga| J} = \rp{}{\sigma}{g| J'}$.
(Indeed, this is immediate when $J$ is reachable, since $\sigma_p(g,a)$ is the same for all $g\in I$.
For non-reachable $J$, we just take the limit over the amount of uniformly random noise added to $\sigma$.)
We then have
\begin{align}
\Iv{}{\sigma}{J}
= \sum_{ga\in J} \rp{}{\sigma}{ga | J} \hv{p}{\sigma}{ga}
= \sum_{g\in J'} \rp{}{\sigma}{g | J'} \hav{p}{\sigma}{g}{a}
.
\end{align}

Putting the two equations together, we get
\begin{align}
    \sum_{J \in \ims{I,a} } \rp{-p}{\sigma}{I,J} \Iv{}{\sigma}{J}
    & = \sum_{J' \subset I } \rp{}{\sigma}{J' | I} \sum_{g\in J'} \rp{}{\sigma}{g | J'} \hav{p}{\sigma}{g}{a} \\
    & = \sum_{J' \subset I } \sum_{g\in J'} \rp{}{\sigma}{J' | I} \rp{}{\sigma}{g | J'} \hav{p}{\sigma}{g}{a} \\
    & = \sum_{J' \subset I } \sum_{g\in J'} \rp{}{\sigma}{g | I} \hav{p}{\sigma}{g}{a} \\
    & = \sum_{g\in I} \rp{}{\sigma}{g | I} \hav{p}{\sigma}{g}{a}
    = \Iav{}{\sigma}{I}{a}
    .
\end{align}
\end{proof}

\infosetValueCharacterization*

\begin{proof}
$\Iv{p}{\sigma}{\textnormal{root}} = u_p(\sigma)$ follows from (1) since $\hv{p}{\sigma}{\textnormal{root}} = u_p(\sigma)$ by Lemma~\ref{lem:hist-value:characterization}.

\edit{(3) $\iff$ (3')
holds because the two formulas are equivalent.
Indeed, when $J\in \ims{I}$ and $\mc P(I) \neq p$, we have $\rp{p}{\sigma}{I,J} = 1$, so $\rp{}{\sigma}{I,J} = \rp{-p}{\sigma}{I,J}$.
When $\mc P(p) = p$, Lemma~\ref{lem:Q-vals} yields
}
\begin{align}
\sum_{J \in \ims{I} } \rp{}{\sigma}{I,J} \Iv{}{\sigma}{J}
& = \sum_{a\in \mc A(I)} \sum_{J \in \ims{I,a}} \sigma_p(I,a) \rp{-p}{\sigma}{I,J} \Iv{}{\sigma}{J}\\
& = \sum_{a \in \mc A(I)} \sigma_p(I,a) \Iav{}{\sigma}{I}{a}
.
\end{align}

\edit{
(2) $\iff$ (3) $\iff$ (4):
These equivalences follow from Lemma~\ref{lem:tree-characterization}.
Indeed, we get this by applying the lemma to the tree $(T,\sqsubset) := (\mc I_p, \sqsubset)$,
$P := \rp{}{\sigma}{\cdot,\cdot}$ (which, by Lemma~\ref{lem:path-rp-props}, satisfies $\rp{}{\sigma}{I,K} = \rp{}{\sigma}{I,J} \rp{}{\sigma}{J,K}$ whenever $I,J,K \in \mc I_p$ are s.t. $I\sqsubset J \sqsubset K$),
$f(\{z\}) := u_p(z)$,
and $F := \IvNP{}{\sigma}$.
}

\edit{
(1) $\iff$ (3):
Denote by $V'$ the $\IvNP{}{\sigma}$ from
(1)
and by $V''$ the $\IvNP{}{\sigma}$ from
(3).
We will use backwards induction on $\mc I_p$ to show that $V' = V''$.
First, when $z \in \mc Z$, the infoset $I=\{z\}$ satisfies}
\begin{align}
V'(I)
= \sum_{h\in I} \rp{}{\sigma}{h|I} \hv{p}{\sigma}{h}
= \rp{}{\sigma}{z|\{z\}} \hv{p}{\sigma}{z}
= 1 \cdot u_p(z)
= V''(I)
.
\end{align}
\edit{In particular,
the two functions are equal on the leaves of $\mc I_p$.
Second, suppose that $V'(J) = V''(J) $ for all $J \in \ims{I}$.
We then have $V''(I) = $
}
\begin{align*}
    & = \sum_{J\in \ims{I}} \rp{}{\sigma}{I,J} V''(J)
    = \sum_{J\in \ims{I}} \rp{}{\sigma}{I,J} V'(J) \nonumber \\
    & = \sum_{J\in \ims{I}}
        \rp{}{\sigma}{I,J}
        \sum_{h \in J} \rp{}{\sigma}{h|J} \hv{p}{\sigma}{h}
    = \sum_{J\in \ims{I}} \sum_{h \in J} \rp{}{\sigma}{I,J} \rp{}{\sigma}{h|J} \hv{p}{\sigma}{h} \nonumber \\
    & \overset{(X)}{=} \sum_{g \in I} \sum_{h \in \ims{g}} \rp{}{\sigma}{g|I} \rp{}{\sigma}{g,h} \hv{p}{\sigma}{h}
    = \sum_{g \in I} \rp{}{\sigma}{g|I} \sum_{h \in \ims{g}} \rp{}{\sigma}{g,h} \hv{p}{\sigma}{h} \\
    & \overset{(Y)}{=} \sum_{g \in I} \rp{}{\sigma}{g|I} \hv{p}{\sigma}{g}
    = V'(I)
    .
\end{align*}
\edit{(X) follows from the fact that
$\rp{}{\sigma}{I,J} \rp{}{\sigma}{h|I}
= \rp{}{\sigma}{g|I} \rp{}{\sigma}{g,h}
$
holds for the unique $g \in I$ for which $g \sqsubset h$.
(This is easy for fully-mixed $\sigma$ since
$\rp{}{\sigma}{I,J} \rp{}{\sigma}{h|I} 
= \frac{\rp{}{\sigma}{J}}{\rp{}{\sigma}{I}} \frac{\rp{}{\sigma}{h}}{\rp{}{\sigma}{I}}
= \frac{\rp{}{\sigma}{h}}{I}
$
and 
$ \rp{}{\sigma}{g|I} \rp{}{\sigma}{g,h}
= \frac{\rp{}{\sigma}{g}}{\rp{}{\sigma}{I}} \rp{}{\sigma}{g,h}
= \frac{\rp{}{\sigma}{h}}{\rp{}{\sigma}{I}}$.
For general $\sigma$, we take the limit over $\sigma^n = \frac{n-1}{n} \sigma + \frac{1}{n}\textnormal{unif}$.)
For (Y), we used Lemma~\ref{lem:hist-value:characterization}.}
\end{proof}

\CFVproperties*

\begin{proof}
(1): This holds because
$\IvCf{p,}{\sigma}{\textnormal{root}} = \rp{-p}{\sigma}{\textnormal{root}} \hv{p}{\sigma}{\textnormal{root}} = 1 \cdot \hv{p}{\sigma}{\textnormal{root}} $
(and Lemma~\ref{lem:hist-value:characterization} gives $\hv{p}{\sigma}{\textnormal{root}} = u_p(\sigma)$).

(2):
First, note that for any $I$ and $\sigma$, $\rp{-p}{\sigma}{I} \rp{}{\sigma}{h|I} = \rp{-p}{\sigma}{I}$.
(For counterfactually reachable $I$, this holds because $\rp{}{\sigma}{h|I} = \frac{ \rp{-p}{\sigma}{h} }{ \rp{-p}{\sigma}{I} }$ by Lemma~\ref{lem:cond-rp}.
For general $I$, the result holds because $\rp{}{\sigma}{h|I} = \lim_n \rp{}{\sigma^n}{h|I}$, where $\sigma^n$ is fully mixed and $\sigma^n \rightarrow \sigma$.)
Consequently, we have
\begin{align*}
    \rp{-p}{\sigma}{I} \Iv{}{\sigma}{I}
    = \rp{-p}{\sigma}{I} \sum_{h\in I} \rp{}{\sigma}{h|I} \hv{p}{\sigma}{h}
    = \sum_{h\in I} \rp{-p}{\sigma}{h} \hv{p}{\sigma}{h}
    .
\end{align*}

(3):
First, observe that
\begin{equation*}
\rp{-p}{\sigma}{I} \rp{}{\sigma}{I,J}
= \rp{-p}{\sigma}{I} \rp{-p}{\sigma}{I,J} \rp{p}{\sigma}{I,J}
= \rp{-p}{\sigma}{J} \rp{p}{\sigma}{I,J}
.
\end{equation*}
Combining this fact with Theorem~\ref{thm:infoset-v-characterization}, we get
\begin{align*}
    \IvCf{}{\sigma}{I}
    & = \rp{-p}{\sigma}{I} \Iv{}{\sigma}{I} 
    = \sum_{J \in \mc L } \rp{-p}{\sigma}{I} \rp{}{\sigma}{I,J} \Iv{}{\sigma}{J} \\
    & = \sum_{J \in \mc L } \rp{-p}{\sigma}{J} \rp{p}{\sigma}{I,J} \Iv{}{\sigma}{J}
    = \sum_{J \in \mc L } \rp{p}{\sigma}{I,J} \IvCf{}{\sigma}{J}
    .
\end{align*}

(4):
(a) holds because $\IvCf{}{\sigma}{Z} = \sum_{z\in Z} \rp{-p}{\sigma}{z} \hv{p}{\sigma}{z}$ (by definition of $\IvCfNP{}{\sigma}$) and $\hv{p}{\sigma}{z} = u_p(z)$ whenever $z\in \mc Z$.
To prove (b), observe first that
\begin{align*}
\IavCf{}{\sigma}{I}{a}
& = \sum_{h\in \mc A(I)} \havCf{p}{\sigma}{h}{a}
= \sum_{h\in \mc A(I)} \hvCf{p}{\sigma}{ha}
\\
& = \sum_{J \in \ims{I,a}} \sum_{ha \in J} \hvCf{p}{\sigma}{ha}
= \sum_{J \in \ims{I,a}} \IvCf{p}{\sigma}{J}
.
\end{align*}
Moreover, when $\mc P(I) = p$, we have $\rp{p}{\sigma}{I,J} = \sigma_p(I,a)$ for any $J \in \ims{I,a}$.
Combining this with (3) applied to $\mc L := $ `some slice through $\mc I_p$ containing $\ims{I}$', we get
\begin{align*}
    \IvCf{}{\sigma}{I}
    & \overset{(3)}{=} \sum_{a \in \mc A(I)} \sum_{J\in \ims{I,a}} \rp{p}{\sigma}{I,J} \IvCf{}{\sigma}{J} \\
    & = \sum_{a \in \mc A(I)} \sum_{J\in \ims{I,a}} \sigma_p(I,a) \IvCf{}{\sigma}{J}
    = \sum_{a \in \mc A(I)} \sigma_p(I,a) \IavCf{}{\sigma}{I}{a}
    .
\end{align*}
The proof of (c) is analogous to the proof of (b), except that we use the fact that when $\mc P(I) \neq  p$, we have $\rp{p}{\sigma}{I,J} = 1$ for any $J \in \ims{I,a}$.
\end{proof}

\EqPreservation*

\begin{proof}
Let $\vv$ be a value function for $\trunk$ that satisfies the assumption of the lemma.

\edit{To prove that any $\trunkS \in \textnormal{NE}(G)|_\trunk$ is a solution of $(\mc T, \vv)$,} suppose that $\trunkS \in \Sigma^\trunk$ admits some extension $\sigma^* \in \textnormal{NE}(G)$.
Since each player can extend $\trunkS$ into a non-exploitable strategy in $G$, we have $\gv{p}{G(\trunkS)} \geq \gv{p}{G}$ for both $p$, which implies that $\gv{p}{G(\trunkS)} = \gv{p}{G}$.
If $p$ were to deviate from $\trunkS$ in $(\trunk,\vv)$ and use some strategy $\rho_p^\trunk$ instead, the opponent could use $\sigma_{-p}^\trunk$ which to ensure that $\gv{-p}{G(\rho_p^\trunk, \sigma_{-p}^\trunk)} \geq \gv{-p}{G}$.
Since $G$ is zero-sum, we have $\gv{p}{G(\rho_p^\trunk, \sigma_{-p}^\trunk)} \leq \gv{p}{G}$.
Taken together, this implies that
\begin{equation*}
\DLu{p}{\trunkS} = \gv{p}{G} \geq \gv{p}{G(\rho_1^\trunk, \sigma_2^\trunk)} = \DLu{p}{\rho_p^\trunk,\sigma_{-p}^\trunk} .
\end{equation*}
In other words, no player can gain utility in $(\trunk,\vv)$ by deviating from $\trunkS$, so $\trunkS$ is a \edit{solution of} $(\trunk,\vv)$.

In the opposite direction, suppose that $\trunkS$ is a \edit{solution of} $(\trunk,\vv)$.
We will show that $\trunkS$'s trunk exploitability is $0$, implying that it can be extended into a Nash equilibrium in $G$ (Proposition~\ref{prop:trunk_expl}).
Since $\trunkS$ is a \edit{solution} of $(\trunk,\vv)$, it must satisfy
\begin{align}\label{eq:BR-and-NE-in-DL-G}
\left(\forall p \right) \left(\forall \rho_p^\trunk \in \Sigma_p^\trunk \right) : \DLu{p}{\rho_p^\trunk,\sigma_{-p}^\trunk} \leq \DLu{p}{\trunkS} .
\end{align}
Using the assumption of the lemma, we get $\gv{p}{G(\rho_p^\trunk,\sigma_{-p}^\trunk} \leq \gv{p}{G(\trunkS)}$ for each $\rho_p^\trunk$.
Substituting $\rho_p^\trunk := \sigma^*_1|_\trunk$ for some $\sigma^* \in \textnormal{NE}(G)$, we get
\begin{equation}
\gv{p}{G(\trunkS\edit{)}} \geq \gv{p}{(G(\edit{\sigma^*_1|_\trunk},\sigma_{-p}^\trunk)} \geq \gv{p}{G} .
\end{equation}
Since the game is zero-sum, it follows that
\begin{equation}\label{eq:u_in_Tv_equal_gv_G}
\gv{p}{G} = \gv{p}{G(\trunkS)} .
\end{equation}
By Proposition~\ref{prop:trunk_expl}, we have $ \expl{p}{}{\sigma_p^\trunk} = \gv{p}{G} - \gv{p}{G(\trunkS_p)} = 0$.


\end{proof}

\ComputingReachOptVf*

\begin{proof}
\edit{Recall that $\vv$ is reachably optimal if every trunk strategy $\trunkS$ admits a reachably-optimal extension for which $\vf{}{\trunkS}{h} = \hv{1}{\sigma}{h}$ (for all histories in reachable infosets in ${\mc Z}^\trunk$.
By the assumption of the proposition, each $\vf{}{\trunkS}{\cdot}$ coincides with $\hv{1}{\sigma}{\cdot}$ for some $\sigma \in NE(G(\trunkS))$.
To prove the result, we thus need to show that for any $\trunkS$, the corresponding extension $\sigma \in NE(G(\trunkS))$ is reachably optimal.
}

Let $\trunkS \in \Sigma^\trunk$ and let $\sigma$ be the corresponding element of $\textnormal{NE}(G(\trunkS))$.
Suppose that $\sigma$ \emph{isn't} reachably-optimal, i.e., suppose that there was some reachable $I_0\subset \mc Z^\trunk$ for which $p$ could increase their value of $I_0$ by switching to some strategy $\rho_p \neq \sigma_p$.
Recall that as far as $p$'s strategy is concerned, the values $\Iv{p}{(\cdot),\sigma_{-p}}{I}$ only depend on what $p$ does \textit{below $I$} --- not on what they do below other infosets in $\mc Z^\trunk$, nor on what they do in the trunk.
Without loss of generality, we can therefore assume that (i) $\Iv{p}{\rho_p,\sigma_{-p}}{I} = \Iv{p}{\sigma_p,\sigma_{-p}}{I}$ for all $p$'s infosets $I\neq I_0$ in $\mc Z^\trunk$ and (ii) $\rho_p$ is an extension of $\trunkS_p$.
By Theorem~\ref{thm:infoset-v-characterization}, we have 
\begin{align*}
u_p(\rho_p, \sigma_{-p} )
&  = \sum_{I \in \mc I_p, \ I \subset \mc Z^\trunk} \rp{}{\rho_p,\sigma_{-p}}{I} \Iv{p}{\rho_p,\sigma_{-p}}{I} \\
&  = \sum_I \rp{}{\sigma}{I} \Iv{p}{\rho_p,\sigma_{-p}}{I} \\
&  = \rp{}{\sigma}{I_0} \Iv{p}{\rho_p,\sigma_{-p}}{I_0} + \sum_{I \neq I_0} \rp{}{\sigma}{I} \Iv{p}{\sigma_p,\sigma_{-p}}{I} \\
&  > \rp{}{\sigma}{I_0} \Iv{p}{\sigma_p,\sigma_{-p}}{I_0} + \sum_{I \neq I_0} \rp{}{\sigma}{I} \Iv{p}{\sigma_p,\sigma_{-p}}{I} = u_p(\sigma_p,\sigma_{-p}) 
\end{align*}
(where the inequality required the assumption that the reach probability of $I_0$ is positive).
This shows that $p$ could increase their utility in $G(\trunkS)$ by deviating from $\sigma_p$, which contradicts our assumption of $\sigma$ being Nash equilibrium.
\edit{This shows that $\sigma$ must be a reachably-optimal extension of $\trunkS$, which }concludes the proof.
\end{proof}

\EnableDLu*

\begin{proof}
Let $\trunkS$ be a trunk strategy
\edit{and denote $\Sigma_p^\downarrow := \Sigma_p |_{\mc H \setminus \trunk}$}.
First, observe that since $G(\trunkS)$ is a zero-sum EFG, we have
\begin{equation}
\gv{p}{G(\trunkS)}
= \max_{\sigma'_1 \supset \sigma_1^\trunk} \min_{\sigma'_2 \supset \sigma_2^\trunk} u_1 (\sigma')
= \min_{\sigma'_2 \supset \sigma_2^\trunk} \max_{\sigma'_1 \supset \sigma_1^\trunk} u_1 (\sigma') 
\end{equation}
(by minimax theorem).

Second, \edit{denote $\DLIv{}{\trunkS}{I} := \sum_{h\in I} \rp{}{\trunkS}{h|I} \vf{p}{\trunkS}{h}$ for $I\subset {\mc Z}^\trunk$}
and let $\sigma = \trunkS \cup \sigma^\downarrow$ be the reachably optimal extension of $\trunkS$ from the definition of $\vv$ being reachably optimal.
\edit{Expanding the definition of $\DLu{p}{\trunkS}$ yields}
\begin{align*}
 \DLu{p}{\trunkS}
& = \sum_{h \in \mc Z^\trunk} \!\! \rp{}{\trunkS}{h} \vf{p}{\trunkS}{h}
= \sum_{I \subset \mc Z^\trunk} \!\! \rp{}{\trunkS}{I} \DLIv{}{\trunkS}{I} \\
& = \sum_{I \subset \mc Z^\trunk} \!\! \rp{}{\sigma}{I} \Iv{}{\sigma}{I}
= u_p(\sigma)
.
\end{align*}
(The first identity holds by Theorem~\ref{thm:infoset-v-characterization}.
The third holds because an infoset $I$ is either
reachable and $\Iv{p}{\trunkS, \vv}{I} = \Iv{p}{\sigma}{I}$ or
unreachable and $\rp{}{\trunkS}{I} \DLIv{}{\trunkS}{I} = \rp{}{\sigma}{I} \Iv{}{\sigma}{I}$ because $\rp{}{\trunkS}{I} = \rp{}{\sigma}{I} = 0$.)

Third, for $(\rho_1^\downarrow, \rho_2^\downarrow) \in \Sigma_1^\downarrow \times \Sigma_2^\downarrow$, denote
$f(\rho_1^\downarrow, \rho_2^\downarrow) := u_1(\trunkS_1 \cup \rho_1^\downarrow, \trunkS_2 \cup \rho_2^\downarrow)$.
To prove the proposition, it suffices to show that
\begin{equation}\label{eq:f_max_min}
    f(\sigma_1^\downarrow, \sigma_2^\downarrow) = \max_{\rho_1^\downarrow} f(\rho_1^\downarrow, \sigma_2^\downarrow) = \min_{\rho_2^\downarrow} f(\sigma_1^\downarrow, \rho_2^\downarrow)
    .
\end{equation}
Indeed, this is because \eqref{eq:f_max_min} enables the following calculation to go through:
\begin{align}
    u_1(\sigma)
    & = f(\sigma_1^\downarrow, \sigma_2^\downarrow)
    = \max_{\rho_1^\downarrow} f(\rho_1^\downarrow, \sigma_2^\downarrow)
    \\
    & \geq \min_{\rho_2^\downarrow} \max_{\rho_1^\downarrow} f(\rho_1^\downarrow, \rho_2^\downarrow)
    = \gv{p}{G(\trunkS)}
    = \max_{\rho_1^\downarrow} \min_{\rho_2^\downarrow} f(\rho_1^\downarrow, \rho_2^\downarrow)
    \\
    & \geq \min_{\rho_2^\downarrow} f(\sigma_1^\downarrow, \rho_2^\downarrow)
    = f(\sigma_1^\downarrow, \sigma_2^\downarrow)
    = u_1(\sigma) .
\end{align}

Finally, we prove the $\max$ part of \eqref{eq:f_max_min}. (The proof of the $\min$ part is analogous.)
By Theorem~\ref{thm:infoset-v-characterization} and reachable optimality, we have 
\begin{align}
u_p(\sigma)
 = \sum_{I \subset \mc Z^\trunk} \!\! \rp{}{\sigma}{I} \Iv{p}{\sigma}{I}
 = \sum_{I \subset \mc Z^\trunk} \!\! \rp{}{\sigma}{I} \max_{\rho_p \supset \trunkS_p} \Iv{p}{\rho_p, \sigma_{-p}}{I} 
\end{align}
Since the infoset values $\Iv{}{\tilde \sigma_p,\sigma_{-p}}{I}$ do not depend on the trunk portion of $p$'s strategy and can be separately optimized in each $I$, the maximum and summation can be swapped, which concludes the proof:
\begin{align}
f(\sigma^\downarrow)
 = u_1(\sigma)
 = \max_{\rho_1 \supset \trunkS_1} \sum_{I \subset \mc Z^\trunk} \!\! \rp{}{\sigma}{I} \Iv{}{\rho_1, \sigma_{2}}{I}
 = \max_{\rho^\downarrow_2} f(\sigma^\downarrow_1, \rho_2^\downarrow)
 .
\end{align}
\end{proof}

\CompCfOptVf*

\begin{proof}
Firstly, $\Iv{p}{\mu_p,\sigma_{-p}}{I} = \max_{\mu'_p} \Iv{p}{\mu'_p,\sigma_{-p}}{I} =: \Iv{p}{*,\sigma_{-p}}{I}$ holds for all reachable $I\subset \mc Z^\trunk$, since $\mu_p$ and $\sigma_p$ coincide there, and $\sigma$ couldn't be a Nash equilibrium otherwise (as we have shown in detail in the proof of Proposition~\ref{prop:comp-reach-optimal}).
Secondly, an elementary backwards-induction argument implies that all infosets $I\subset \mc Z^\trunk$ that are unreachable but counterfactually reachable by $p$ satisfy $\Iv{p}{\mu_p,\sigma_{-p}}{I} = \Iv{p}{*,\sigma_{-p}}{I}$.
Taken together, the two observations imply that $\Iv{p}{\mu_p,\sigma_{-p}}{I} = \Iv{p}{*,\sigma_{-p}}{I}$ holds for all counterfactually reachable sets.
To prove the proposition, we need to show that the same holds if $\sigma_{-p}$ is replaced by $\mu_{-p}$.

Let $I\subset \mc Z^\trunk$, $I \in \mc I_p$ be counterfactually reachable by $p$.
Applying Lemma~\ref{lem:infoset-reach} to $\Iv{p}{\rho_p,\sigma_{-p}}{I}$, we get
$\Iv{p}{\rho_p,\sigma_{-p}}{I}
= \sum_{h\in I} \frac{\rp{-p}{\trunkS}{h}}{\rp{-p}{\trunkS}{I}} \hv{p}{\rho_p,\sigma_{-p}}{h}$ for any $\rho_p \in \Sigma_p$.
The histories $h\in I$ can be divided into two parts: those for which $\rp{-p}{\trunkS}{h}$ is positive, and those for which it is zero. The above formula shows that $\Iv{-p}{\rho_p,\sigma_{-p}}{I}$ depends on the values of histories of the first type.
However, such histories will be either fully reachable (because $\rp{p}{\trunkS}{h}>0$) or counterfactually-unreachable by $p$ (because $\rp{p}{\trunkS}{h}=0$).
In either case, they do not fit the criterion ``cf. reachable but not reachable'' \textit{for the opponent} from the assumptions of this proposition, so $-p$ wasn't allowed to change their strategy below them and we have $\hv{p}{\rho_p,\mu_{-p}}{h} = \hv{p}{\rho_p,\sigma_{-p}}{h}$.
It follows that $\Iv{p}{\rho_p,\sigma_{-p}}{I} = \Iv{p}{\rho_p,\mu_{-p}}{I}$ for every strategy of $p$, which concludes the proof.
\end{proof}

\EnablingDLCFR*

\begin{proof}
The ``reachably-optimal value functions'' part holds by Example~\ref{ex:DL-CFR-fail}.

Comparing DL-CFR to CFR-D (\cite{Neil_thesis}[Alg.\,6]), we see that CFR-D is essentially a CFR run in the trunk (like DL-CFR) that additionally requires a "Solve Subgames" method. This method takes the trunk strategy as an input and returns counterfactual values of infosets in $\mc Z^\trunk$ as output. In DL-CFR, these can be obtained by calling the value function, summing the values over $I$, and weighting them by the counterfactual reach probability of $I$. This proves (1).
Theorem~17 from \cite{Neil_thesis} then guarantees that the resulting strategy is an Nash equilibrium if the counterfactual values $\IvCf{}{(\cdot)}{I} = \rp{-p}{(\cdot)}{I} \Iv{}{(\cdot)}{I}$ are maximal for all $I\subset \mc Z^\trunk$ (in the sense of corresponding to a strategy from which $p$ doesn't want to deviate).
This is equivalent to $V^{(\cdot)}_p(I)$ being maximal in counterfactually-reachable infosets.
In DL-CFR with a counterfactually-optimal value function, this holds trivially, since $\vv$ corresponds to strategies that satisfy this by definition.
\end{proof}

\NaiveBR*

\begin{proof}
Let $\trunkS \in \Sigma^\trunk$ be s.t. each $\sigma_p^\trunk$ is a naive best response to $\trunkS$ in $(\trunk,\vv)$.
Denote by $\sigma$ the counterfactually-optimal extension of $\trunkS$ that $\vv$ corresponds to.
We will show that $\sigma$ is a Nash equilibrium in $G$.

Suppose that $p$ is considering to switch over to a different strategy $\rho_p \in \Sigma_p$ in $G$.
We then have
\begin{align}
u_p(\rho_p, \sigma_{-p})
& = \sum_{I\subset \mc Z^\trunk} \rp{}{\rho_p,\sigma_{-p}}{I} \Iv{p}{\rho_p,\sigma_{-p}}{I} \nonumber \\
& \leq \sum_{I\subset \mc Z^\trunk} \rp{}{\rho_p,\sigma_{-p}}{I} \max_{\rho'_p\in \Sigma_p} \Iv{p}{\rho'_p,\sigma_{-p}}{I} . \label{eq:naive-DL-BR}
\end{align}
Since $\vv$ is counterfactually \edit{optimal}, the maximum in \eqref{eq:naive-DL-BR} is attained by $\sigma_p$ whenever $\rp{}{\rho_p,\sigma_{-p}}{I}$ is non-zero, allowing us to continue as follows:
\begin{align*}
\phantom{u_p(\rho_p, \sigma_{-p})}
& \leq \sum_{I\subset \mc Z^\trunk} \rp{}{\rho_p,\sigma_{-p}}{I} \Iv{p}{\sigma}{I} 
\edit{\ = \sum_{h \in \mc Z^\trunk} \rp{}{\rho_p, \sigma_{-p}}{h} \hv{p}{\sigma}{h} }\\
& = \sum_{h\in \mc Z^\trunk} \rp{}{\rho_p|_\trunk, \sigma_{-p}^\trunk}{h} \vf{p}{\trunkS}{h}
\leq \max_{\mu_p^\trunk \in \Sigma_p^\trunk} \sum_{h\in \mc Z^\trunk} \rp{}{\mu_p^\trunk, \sigma_{-p}^\trunk}{h} \vf{p}{\trunkS}{h}
,
\end{align*}
\edit{where the first identity also used the fact that either $\rp{}{\rho_p,\sigma_{-p}}{I} = 0$ or
\begin{align*}
\rp{}{\rho_p,\sigma_{-p}}{I} \Iv{p}{\sigma}{I}
=
\rp{p}{\rho_p}{I} \rp{-p}{\sigma_{-p}}{I} \sum_{h\in I} \frac{ \rp{-p}{\sigma_{-p}}{h} }{ \rp{-p}{\sigma_{-p}}{I} } \hv{p}{\sigma}{h}
= \sum_{h\in I} \rp{}{\rho_p, \sigma_{-p}}{h} \hv{p}{\sigma}{h}
.
\end{align*}
}
Since $\sigma_p^\trunk$ is a naive best response to $\trunkS$, the last term is equal to $\DLu{p}{\trunkS}$, which is further equal to $u_p(\sigma)$.
We have shown that $p$ cannot increase their utility by deviating from $\sigma_p$, which concludes the proof.
\end{proof}

\pureUndomPreserveNE*

\begin{proof}
\edit{By Theorem~\ref{thm:reach_opt_preserve_NE}, the conclusion holds for reachably optimal value functions (those from Definition~\ref{def:optimal-vf}, i.e., those optimal w.r.t. all strategies).
We will show that if $\vv$ and $\portfolio$ satisfy the assumptions of the theorem, $\vv$ must be reachably optimal.
To do this, it suffices to show that any $\sigma \supset \trunkS$ extension that is reachably optimal w.r.t. $\portfolio$ is reachably optimal.}

\edit{Suppose that $\sigma \supset \trunkS$ is \textit{not} reachably optimal.
By definition, this means that there is some trunk-leaf infoset $I$ with $\rp{}{\trunkS}{I} > 0$ and $p$ such that $\Iv{p}{\rho_p, \sigma_{-p}}{I} > \Iv{p}{\sigma_p, \sigma_{-p}}{I}$.
Without loss of generality, assume that $\rho_p$ is an undominated pure strategy.
(If such $\rho_p$ exists within the space of \textit{all} strategies, there will also exist some pure $\rho_p'$ with the same property, and some undominated pure $\rho_p''$ with the same property.)
Since $\rho_p|_{\mc H \setminus \trunk} \in \portfolio_p$, this shows that the extension $\sigma$ is not reachably optimal w.r.t. $\portfolio$.}
\end{proof}

\portfolioGame*

\begin{proof}
\edit{First, we prove a lemma about the equality of the depth-limited expected utility and the game value of the game $G(\trunk, \portfolio)(\trunkS)$ where the trunk strategy is fixed to $\trunkS$ (so the players only select which portfolio strategy to use below each infoset).\footnote{Recall that for a game $G$ and partial strategy $\rho$, $G(\rho)$ denotes the game where the players are ``forced'' to play $\rho$ in infosets where it is defined.}}
\begin{lemma*}\label{lem:portfolio-and-gv}
\edit{If $\vv$ is reachably optimal w.r.t. $\portfolio$,  we have $\DLu{p}{\trunkS} = \gv{p}{G(\trunk, \portfolio)(\trunkS)}$ for every $\trunkS \in \Sigma^\trunk$.}
\end{lemma*}
\edit{To prove the lemma, suppose that $\vv$ is reachably optimal w.r.t. $\portfolio$ and let $\trunkS \in \Sigma^\trunk$.
By Definition~\ref{def:optimality:generalized-strat}, there is some extension $\sigma$ of $\trunkS$ which is reachably optimal w.r.t. $\portfolio$ and satisfies $\hv{p}{\sigma}{z} = \vf{p}{\trunkS}{z}$ for all reachable $z \in \mc Z^\trunk$.
Since only the reachable $z$ make a difference when calculating utility, we have
$\DLu{p}{\trunkS}
= \sum_{z \in \mc Z^\trunk} \rp{}{\trunkS}{z} \vf{p}{\trunkS}{z}
= \sum_{z \in \mc Z^\trunk} \rp{}{\sigma}{z} \hv{p}{\sigma}{z}
= u_p(\sigma)$.
By $\sigma^{\portfolio}$, we denote the strategy from the game $G(\trunk, \portfolio)$ which corresponds to $\sigma$ (formally, $\sigma^{\portfolio}$ coincides with $\trunkS$ on $\trunk$, it is defined by condition (i) from Definition~\ref{def:optimality:generalized-strat} in reachable $I\subset \mc Z^\trunk$, and it is defined arbitrarily for unreachable $I$).
From definition of $G(\trunk, \portfolio)$, it follows that $u_p(\sigma)$ is equal the expected utility of $\sigma^{\portfolio}$ in $G(\trunk, \portfolio)$.}

\edit{To finish the proof of the lemma, it remains to show that $\sigma^{\portfolio}$ is an NE of $G(\trunk, \portfolio)(\trunkS)$ (and hence its expected utility is equal to the value of this game).
To see how the converse would lead to a contradiction, note that if $\sigma^{\portfolio}$ wasn't an equilibrium, one player $q$ would be able to increase their utility in $G(\trunk, \portfolio)$ by deviating from $\sigma^{\portfolio}$.
Since $\trunkS$ is fixed, this deviation would need to happen in some $I \subset \mc Z^\trunk$.
Moreover, to affect the total expected utility, $I$ would need to be reachable.
And without loss of generality, the new strategy in $G(\trunk, \portfolio)$ would be pure --- in $G$, this would translate to a deterministic choice of some $\rho_q^\downarrow \in \portfolio_q$.
We would then have $\Iv{}{\rho_q^\downarrow, \sigma_{-p}}{I} > \Iv{}{\sigma}{I}$ --- a contradiction with the assumption that $\sigma$ is reachably optimal w.r.t. $\portfolio$.}

\edit{We now finish the proof of the theorem.
Let $\trunkS \in \Sigma^\trunk$.
We will show that $\trunkS$ is not a solution of $(\trunk, \vv)$ if and only if it is not a restriction of some NE of $G(\trunk, \portfolio)$.
Using the lemma, we see that (A) the existence of some $\rho^\trunk_p$ for which $\DLu{p}{\rho^\trunk_p, \sigma^\trunk_{-p}} > \DLu{p}{\trunkS}$ is equivalent to (B) the existence of some (or rather, the same one) $\rho^\trunk_p$ for which $\gv{p}{G(\trunk, \portfolio)(\rho_p^\trunk, \trunkS_{-p})} > \gv{p}{G(\trunk, \portfolio)(\trunkS)}$.
By Proposition~\ref{prop:trunk_expl}, (B) is equivalent to $\trunkS$ being exploitable and thus (by Lemma~\ref{lem:trunk-expl}) not being a restriction of some Nash equilibrium of $G(\trunk, \portfolio)$.
Since (A) is literally the definition of $\trunkS$ not being the solution $(\trunk, \vv)$, this concludes the proof.}
\end{proof}

\SufficientStats*

\begin{proof}
The first case follows from Proposition~\ref{prop:comp-reach-optimal}.
Indeed, the proposition states that a reachably-optimal value function can be found by identifying a solution of $\sigma \in G(\trunkS)$ and computing the corresponding values $\hv{1}{\sigma}{h}$, $h\in \mc Z^\trunk$.
To calculate $\hv{1}{\sigma}{h}$, we only need to know how $\sigma$ looks in the bottom of the game (i.e., its trunk-portion isn't needed).
To find this bottom part of $\sigma$, recall that $G(\trunkS)$ looks like $G$, except that all decisions in $\trunk$ are done by chance, according to the probabilities that $\trunkS$ prescribes for the given state.
We can replace the whole trunk by a single chance node, where the probability of transitioning to $h\in \mc Z^\trunk$ is $\rp{}{\trunkS}{h}$.
This modified game only requires the knowledge of the reach probabilities over $\mc Z^\trunk$ while still being able to recover $\sigma|_{\mc H \setminus \trunk}$.

The second case follows from Proposition~\ref{prop:computing-cf-opt}.
Indeed, the ``in particular'' part is true because $\rp{p}{\trunkS}{I} = \rp{p}{\trunkS}{h}$ holds whenever $h \in I \in \mc I_p$.
In turn, the separated reach probabilities of individual histories allow us to recover the joint probabilities via $\rp{}{\trunkS}{h} = \rp{1}{\trunkS}{h} \rp{2}{\trunkS}{h} \rp{c}{}{h}$.
By the first part of this proposition, this enables us to find some (bottom part of) $\sigma \in \textnormal{NE}(G(\trunkS))$.
To perform the post-processing step, we need to know the beliefs $\rp{}{\sigma}{\, \cdot \,|J}$ for all counterfactually reachable infosets $I\subset \mc H \setminus \trunk$.
However, these can all be calculated by starting with the beliefs $\rp{}{\trunkS}{\, \cdot \,|I}$, $I\subset \mc Z^\trunk$, and using $\sigma|_{\mc H \setminus \trunk}$.
The beliefs $\rp{}{\trunkS}{\, \cdot \,|I}$ are only known for counterfactually reachable infosets --- however, they are also only \textit{needed} for such sets (since the post-processing is only needed in counterfactually reachable $J$).

\end{proof}

\Localization*

\begin{proof}
Suppose we are given some statistic $X : \Sigma^\trunk \times \{ S \subset \mc Z^\trunk \mid S \in \mc S \} \to \mc X$ as described by the assumptions of one of the cases of the proposition (e.g., $X(\trunkS,S) = (\rp{}{\trunkS}{h})_{h\in S}$).
To prove the result, we need to construct a value function that is optimal in the appropriate sense and show that there exists a function $\tilde \vv : \mc Z^\trunk \times \mc X \to \R$ for which ${\tilde {\mathbf v}}^{X(\trunkS,S(h))}(h) = \vf{}{\trunkS}{h}$ (where $S(h)$ denotes the public state that $h$ belongs to).

Without loss of generality, suppose that the function $X$ is surjective.
For $S \subset \mc Z^\trunk$ and $X\in \mc X$, pick an arbitrary $\trunkS \in \Sigma^\trunk$ with $X(\trunkS, S) = X$ and denote by $\sigma$ its extenstion that is optimal in the appropriate sense. (By Proposition~\ref{prop:comp-reach-optimal}, resp. \ref{prop:computing-cf-opt},
this is always possible.)
Denote by $\rho^{S,X}$ the restriction of $\sigma$ to $G(S)$.

Observe that while there was a lot of ambiguity in the choice of $\trunkS$ and $\sigma$, any extension $\mu \in \Sigma$ of $\rho^{S,X}$ that satisfies $X(\mu|_\trunk,S) = X$ will also satisfy the appropriate definition of optimality for $\mu|_\trunk$ on $S$.
Indeed, this is true because
\begin{enumerate}[(i)]
\item $\sigma$ satisfies the appropriate definition for $\sigma|_\trunk$ and $S$ and
\item all the terms present in the corresponding definition only depend on $\mu|_{G(S)}$ and $X(\mu|_\trunk,S)$, which coincide with $\sigma|_{G(S)}$ and $X(\sigma|_\trunk,S)$.
\end{enumerate}
(This is the crucial part of the proof which only goes through because $S$ is closed under infosets --- without this, we might not have all information necessary to compute $\Iv{p}{(\cdot)}{I}$.)

For $\trunkS$, define $\mu(\trunkS) := \trunkS \cup \bigcup_{S\subset \mc Z^\trunk} \rho^{S,X(\trunkS,S)}$ and $\vf{}{\trunkS}{h} := \vf{1}{\mu(\trunkS)}{h}$.
By the above observation, $\vv$ is a value function that is optimal in the appropriate sense.

Finally, denote by $\rho(S,X)$ an arbitrarily chosen extension of $\rho^{S,X}$ which satisfies $X( \rho(S,X)|_\trunk, S) = X$ and define $\tilde{\mathbf v}^X(h) := \vf{1}{\rho( S(h), X)}{h}$.
To verify the definition of $X$ being a sufficient statistic for $\vv$, let $\trunkS$ be a trunk strategy $\trunkS$ and $h \in S \subset \mc Z^\trunk$.
By their definitions, the strategies $\rho( S, X(\trunkS,S))$ and $\mu(\trunkS)$ coincide on $G(S)$.
In particular, they coincide on all descendants of $h$ and we have
\begin{equation*}
\vf{}{\trunkS}{h} = \vf{1}{\mu(\trunkS)}{h} = \vf{1}{\rho( S, X(\trunkS,S))}{h} = \tilde{\mathbf{v}}^{X(\trunkS,S)}(h) ,
\end{equation*}
which concludes the proof.
\end{proof}

\wholeRangeDep*

\begin{proof}
Let $\mc T$, $\mc Z^\trunk$, $S$, $h_0,g $, $\trunkS$ and $\mu^{\mc T}$ be as in Theorem~\ref{thm:wholeRangeDep}.
We shall prove the theorem by constructing $G$ and showing that it has the desired properties.

We start by making two simplifying assumptions.
First, we assume that each $h\in S$ only has one legal (dummy) action that we denote $d$. In the general case, each $ha$, $hb$ would be extended identically, complicating the notation but not introducing any real challenges.
Since the public state cannot be further refined, there exists a sequence of histories satisfying $h_0 \sim h_1 \sim \dots \sim h_n = g$ in $S$, where $\sim$ denotes ``one of the players cannot distinguish between the two histories''.
We assume, without loss of generality, that $h_0,\dots,h_n$ is the shortest among such sequences. In particular, it follows that the sequence looks like the one in Figure~\ref{fig:wholeRangeDep} (histories and information sets don't repeat, players unable to distinguish them alternate).
We only show the proof in the case where both $h_0, h_1$ and $h_{n-1}, h_n$ are indistinguishable by the first player (the proofs of the remaining three cases are similar).
(Note that the proof only requires the reach probabilities to be non-zero for the histories $h_i$, rather than on the whole $S$.)

The history tree $\mc H$ of $G$ is s.t. $\trunk$ is a trunk in $G$ and $\mc Z^\trunk$ its leaves.
Outside of $S$, $G$ can continue arbitrarily --- e.g., by each node $h \in \mc Z^\trunk \setminus S$ being terminal with $u_1(h) = 0$.
Finally, at $\{ hd \mid h\in S\}$ and below, $G$ is defined as follows :
\begin{itemize}
\item For $h \in S \setminus \{h_i \mid i=0,\dots,n \}$, $hd$ is a terminal node with utility 0 (for all strategic considerations, this replaces $S$ by $\{h_i \mid i=0,\dots,n \}$).
\item For $h_i$, $i=0,\dots,n-1$, $h_id$ leads to a matching pennies game (a matrix game with actions $U$, $D$ for player 1, actions $L$, $R$ for player 2, and corresponding utilities $1$ for $U,L$ and $D,R$, resp. $0$ for $U,R$ and $D,L$).
\item For $h_n$, $h_nd$ leads to a game where only player 1 acts, choosing between $U$ (utility 0) and $D$ (utility 1).
\item The information sets below $h_i$ are defined in such a way that player 1 has to use the same strategy below $h_0$ and $h_1$, $h_2$ and $h_3$, \dots, $h_{n-1}$ and $h_n$, and player 2 has to use the same strategy below $h_1$ and $h_2$, $h_3$ and $h_4$, \dots, $h_{n-2}$ and $h_{n-1}$ (player 2 strategy below $h_0$ is independent of everything else).
\end{itemize}

Since $h_n$ is unreachable under $\mu^\trunk$, $\widetilde G(\mu^\trunk)$ below $S$ is effectively a collection of (interconnected) matching pennies games.
It follows that the uniform strategy of both players is a Nash equilibrium (clearly, no player can improve his overall utility).
On the other hand, it is \emph{not} a NE strategy in $\widetilde G(\trunkS)$ below $S$ (since player 1 could improve his utility be deviating to ``$D$ everywhere'').

In particular, $0.5$ is an expected utility of $h_0$ under some NE strategy in $G(\mu^\trunk)$.
Suppose that some NE strategy $\sigma$ in $G(\trunkS)$ has $\hv{1}{\sigma}{h_0} = 0.5$.
We will show that such $\sigma$ has to be uniformly random, and thus prove the theorem by contradiction.

Firstly, if $\sigma_1$ wasn't uniformly random at the information set $\{h_0d,h_1d\}$, player 2 could increase his overall utility by changing his strategy below $h_0$ to either $L$ or $R$ and thus $\sigma$ wouldn't be a NE.

We proceed inductively.
We know that $u^\sigma_1(h_0) = 0.5$, and that in $\{h_0d,h_1d\}$, $\sigma_1$ takes both $U$ and $D$ with non-zero probability. 
If $\sigma_1$ is to be a NE, player 1 has to be indifferent between playing $U$ and $D$ in $\{h_0d,h_1d\}$.
Since $\hv{1}{\sigma}{h_0} = 0.5$, this can only be achieved if $\sigma_2$ takes both $L$ and $R$ below $h_1$ with the same probability.
In particular, $u^\sigma_1(h_1)=0.5$.
Since $G$ forces the strategy of player 2 to be the same below $h_1$ and $h_2$, we get that $\sigma_2$ is uniformly random below $h_2$ as well.

We repeat the argument above for each $h_i$, eventually showing that if the players are to be indifferent between the actions they take with non-zero probability, $\sigma_1$ has to be uniformly random in the whole $G(S)$ and $\sigma_2$ has to be uniformly random below $h_1,\dots,h_{n-1}$.
Finally, if $\sigma_1$ wasn't uniform below $h_0$ (but was below $h_1$), player 1 could increase his utility by deviating to either $U$ or $D$.
This implies that the whole $\sigma$ is uniformly random, which contradicts our earlier observation.
\end{proof}

\section{Detailed Descriptions of the Domains}\label{sec:app:domains}

In this section, we give a more specific description of the domains used for evaluation.


\subsection{The Rules of Leduc hold'em (\textbf{LH})}
\edit{Leduc Hold’em, which was first introduced in \cite{southey2012bayes}, is played with six cards and two suits: Two Jacks, two Queens, and two Kings; on of each per suit.
Each player gets dealt a card and submits an ante of $1$.
Each betting round has a maximum of two raise actions.
    (In other words, the possible actions are either raise or check if acting first in a betting round, re-raise, call or fold if faced with a raise, and call or fold if faced with a re-raise.)
There are two betting rounds.
After the first betting round, a public card will be shown and a second second betting round is performed using the same rules.
The raise size is 2 in the first betting round and 4 in the second.
A player wins if their private card matches the public card and the opponent's card does not match, or if none of the players' cards matches the public card and their private card is higher than the private card of the opponent.
If no player wins, the game is a draw, and the pot is split.
The maximum utility in this game is 13 (but we normalize it to 1 in the experiments).}

\subsection{The Rules of Imperfect-Information Goofspiel (\textbf{GS})}

\edit{In goofspiel with $N$ cards, each player is given a private hand of \textit{bid cards} with values $1$ to $N$.
A different deck of $N$ \textit{point cards} is placed face up in a stack.
Each turn, one of the point cards is revealed and each player bids for it by secretly choosing a single card in their hand.
The highest bidder gets the point card and adds the point total to their score (in case of a tie, the point card is discarded).
This is repeated $N$-times until both players run out of cards, and the player with the highest score wins.
In imperfect-information goofspiel, the players only discover who won or lost a bid, but not which bid cards were chosen.
We assume that the point cards are revealed in a decreasing order and that this is common knowledge.
We use $N=5$ and define the each player's utilities as the sum of the cards they won minus the sum of the opponent's cards.
As a result, the maximum utility is $(5+4+3+2) - 1 = 13$ (obtained when the opponent only wins the last card).Note that the structure of the game is different from Leduc hold'em:First, the actions are not perfectly observable. Second, if a player plays a card and loses, the opponent might have played any higher card --- this implies that the size of the information sets and public states first grows at the start of the game and then shrinks towards the end}

\subsection{The Rules of Oshi-zumo (\textbf{OZ})}

\edit{Oshi-zumo \cite{oshizumo} is played by two players,  both of whom start with $N$ coins.
At the beginning of the game, a sumo wrestler is positioned at the center of a one-dimensional playing field which consists of $2K + 1$ locations.
Each round each player secretly spends between some number of coins --- at least $M$ and at most the number of coins they have remaining.
(If a player ends up with no coins, while their opponent still has some, the opponent immediately wins.)
Whoever spent more coins then gets to push the wrestler one location towards the opponent's side.
If this causes the wrestler to fall off the playing field, the player who did the pushing wins.
(If the bids are equal, the wrestler does not move.)
We use an imperfect-information variant of the game, where the winner is publicly revealed, but it is not revealed how many coins each of the players spent.
If both players run out of coins at the same time, the player who pushed the wrestler further (i.e., the player who does not have the wrestler on their side) wins.
(If the wrestler is located at the center, the game result is a draw.)
In our variant of the game, winning yields $1$ utility.
We use $N=8$ and $K=1$.}

\edit{Like goofspiel, oshi-zumo is comprised of variably-sized information sets resulting from the uncertainty about the opponent's bid.
Unlike in GS, not all terminal branches in OZ have the same length (because of the possibility to run out of coins prematurely).}
\section{Choice of the Loss Function}\label{app:sec:loss}
\edit{A crucial part of approximating value functions is to understand which loss to minimize to achieve the best performance.}

\paragraph{Minimizing Huber, $l_1$ or $l_\infty$}

\edit{To explore and determine which of the standard losses are best, we used both the same architecture and data to train value networks, validating on all three candidate loss functions, however each of them minimizing either Huber, $l_1$ or $l_\infty$ loss. Across the three cases, we then compared the minimum validation losses which could be achieved under the particular setting. This analysis revealed that there is no significant difference between all three losses in terms of their achieved validation losses.}

\edit{We also performed an extensive analysis of alternative loss functions Below, we describe the high-level ideas.}

\paragraph{Range-weighting}

\edit{Since the main metric, we care about in the value network, is the achieved exploitability when used in conjunction with \DLCFR, an intuitive idea would be to prioritize the error on infosets which are actually played to under a certain trunk strategy, i.e., the range of the infoset. If an arbitrary infoset's range is zero, the value network's error on it should be less important than one which has high range.Hence, we implemented a loss function which computes the standard Huber, $l_1$ or $l_\infty$ error on every sample, however multiplies the error by the particular range of the infoset.}

\paragraph{Public state value weighting}
\edit{A similar idea was to emphasize errors not just on particular infosets but on full public states, i.e., a full sample. As some public states have higher value than others, doing badly in them might hurt the resulting exploitability more than in low value public states. We did so by precomputing the values of each public state and then multiplying the resulting errors with these values.}

\paragraph{Zero-sum Loss}
\edit{Another idea, already realized by \cite{DeepStack}, was the zero-sum loss which is computed by multiplying each player's ranges with their corresponding counterfactual values and summing the products up to a scalar for each player. The sum of both of these numbers should then be equal to $0$. This loss can then be minimized in conjunction with standard losses.}

\paragraph{Enforcing the right ratio between values}
\edit{A general phenomenon we observed when analysing the game-solving behavior of \DLCFR was also the algorithm's vulnerability in regard to values which should be negative or exact zero but are, due to naturally occurring neural network inaccuracies, very small positive numbers. Relatedly, in regret matching absolute values of regrets do not really matter, but rather the ratio between all values. In the case where an infoset's correct regret matching strategy should be pure for one action this problem can change a strategy from pure to mixed and cause the algorithm to visit states which result in high exploitability. This leads us to conclude that predicting the right ratio between counterfactual values is key and should be enforced through the loss signal. Since negative values in regret matching are pruned to $0$, we implemented a loss function which first shifts all values into the positive range using a constant and then applies softmax to both the targets and predictions and minimizes the Kullback-Leibler divergence between the two.}

\paragraph{Summary of explored loss functions}
\edit{All of the tried variations did not improve (1) the speed of convergence (2) proved to be linear transformations of standard loss functions and, most importantly, did not have a stronger correlation with the resulting exploitability. Despite that, we decided to report the findings of this extensive search and concluded that using one of the standard losses is sufficient. Finally, we chose Huber as the loss to run all experiments on. The code for all above described loss functions is accessible under \cite{vfrepo}.}

\section{Detailed Investigation of Value Networks}\label{sec:app:graphs}

\subsection{Hyperparameter Optimization}\label{sec:app:hyperparams}

\edit{We start out by describing the hyperparameter optimization performed in order to train our neural network.
In particular, we try to find the right choice of network width and depth, loss function, and the amount of data.
While our main results are presented using normalized losses, we opted to show non-normalized losses in this section. The reason for that is that the objective of this investigation is strictly about finding optimal settings for certain hyperparameters as opposed to judging certain performance criteria of the full algorithm across domains.}

\subsubsection{Layer Width}
The three following figures show the search through different widths of hidden layers for each domain.
The y-axis denotes the best non-normalized error achieved with a given setting after 1000 epochs. The x-axis denotes the ratio of the number of neurons to the input size.

\begin{figure}[H]
\centering
\includegraphics[trim=0 0 0 47,clip,width=\linewidth]{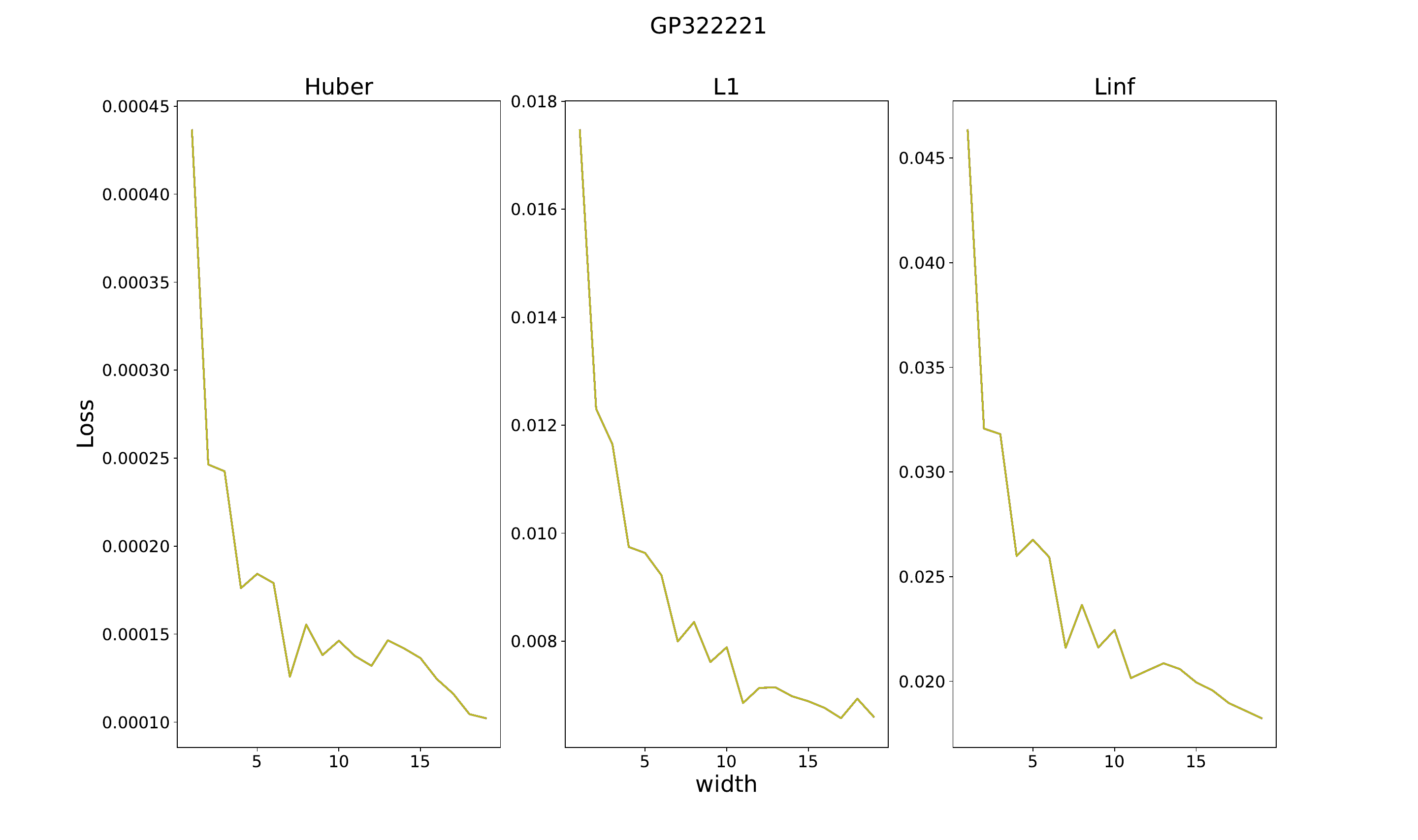}\hfill
\caption{Layer width grid search in Leduc hold'em. The y-axis denotes validation error while the x-axis shows number of neurons as a multiplier of the input size.}
\end{figure}

\begin{figure}[H]
\centering
\includegraphics[trim=0 0 0 47,clip,width=\linewidth]{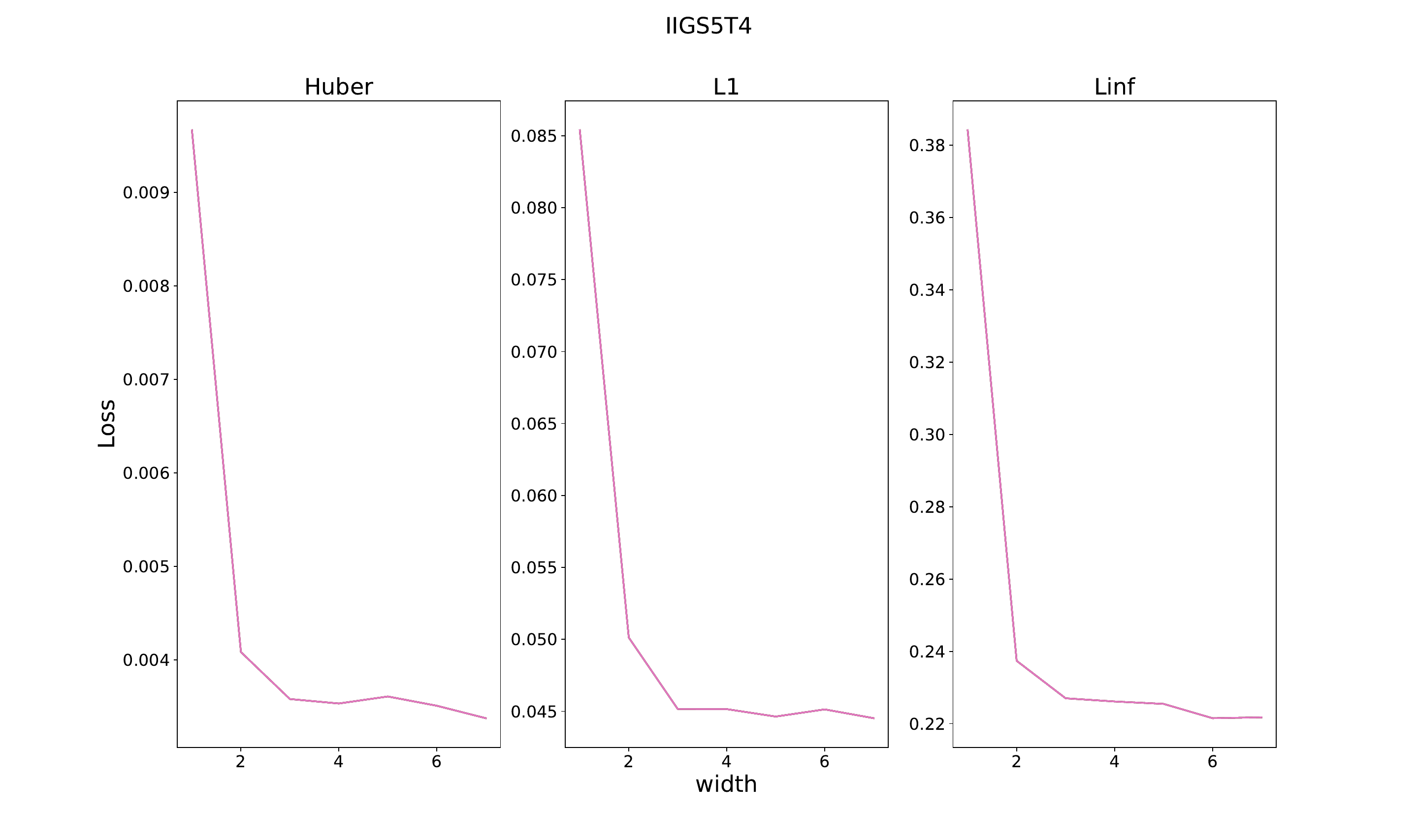}\hfill
\caption{Layer width grid search in goofspiel. The y-axis denotes validation error while the x-axis shows number of neurons as a multiplier of the input size.}
\end{figure}

\begin{figure}[H]
\centering
\includegraphics[trim=0 0 0 47,clip,width=\linewidth]{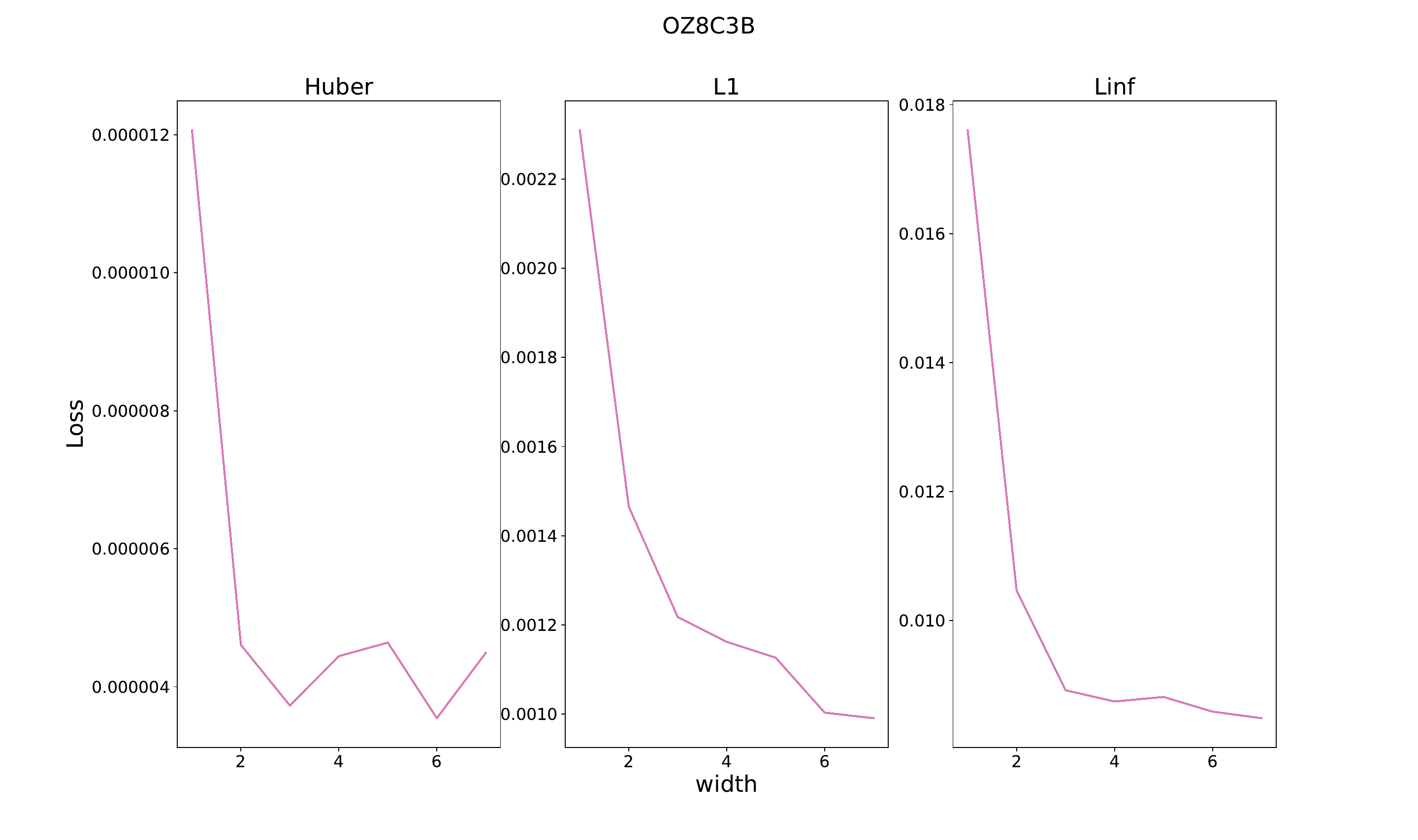}\hfill
\caption{Layer width grid search in oshi-zumo. The y-axis denotes validation error while the x-axis shows number of neurons as a multiplier of the input size.}
\end{figure}

\subsubsection{Network Depth}

We also explored different neural network depths for each domain. We display the best non-normalized loss achieved (the y-axis) with the given number of layers (the x-axis) after 1000 epochs.

\begin{figure}[H]
\centering
\includegraphics[trim=0 0 0 47,clip,width=\linewidth]{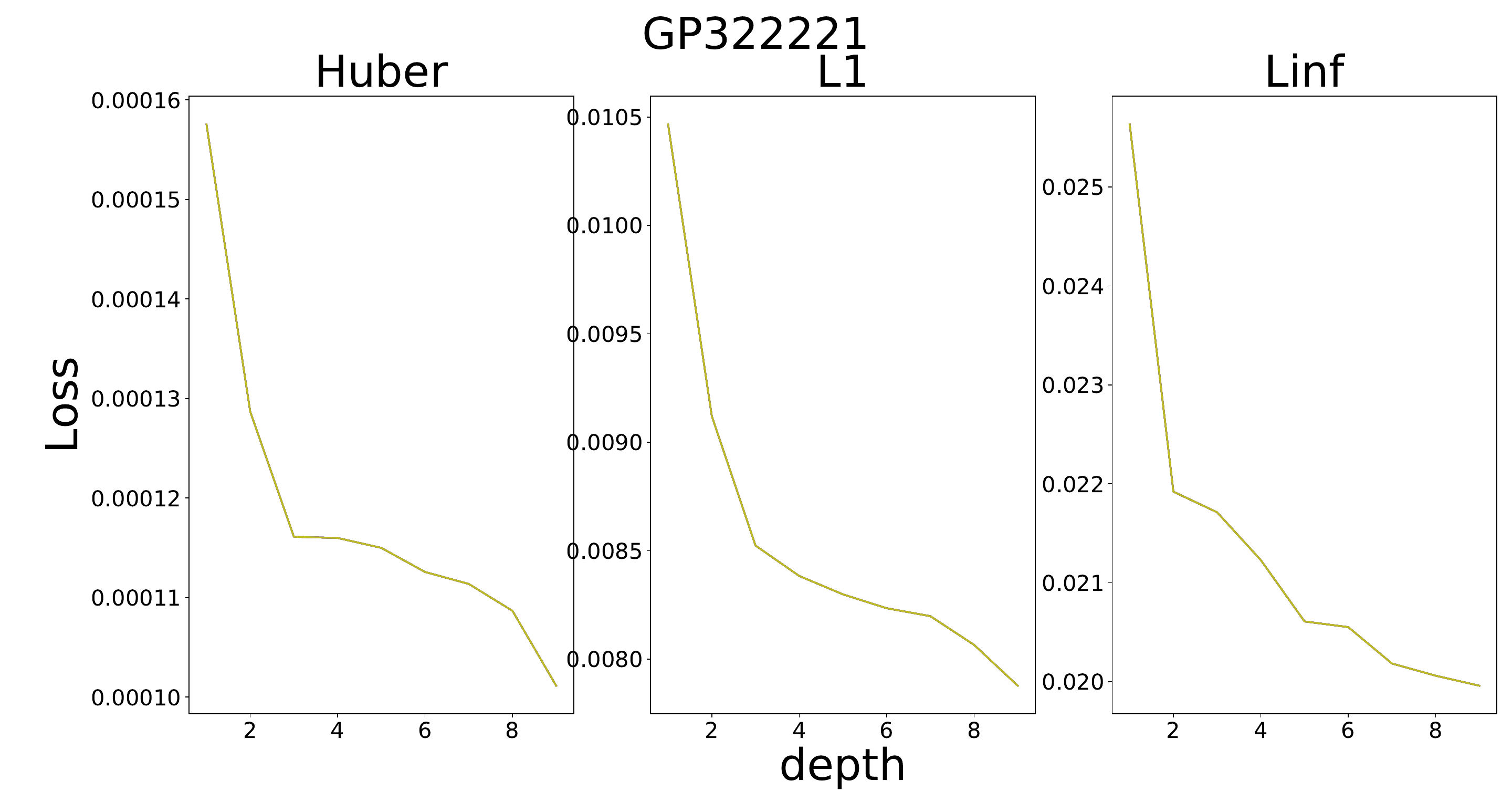}\hfill
\caption{Layer depth grid search in Leduc hold'em. The y-axis denotes validation error while the x-axis shows number of hidden layers.}
\end{figure}

\begin{figure}[H]
\centering
\includegraphics[trim=0 0 0 47,clip,width=\linewidth]{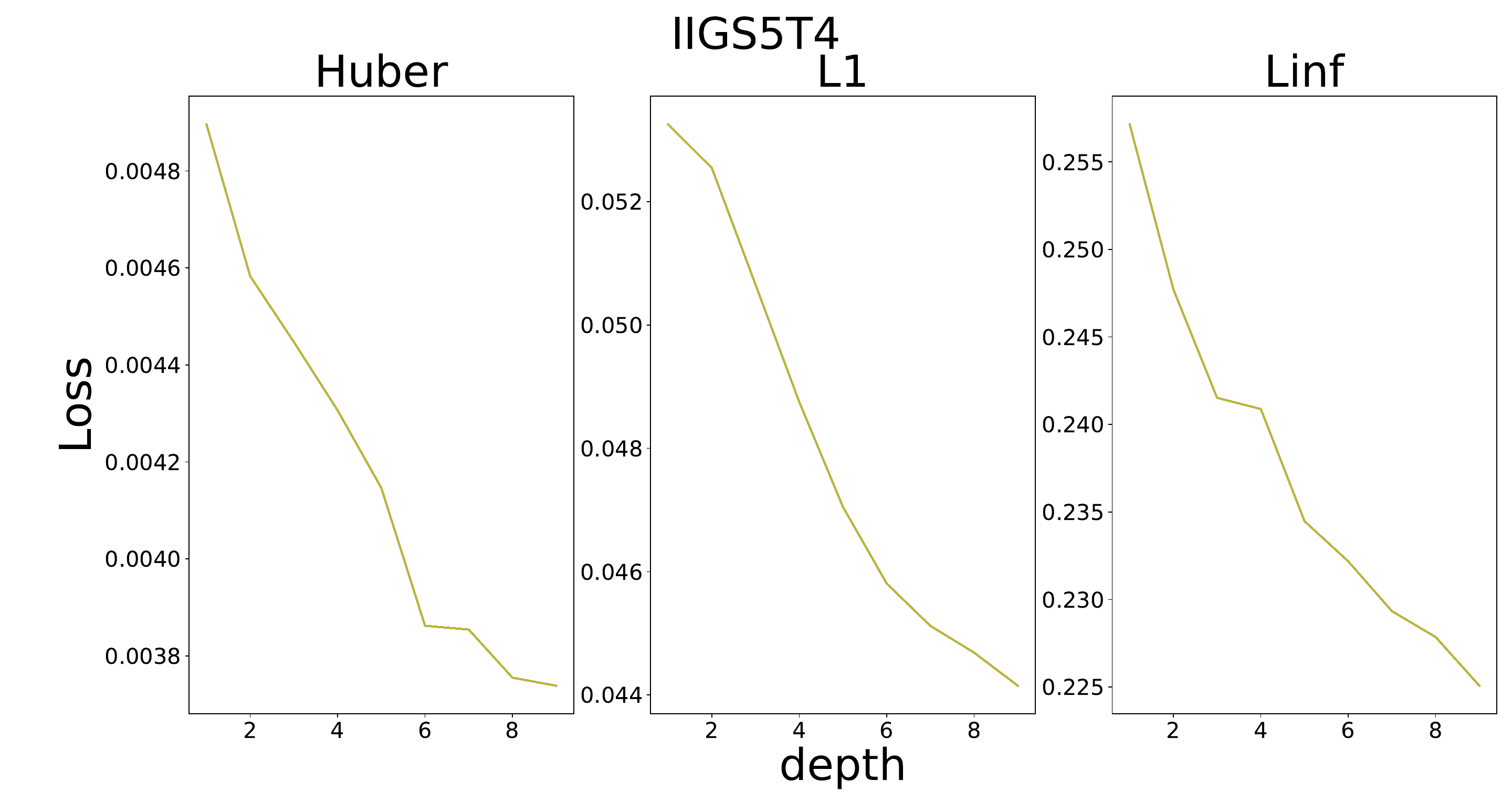}\hfill
\caption{Layer depth grid search in goofspiel. The y-axis denotes validation error while the x-axis shows number of hidden layers.}
\end{figure}

\begin{figure}[H]
\centering
\includegraphics[trim=0 0 0 47,clip,
width=\linewidth]{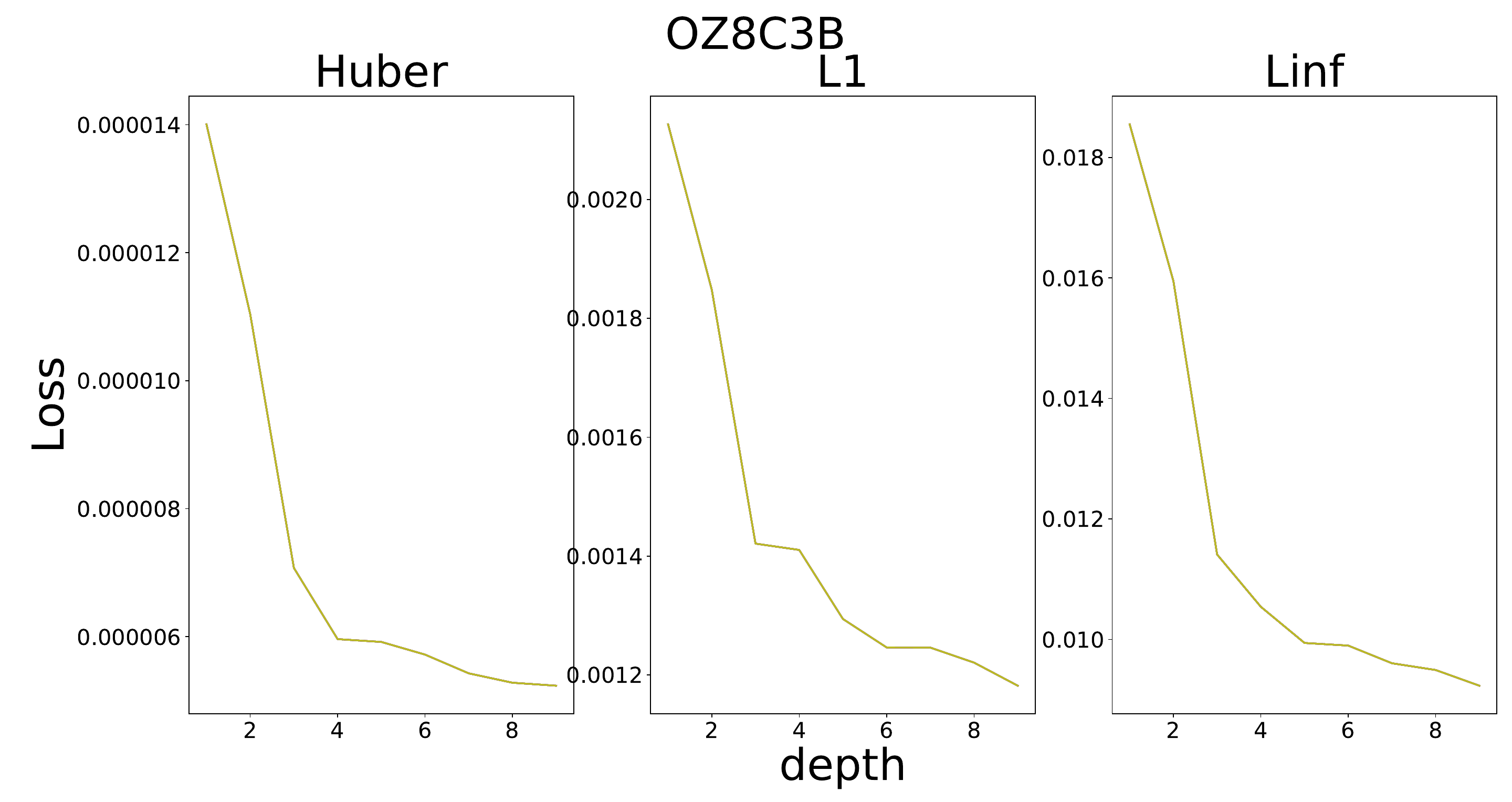}\hfill
\caption{Layer depth grid search in oshi-zumo. The y-axis denotes validation error while the x-axis shows number of hidden layers.}
\end{figure}

Recall, that Table~\ref{table:finalarch2} in the main text shows the final architectures used, which were based on the above described grid searches.

\subsubsection{Data Amount}

\begin{figure}[H]
\centering
\includegraphics[trim=0 0 0 47,clip,width=\linewidth]{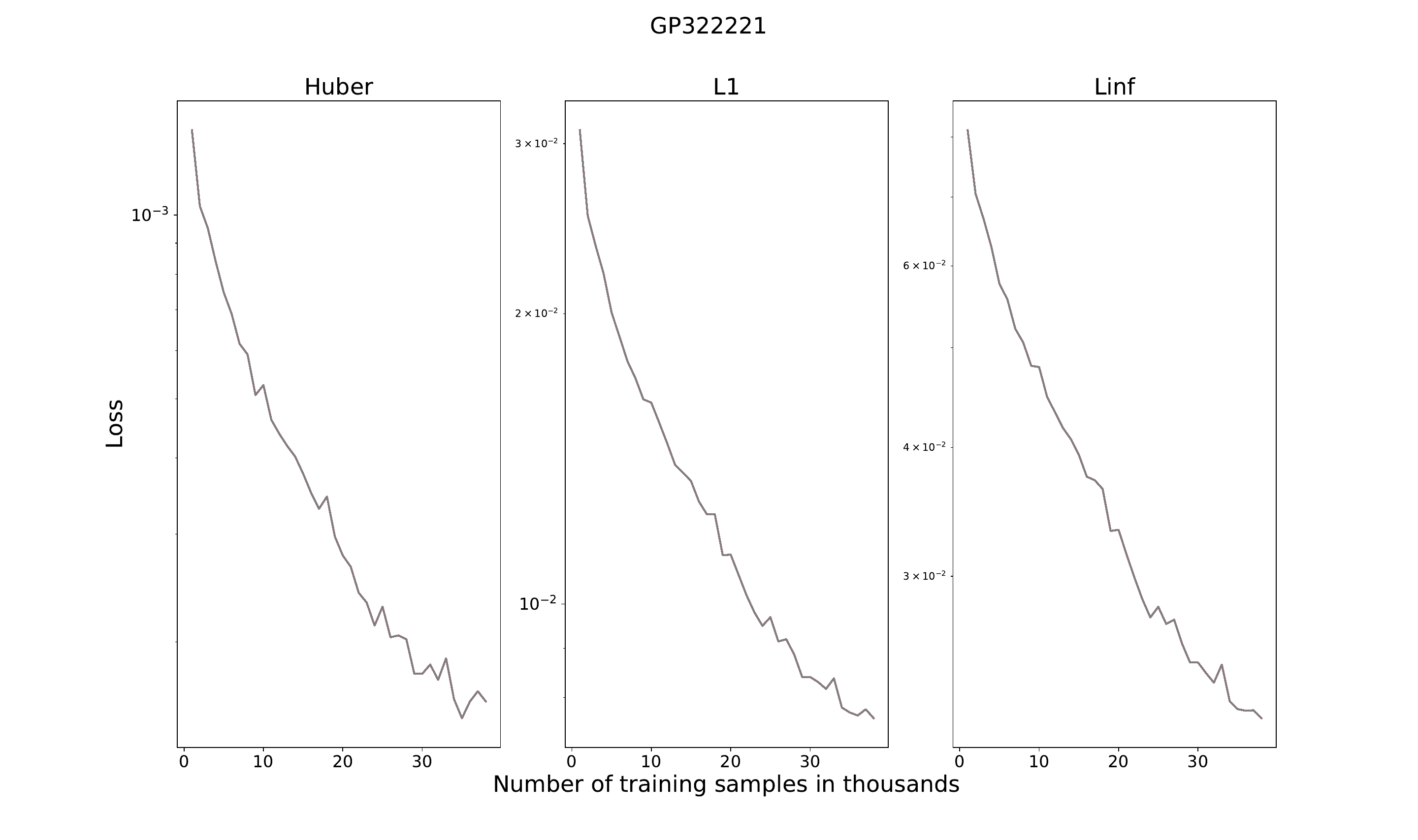}
\caption{Relationship of training samples on the validation errors in LH. We show loss on the y-axis and number of training samples on the x-axis.}
\label{app:fig:app_data_lh}
\end{figure}

\begin{figure}[H]
\centering
\includegraphics[trim=0 0 0 47,clip,width=\linewidth]{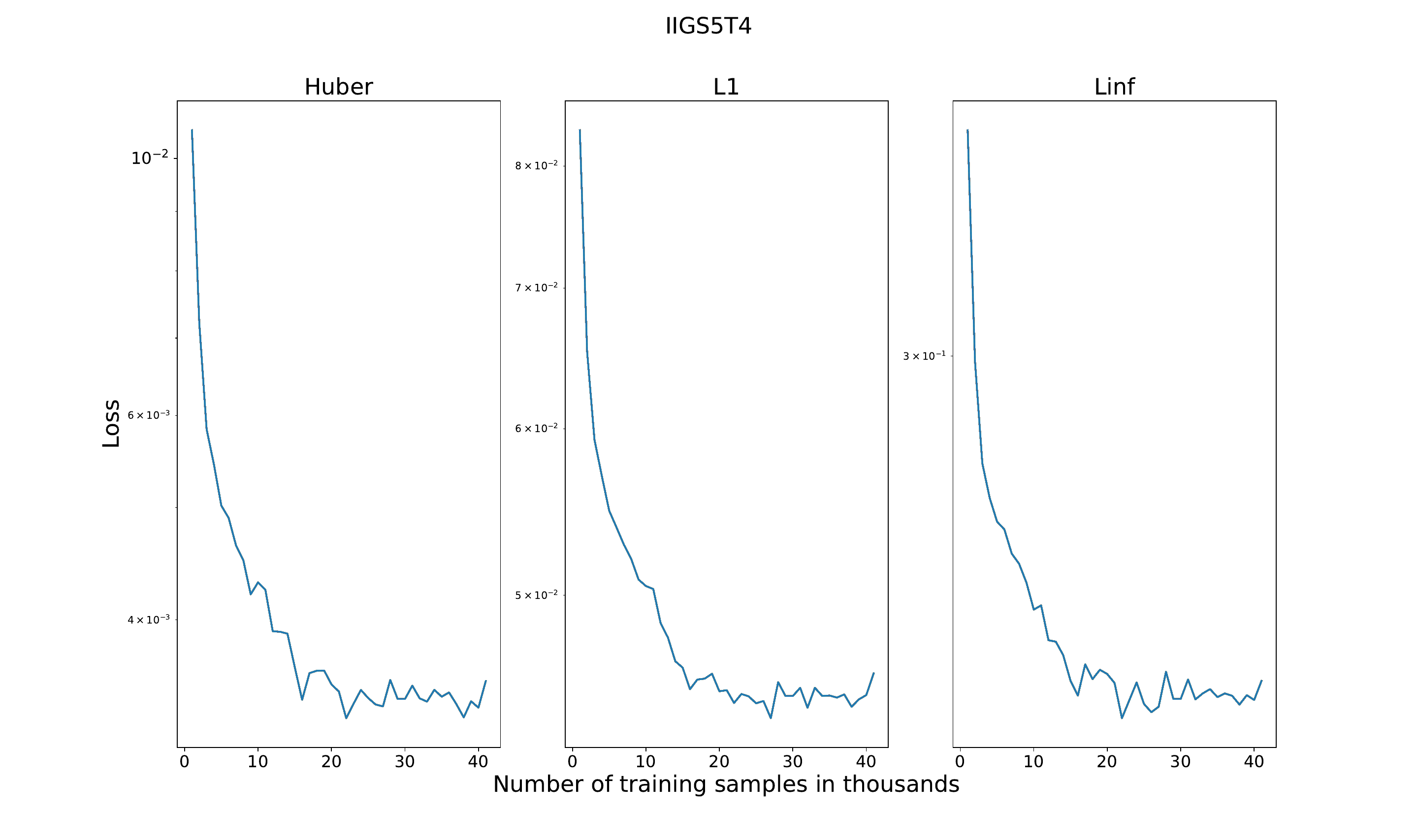}
\caption{Relationship of training samples on the validation errors in GS. We show loss on the y-axis and number of training samples on the x-axis.}
\label{app:fig:app_data_gs}
\end{figure}

\begin{figure}[H]
\centering
\includegraphics[trim=0 0 0 47,clip,width=\linewidth]{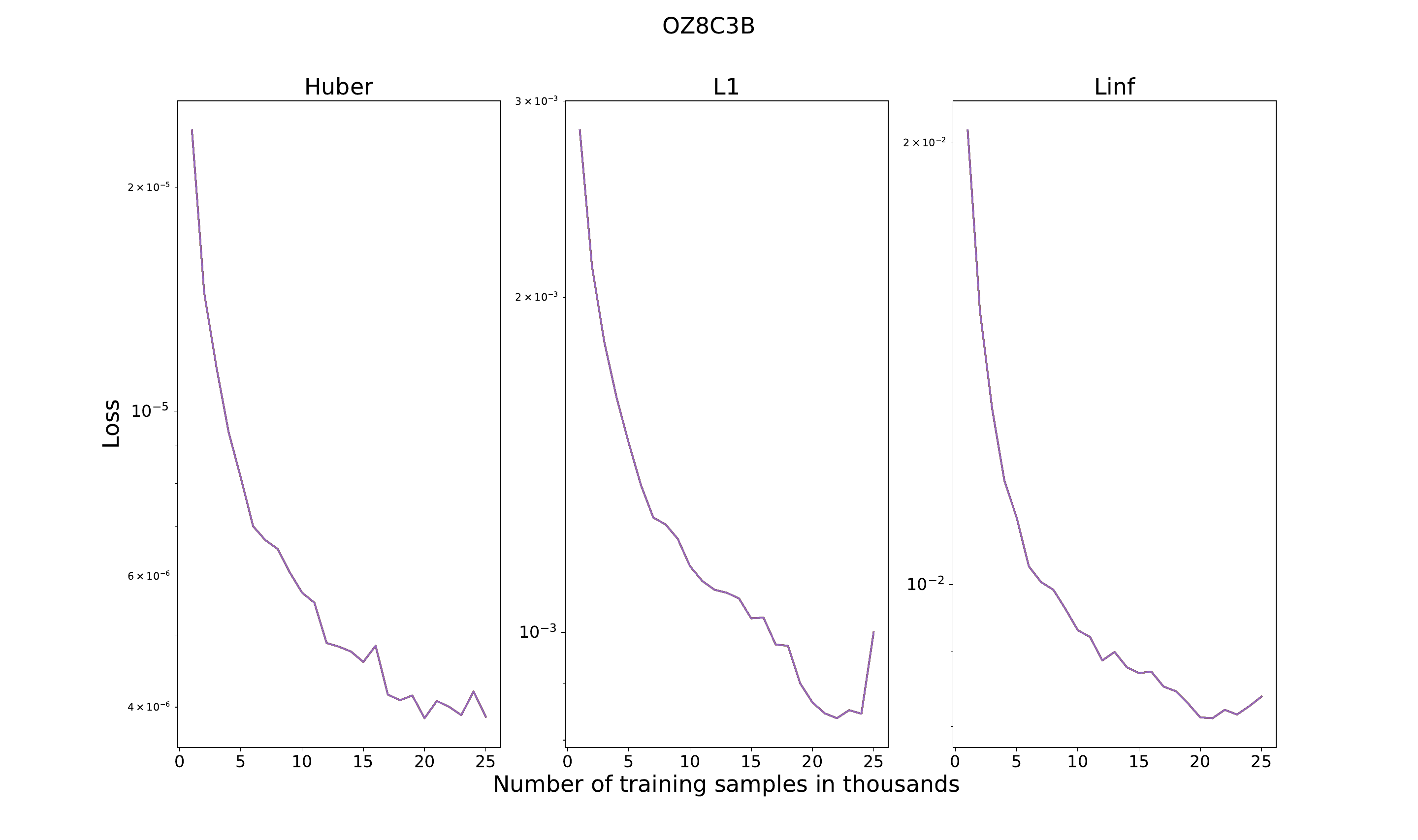}
\caption{Relationship of training samples on the validation errors in OZ. We show loss on the y-axis and number of training samples on the x-axis.}
\label{app:fig:app_data_oz}
\end{figure}

We are also interested in how much data is required to achieve a sufficiently low loss.
Recall that the network always takes some public state $S$ and a range $r$ at $S$ as input (and returns a vector $\vec v$ of values, one for each infoset at $S$).
We thus measure the amount of data in the number of ($S$, $r$, $\vec v$) tuples used for training.
Note that each of the randomly generated training trunk-strategies (Section~\ref{sec:sub:training-data}) corresponds to a as many tuples as there are public states at the depth limit (Table~\ref{tab:domains_description2}).
Hence, the more public states there are in a domain the less random trunk strategies needed.
We show the validation loss after training 1000 epochs for a given data amount setting (LH in Fig.~\ref{app:fig:app_data_lh}, GS in Fig.~\ref{app:fig:app_data_gs}, and OZ in Fig.~\ref{app:fig:app_data_oz}).

\edit{We conclude that even with below 10k datapoints we can achieve low validation errors depending in every domain. However, it is clear that with more data it is possible to reach orders of magnitude better validation losses. This is however not necessary to guarantee low exploitability. Regardless of that, to run our main experiments, we opted to use a larger amount of data to properly be able to explore the algorithm's performance: the numbers we used for our main experiments were $18\,000$ for goofspiel, $70644$ for Leduc and $34\,000$ for oshi-zumo. }

\subsection{Performance on CFR-D Data}\label{sec:app:val_cfrd}

We also wanted to see the performance of the network on the ranges which CFR-D requests when solving the game.
To achieve that, we trained the network on the standard random ranges and validated on the CFR-D ranges.
The following figures summarize this experiment for all domains.

\begin{figure}[H]
\centering
\includegraphics[trim=0 0 0 95,clip,width=\linewidth]{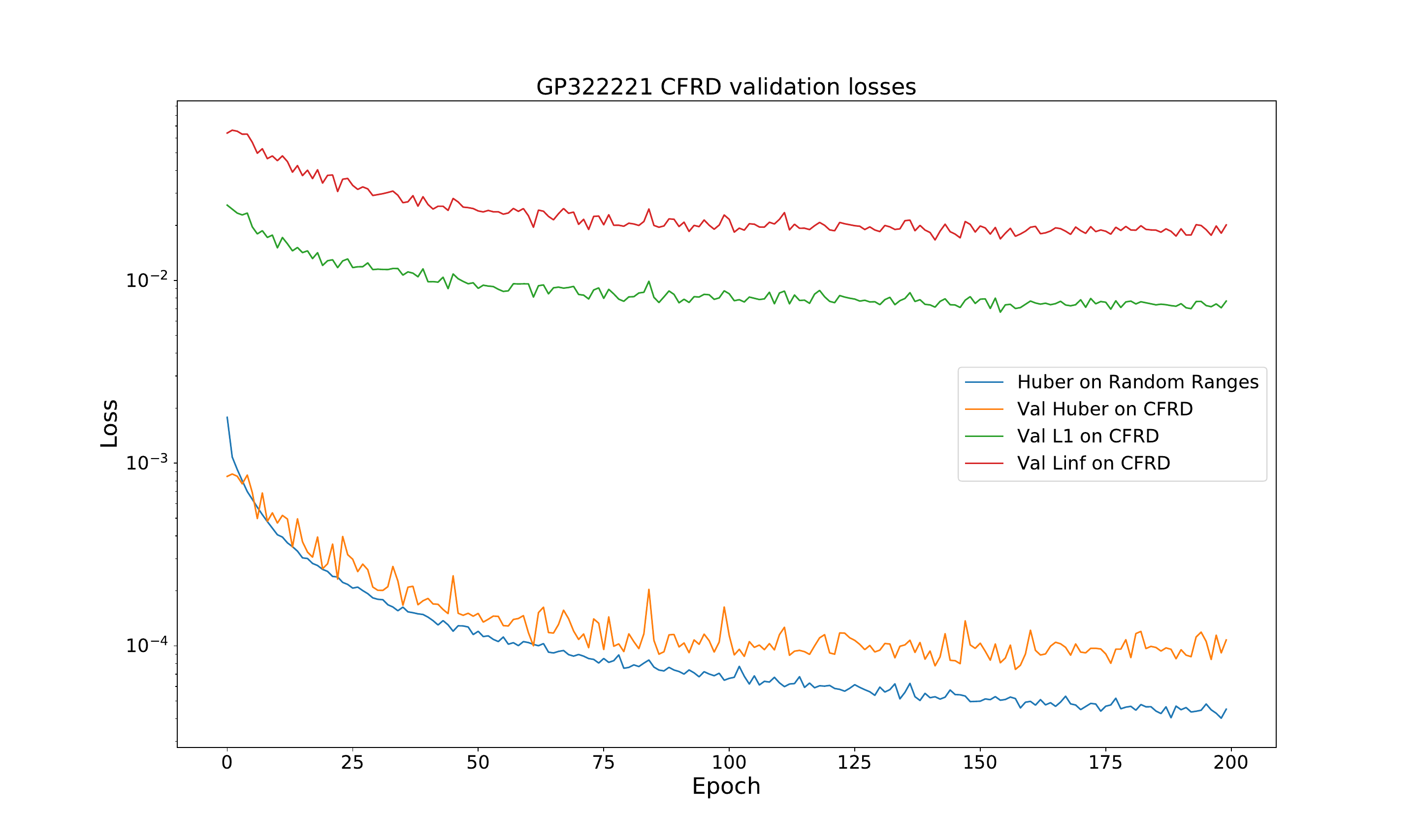}\hfill
\caption{LH: Training a network on random ranges and measuring validation error on CFR-D data.}
\end{figure}

\begin{figure}[H]
\centering
\includegraphics[trim=0 0 0 95,clip,width=\linewidth]{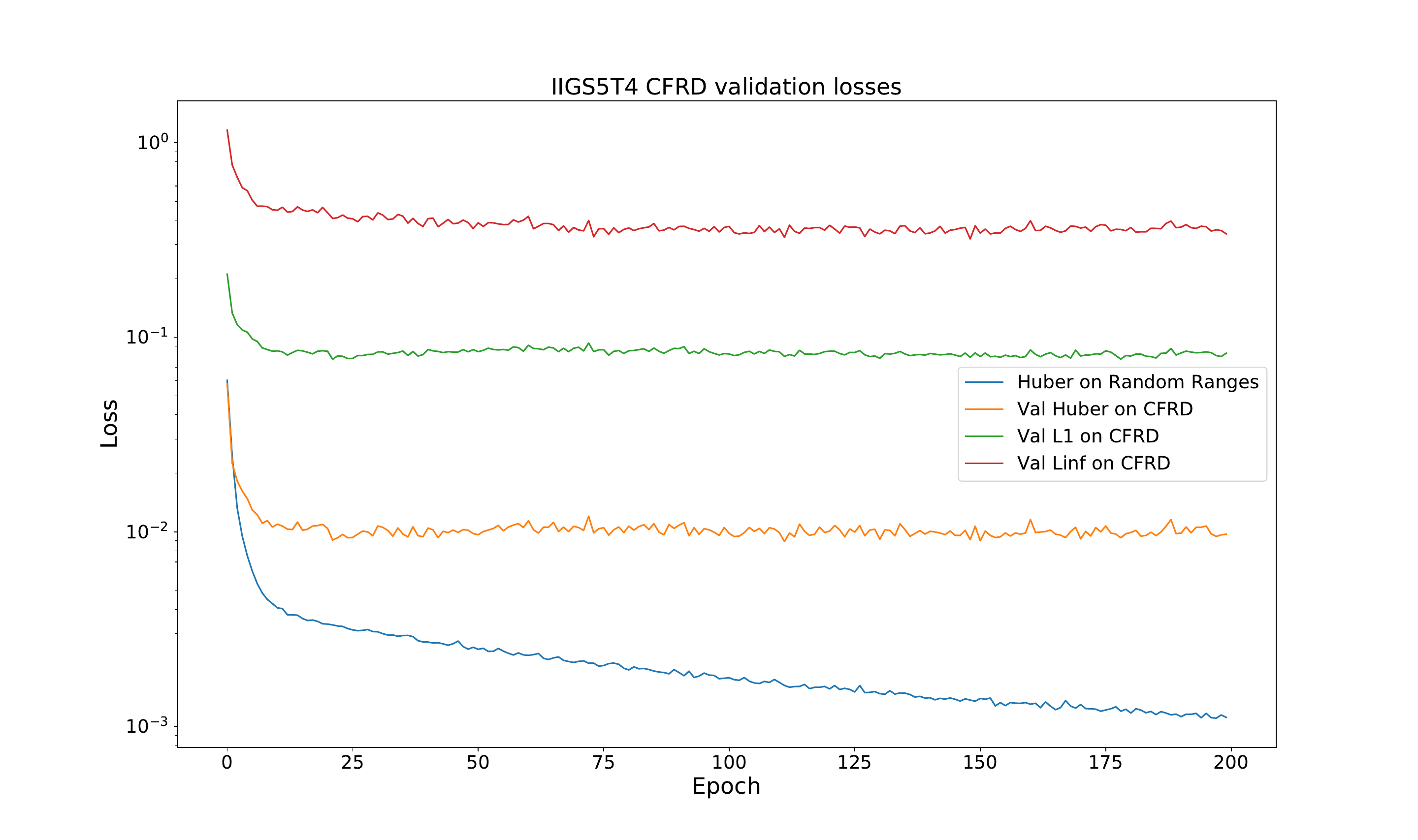}\hfill
\caption{GS: Training a network on random ranges and measuring validation error on CFR-D data.}
\end{figure}

\begin{figure}[H]
\centering
\includegraphics[trim=0 0 0 95,clip,width=\linewidth]{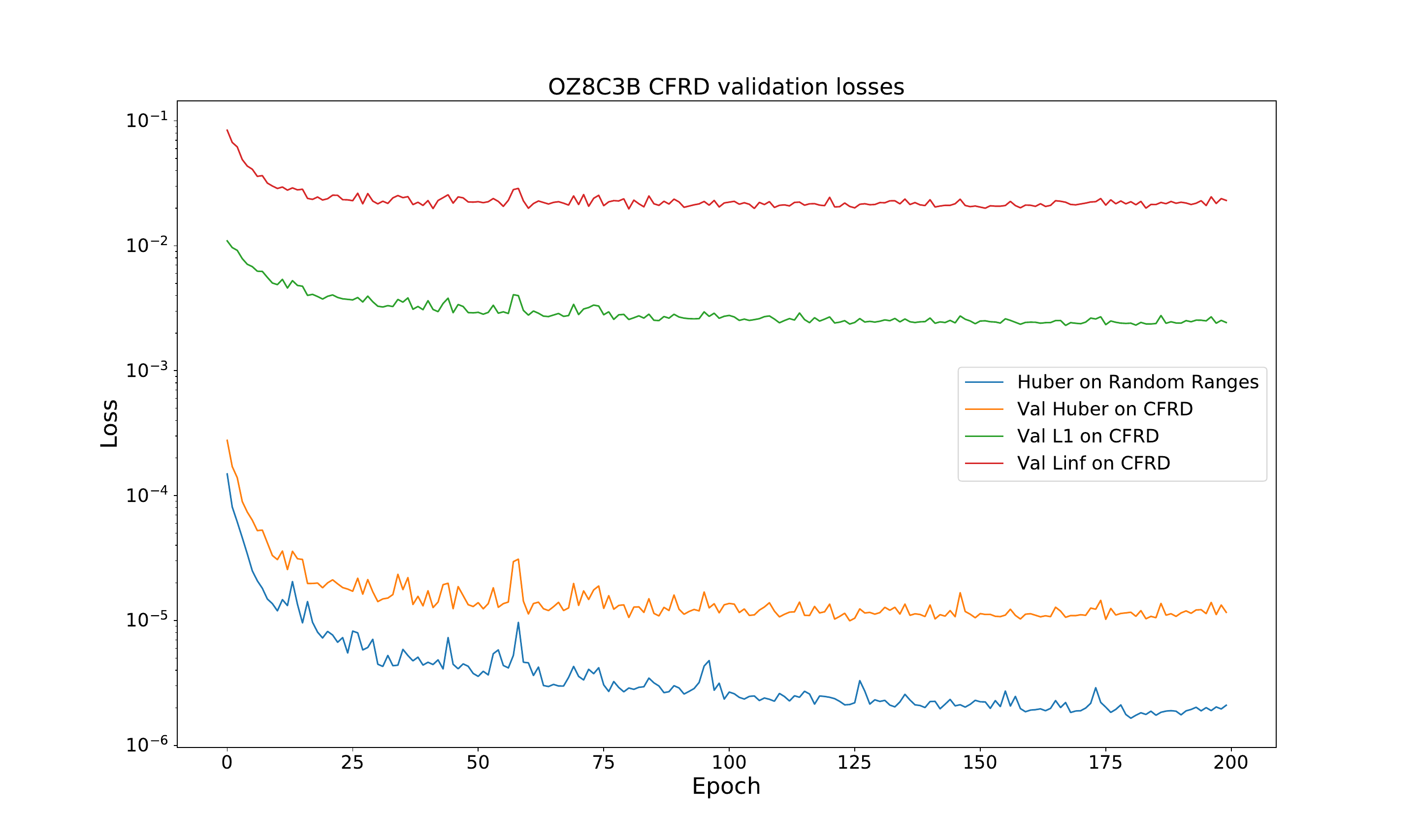}\hfill
\caption{OZ: Training a network on random ranges and measuring validation error on CFR-D data.}
\end{figure}

We observe substantially higher $l_\infty$ and $l_1$-errors on CFR-D data compared to random ranges, while the Huber errors are close to the one on random ranges. We conclude that the CFR-D data contains problematic ranges which are not covered by the training data. Despite that, it is clear (1) that the network can generalize to CFR-D data sufficiently precisely \emph{on average} and (2) that the validation loss on \emph{unseen random ranges} and provides enough confidence about the resulting exploitability as shown in Figure~\ref{fig:expl_vs_error2}.
\subsection{Comparison of \texorpdfstring{\DLCFR}{Depth-limited CFR} and CFR-D \texorpdfstring{\&}{,} Training Ranges}\label{sec:app:heatmaps}
\edit{Here, we show the remaining results for the two other domains. Recall, that we examined the game-solving behaviour of \DLCFR in a particular public state (meaning the which infosets are actually played into, i.e. ranges) and compared it to CFR-D in the same iteration and the closest (in terms of Euclidean distance) random range in the training data set for that particular range of \DLCFR. We show the ranges in the top row and the corresponding values, i.e., solutions to the subgames, in the bottom.}


\begin{figure}[H]
\centering
\includegraphics[width=\linewidth]{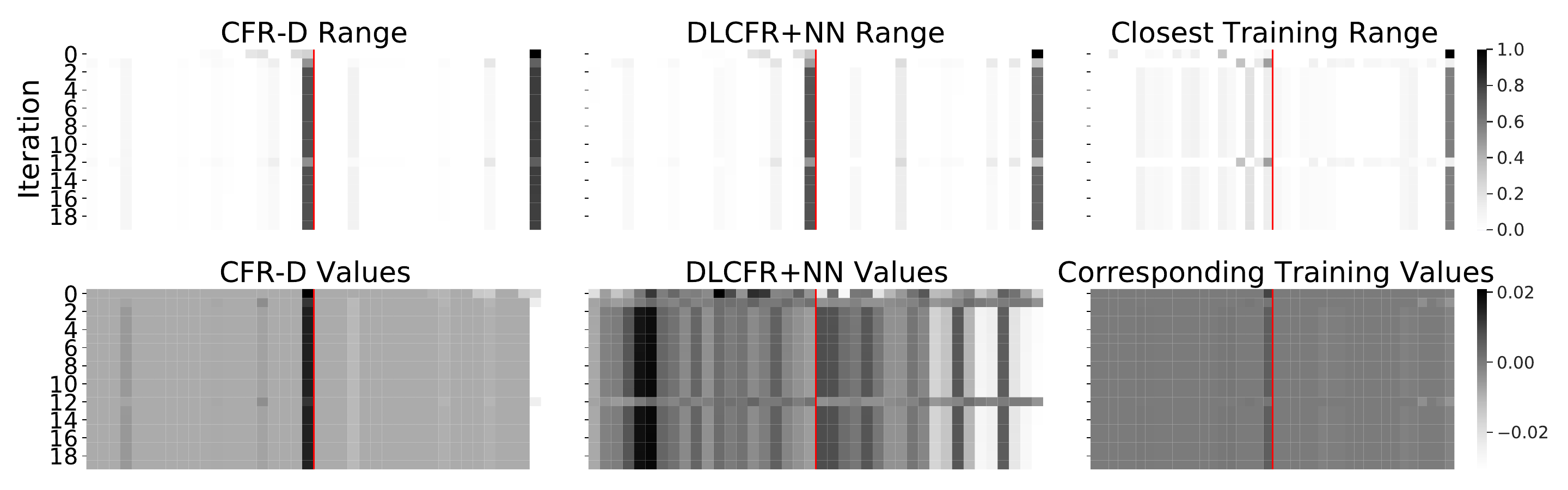}\hfill
\caption{\edit{We show ranges (top row) and values (bottom row) of the first 20 iterations of CFR-D (left),\DLCFR (middle) and closest training ranges (right) in goofspiel.
    The x-axis refers to a information set in a particular public state. The y-axis denotes the range/value in a given iteration.}}
\end{figure}

\begin{figure}[H]
\centering
\includegraphics[width=\linewidth]{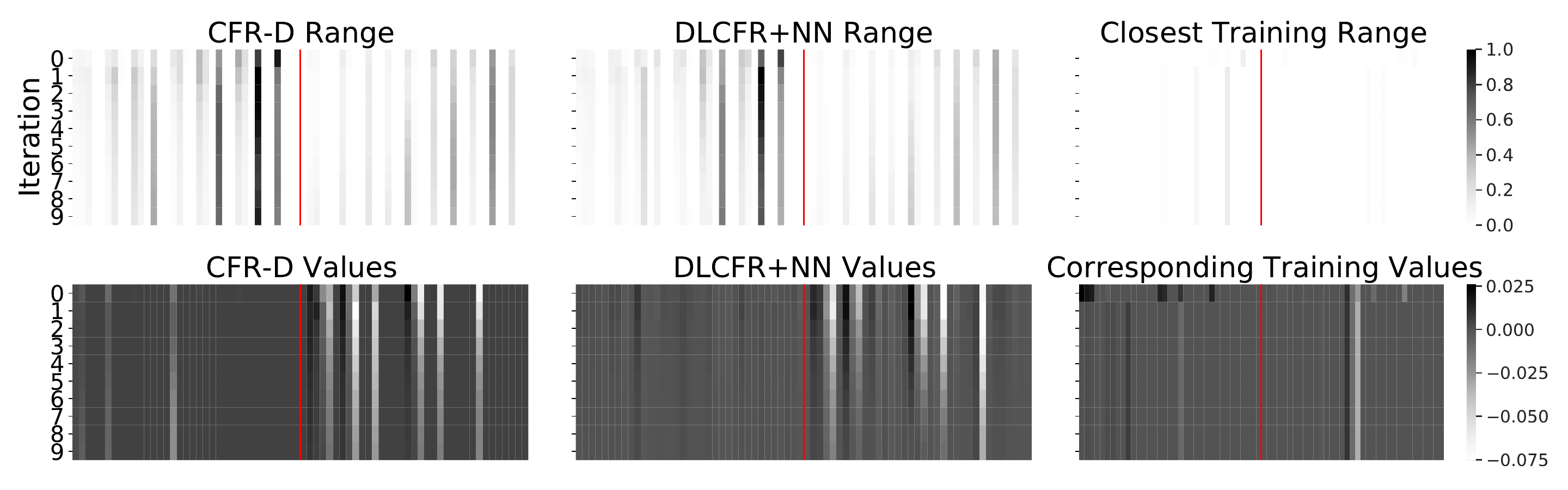}\hfill
\caption{\edit{We show ranges (top row) and values (bottom row) of the first 10 iterations of CFR-D (left),\DLCFR (middle) and closest training ranges (right) in oshi-zumo.
    The x-axis refers to a information set in a particular public state. The y-axis denotes the range/value in a given iteration.}}
\end{figure}

\edit{We observe very similar ranges requested by CFR-D and \DLCFR. The closest training ranges in all domains are still substantially different from \DLCFR. However, in goofspiel, the corresponding predicted values of \DLCFR are different from those computed by CFR-D, hinting at the possibility that \DLCFR was able to identify a different trunk equilibrium. Note that the value networks which were used to generate this data all achieved near-optimal exploitability.}



\subsection{Public State Cross-Validation}\label{sec:app:pubstatecrossval}
\edit{This subsection presents the results of the other two domains of the public state cross-validation experiment.
In all three figures below the x-axis denotes the public state index, while the y-axis denotes the $l_\infty$ validation loss for a network which has not seen the particular public state in training, the validation loss of a fully trained network and the average of 10 random networks.}


\begin{figure}[H]
\centering
\includegraphics[width=0.65\linewidth]{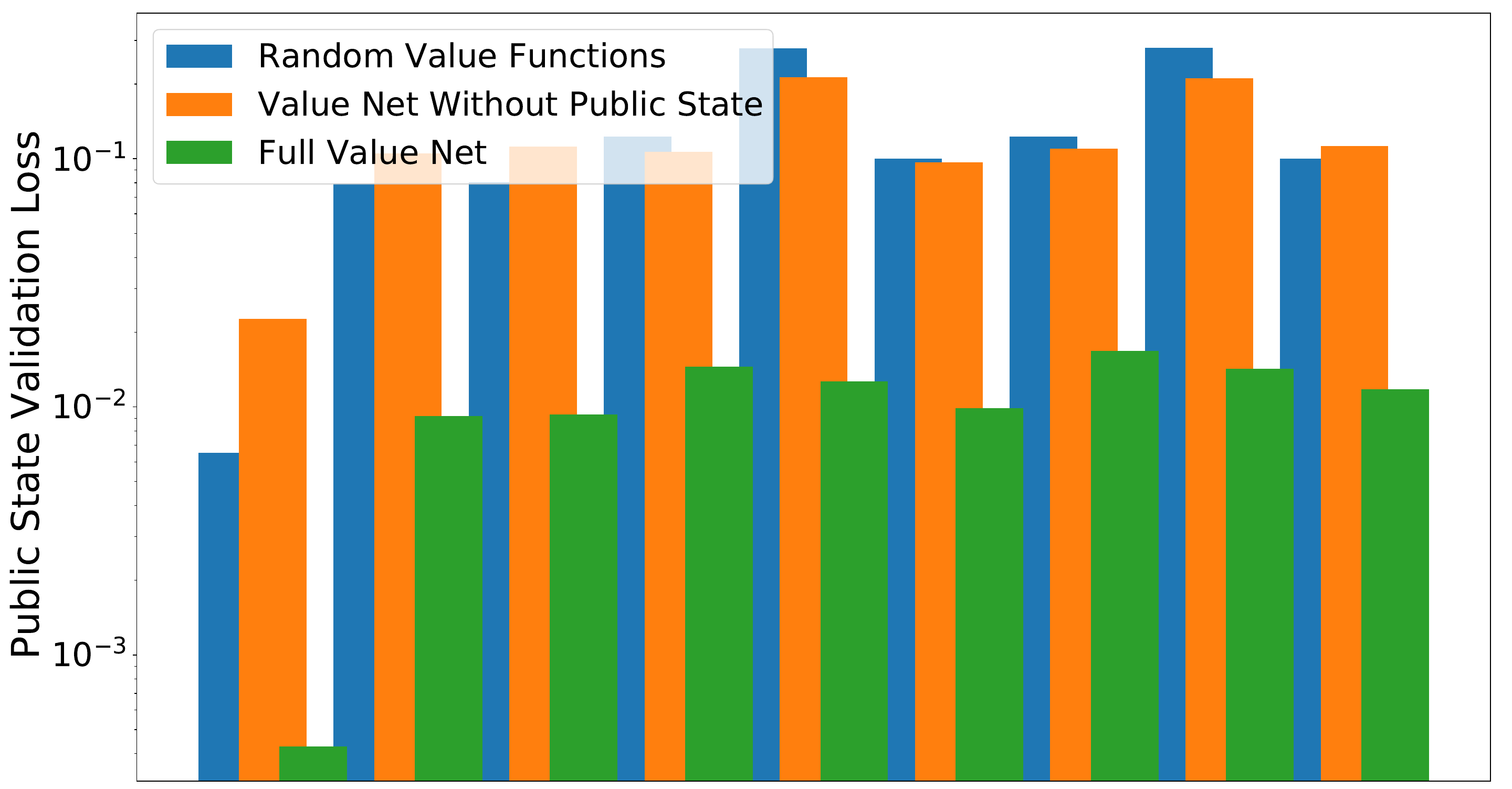}\hfill
\caption{Public state cross-validation in goofspiel.}
\end{figure}

\begin{figure}[H]
\centering
\includegraphics[width=0.7\linewidth]{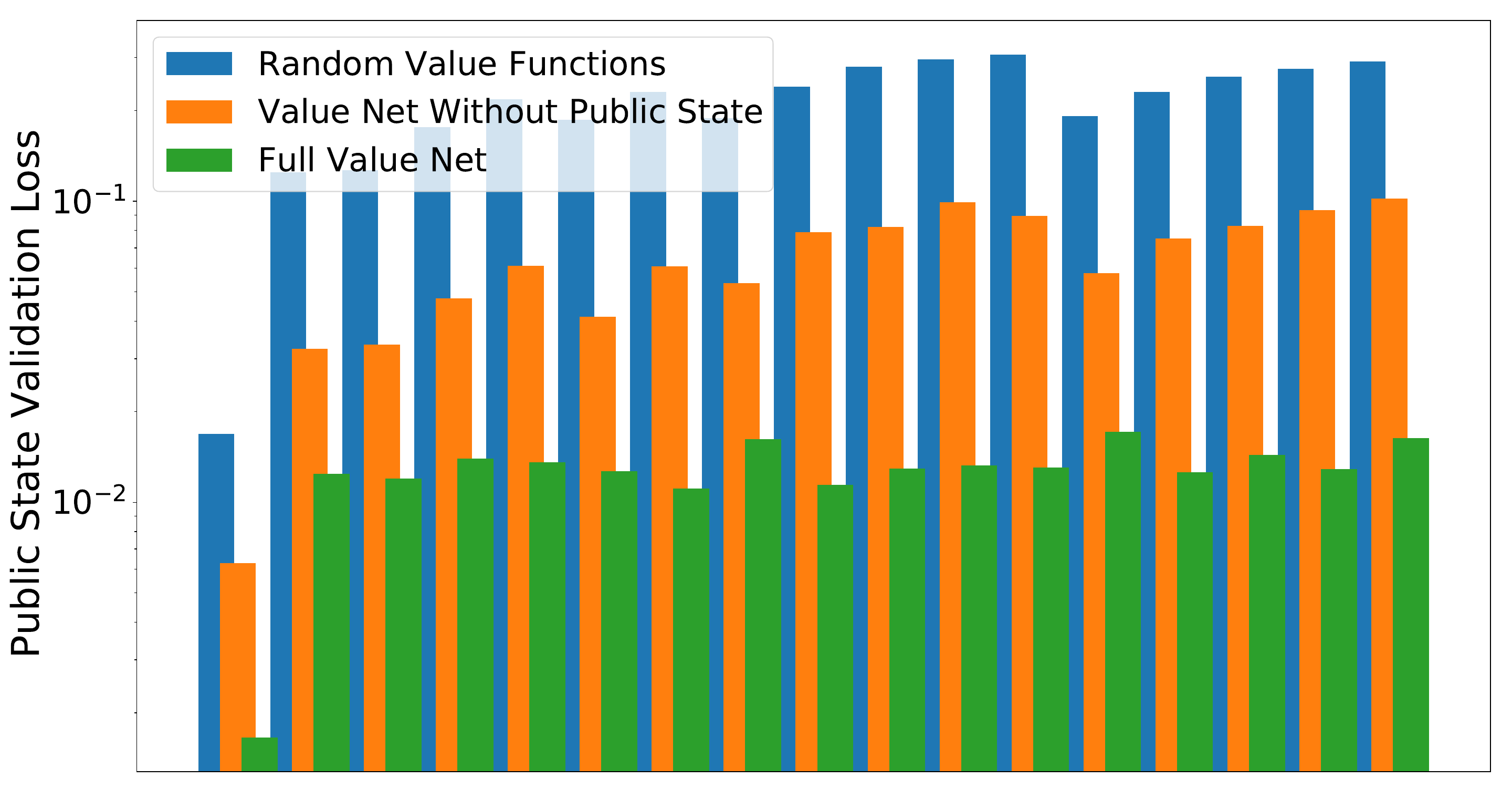}\hfill
\caption{Public state cross-validation in oshi-zumo.}
\end{figure}


\edit{We see that the network with withheld data is nearly as bad as the random network in goofspiel, somewhere in the middle in oshi-zumo, and nearly as good as the full value network in LH.
This is consistent with our hypothesis that the encoding is suitable for generalization:
Goofspiel has a very small number of public states, so there is not much to generalize from.
On the other hand, LH has many public states and there is a good chance that for any given $S$, there will be other public states that are strategically similar to $S$ --- and indeed, the network generalizes very well here.}

\end{document}